\documentclass[10pt,twocolumn,letterpaper]{article}

\usepackage{cvpr}              

\usepackage{graphicx}
\usepackage{amsmath}
\usepackage{amssymb}
\usepackage{booktabs}
\usepackage{enumitem}

\usepackage{algorithm}
\usepackage{bm}
\usepackage{algorithmic}
\usepackage{listings}
\usepackage{makecell}  
\usepackage{diagbox} 
\usepackage{mathtools} 
\usepackage{graphicx} 
\usepackage{multirow}
\usepackage{color}
\usepackage{xcolor}
\usepackage{xcolor,colortbl}

\definecolor{Gray}{gray}{0.5}
\definecolor{LightCyan}{rgb}{0.88,1,1}
\definecolor{cvprblue}{rgb}{0.21,0.49,0.74}

\usepackage[pagebackref,breaklinks,colorlinks,citecolor=cvprblue]{hyperref}

\def\paperID{2163} 
\def\confName{CVPR}
\def\confYear{2024}

\usepackage[capitalize]{cleveref}
\crefname{section}{Sec.}{Secs.}
\Crefname{section}{Section}{Sections}
\Crefname{table}{Table}{Tables}
\crefname{table}{Tab.}{Tabs.}

\newcommand{\topparaheading}[1]{\vspace{-0.2em}\paragraph{#1.}}

\newcolumntype{a}{>{\columncolor{Gray}}c}
\newcolumntype{b}{>{\columncolor{white}}c}

\begin{document}


\title{DiffusionMat: Alpha Matting as Sequential Refinement Learning}

\author{Yangyang Xu$^1$,
       Shengfeng He$^2$,
       Wenqi Shao$^3$,
       Kwan-Yee K. Wong$^1$,
       Yu Qiao$^3$,
       and Ping Luo$^{1,3}$
       \\
       {$^1$Deapartment of Computer Science, The University of Hong Kong} \\
       {$^2$School of Computing and Information Systems, Singapore Management University}\\
       {$^3$Shanghai AI Laboratory}\\
      }

\maketitle


\begin{abstract}
In this paper, we introduce DiffusionMat, a novel image matting framework that employs a diffusion model for the transition from coarse to refined alpha mattes. Diverging from conventional methods that utilize trimaps merely as loose guidance for alpha matte prediction, our approach treats image matting as a sequential refinement learning process. This process begins with the addition of noise to trimaps and iteratively denoises them using a pre-trained diffusion model, which incrementally guides the prediction towards a clean alpha matte.
The key innovation of our framework is a correction module that adjusts the output at each denoising step, ensuring that the final result is consistent with the input image's structures. We also introduce the Alpha Reliability Propagation, a novel technique designed to maximize the utility of available guidance by selectively enhancing the trimap regions with confident alpha information, thus simplifying the correction task. To train the correction module, we devise specialized loss functions that target the accuracy of the alpha matte's edges and the consistency of its opaque and transparent regions. We evaluate our model across several image matting benchmarks, and the results indicate that DiffusionMat consistently outperforms existing methods. Project page at~\url{https://cnnlstm.github.io/DiffusionMat}.
\end{abstract}

\section{Introduction}
\label{secintro}

\begin{figure}[t]
    \centering
    \captionsetup[subfloat]{labelformat=empty,justification=centering}
    \hspace{-2.5mm}
    \subfloat[\scriptsize{Initial}]{
     \begin{minipage}{0.16\linewidth}
     \includegraphics[width=\linewidth]{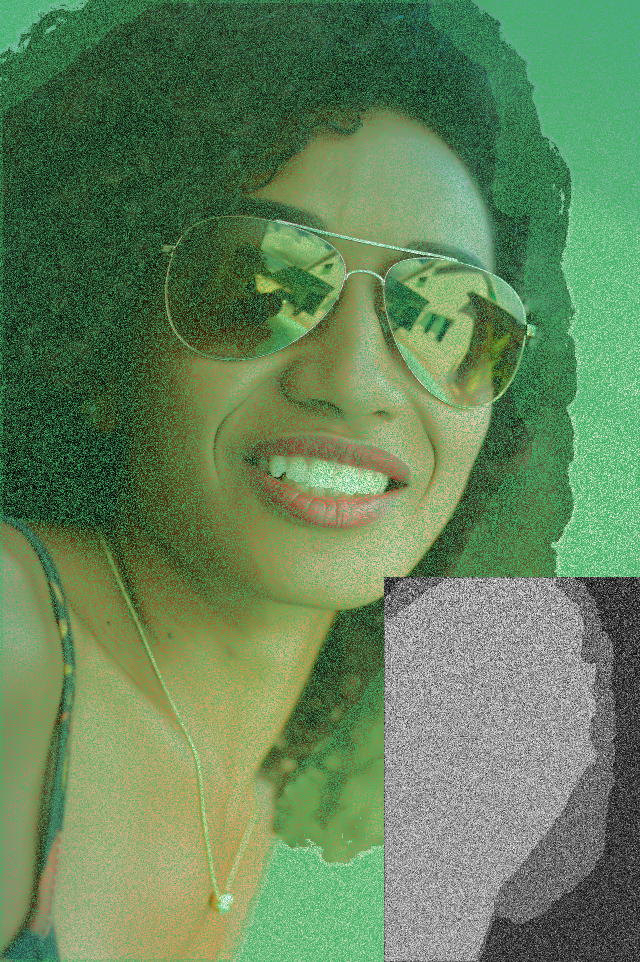}
     \includegraphics[width=\linewidth]{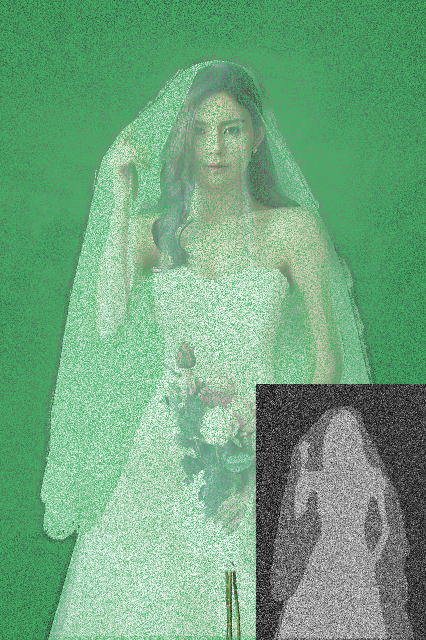}
     \end{minipage}
     }
    \hspace{-2.5mm}
    \subfloat[\scriptsize{$t={200}$}]{
     \begin{minipage}{0.16\linewidth}
     \includegraphics[width=\linewidth]{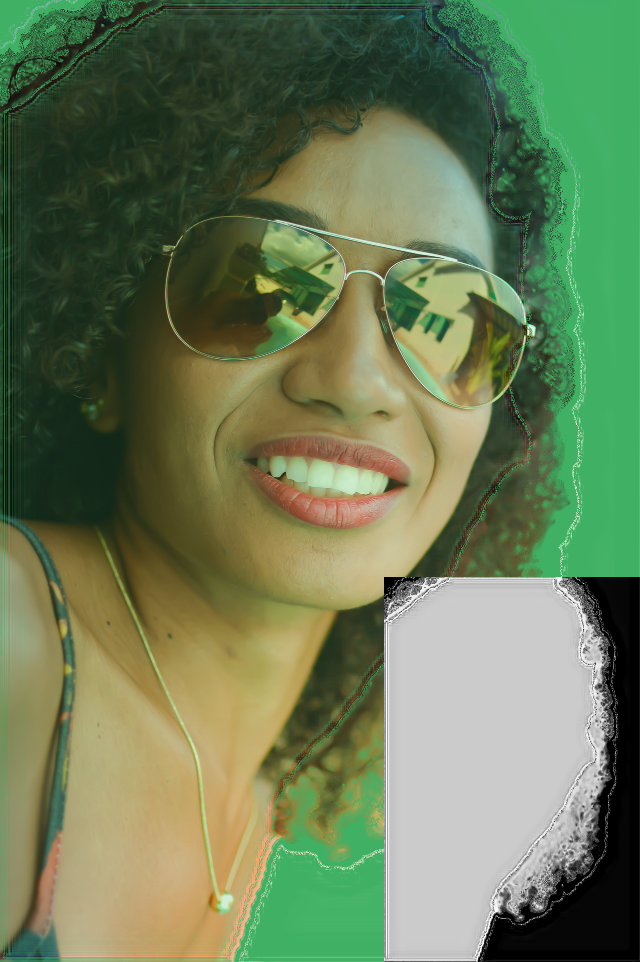}
     \includegraphics[width=\linewidth]{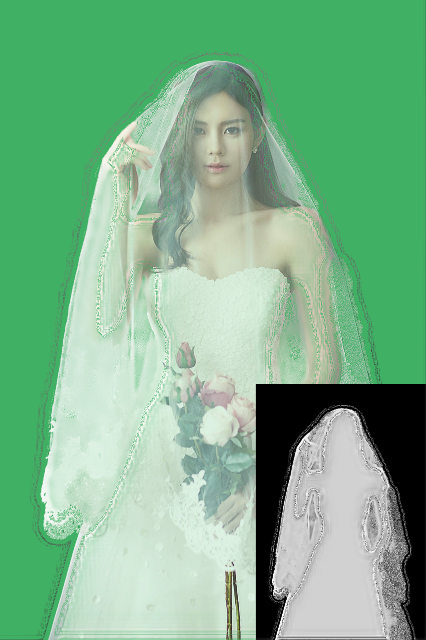}
     \end{minipage}
     }
    \hspace{-2.5mm}
    \subfloat[\scriptsize{$t={150}$}]{
     \begin{minipage}{0.16\linewidth}
     \includegraphics[width=\linewidth]{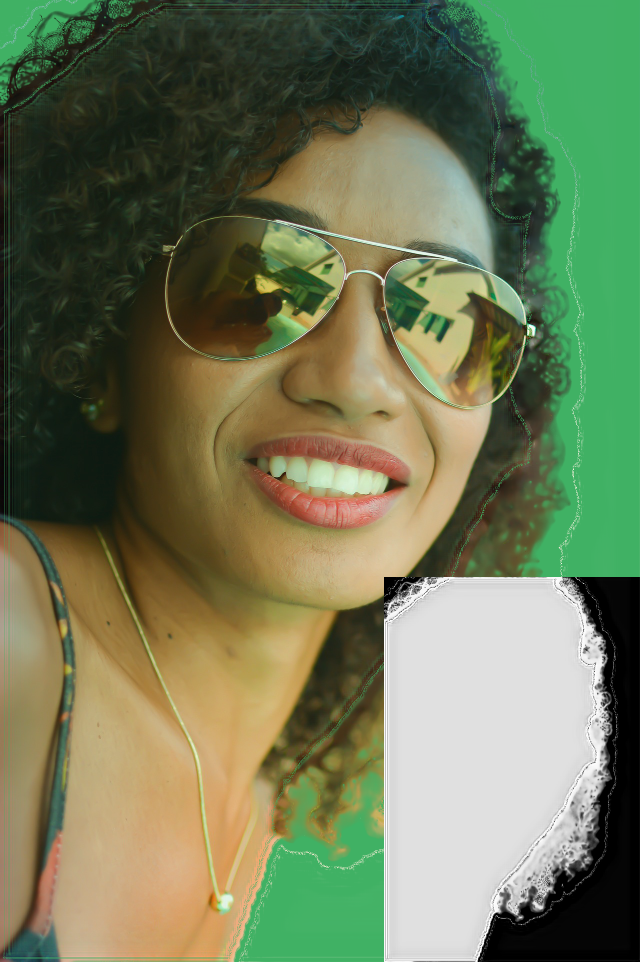}
     \includegraphics[width=\linewidth]{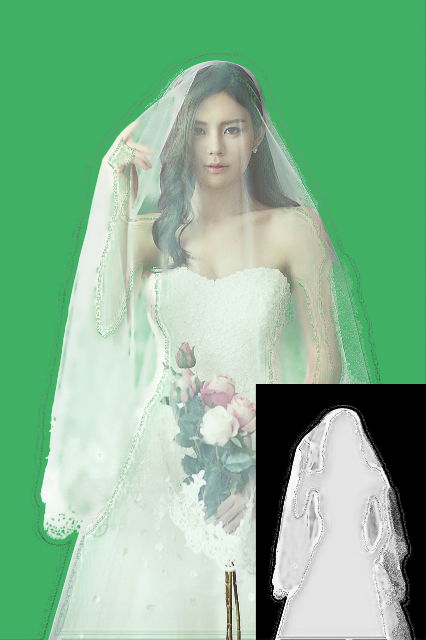}
     \end{minipage}
     }
     \hspace{-2.5mm}
    \subfloat[\scriptsize{$t={100}$}]{
     \begin{minipage}{0.16\linewidth}
     \includegraphics[width=\linewidth]{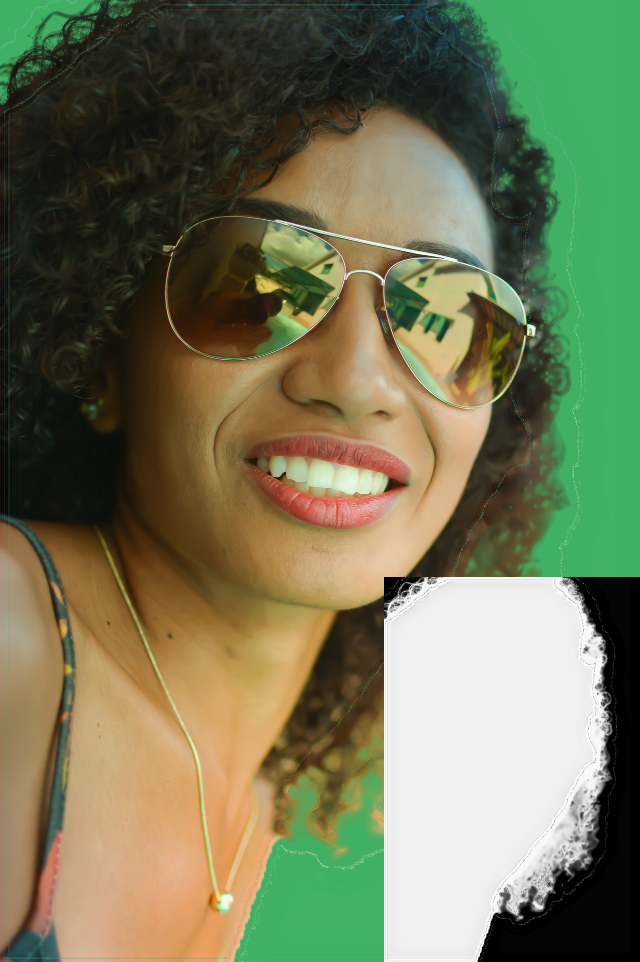}
     \includegraphics[width=\linewidth]{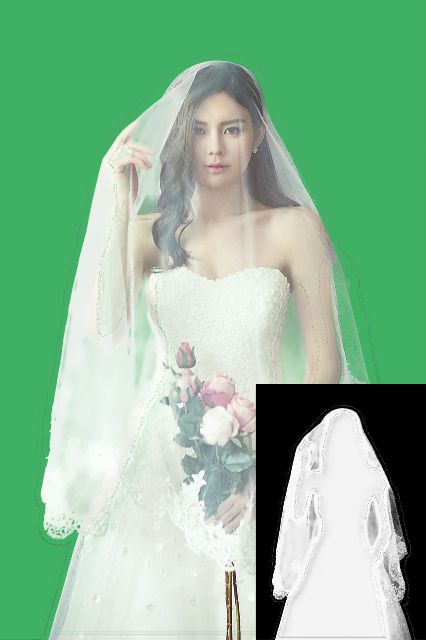}
     \end{minipage}
     }
    \hspace{-2.5mm}
    \subfloat[\scriptsize{$t={50}$}]{
     \begin{minipage}{0.16\linewidth}
     \includegraphics[width=\linewidth]{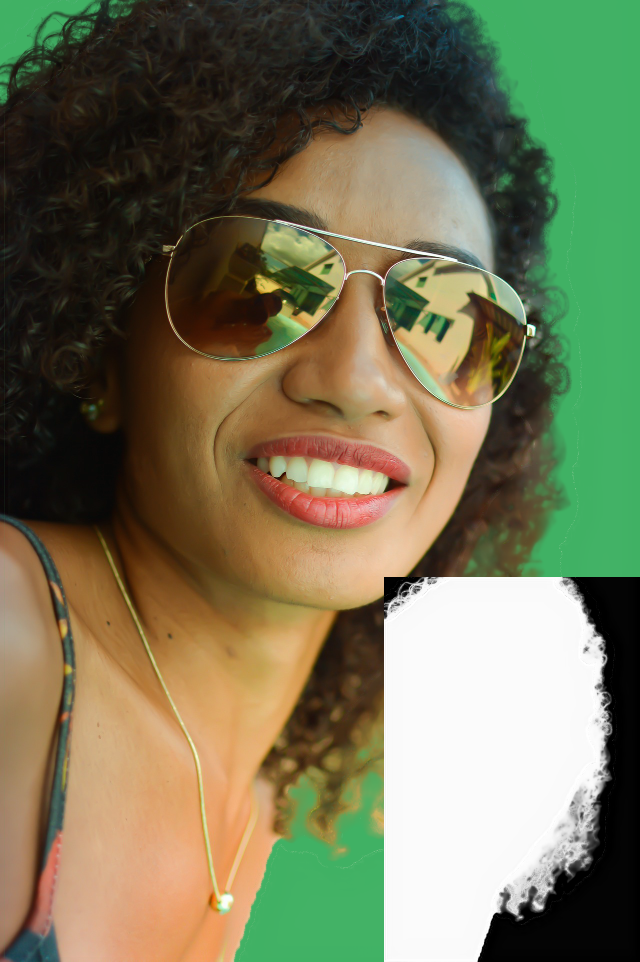}
     \includegraphics[width=\linewidth]{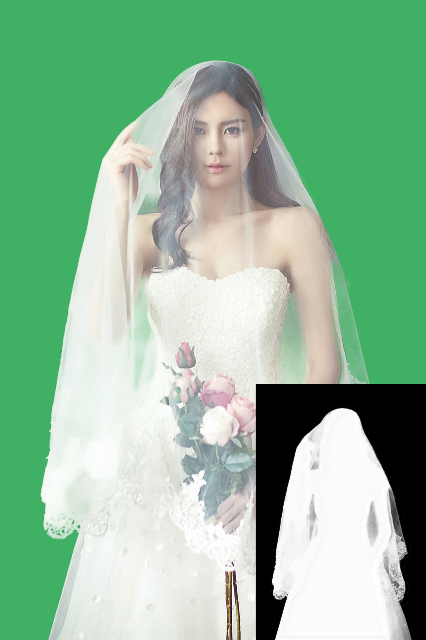}
     \end{minipage}
     }
    \hspace{-2.5mm}
    \subfloat[\scriptsize{$t={0}$}]{
     \begin{minipage}{0.16\linewidth}
     \includegraphics[width=\linewidth]{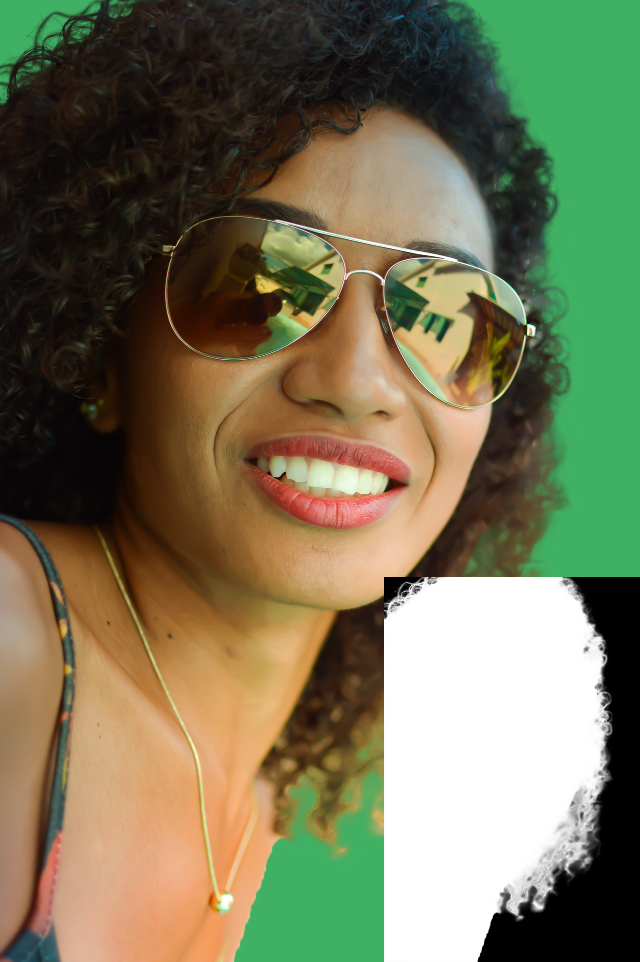}
     \includegraphics[width=\linewidth]{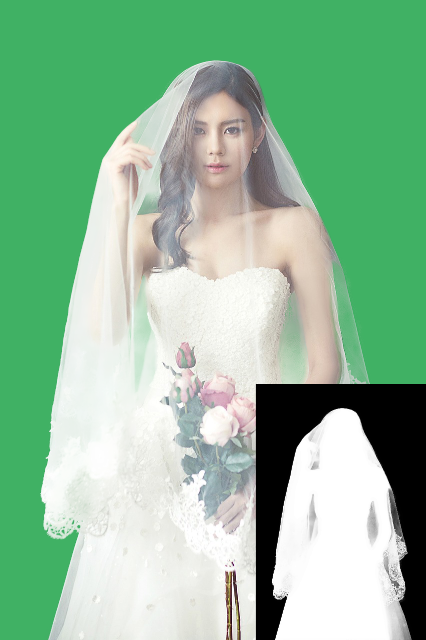}
     \end{minipage}
     }
\vspace{-2mm}
\caption{Our DiffusionMat transforms alpha matting into a sequential refinement process, progressing from a noise-injected trimap to a precisely defined alpha matte (see bottom-right). The resulting composited foregrounds are showcased against a green screen background.}
\vspace{-5mm}
\label{fig:teaser}
\end{figure}

Image matting, a critical process in extracting objects from images, has indispensable applications in numerous fields, including media production, virtual reality, and intelligent editing tools~\cite{4547428,sengupta2020background,cai2019disentangled,xu2017deep,lu2019indices}. The process is mathematically modeled as a decomposition of an image $I$ into its constituent foreground $F$ and background $B$, combined via an alpha matte $\alpha$. The composite model is given by:
\begin{equation}
I_i = \alpha_i F_i + (1 - \alpha_i) B_i,  \quad \alpha_i \in [0,1],
\label{eq:matting}
\end{equation}
where each pixel $i$ entails solving for the transparency $\alpha_i$ alongside the colors of $F_i$ and $B_i$.

The inherent challenge in matting stems from its ill-posed structure, in which each pixel's calculation involves estimating more variables than the available observations. Traditional approaches~\cite{xu2017deep,yang2018active,li2020natural,tang2019learning,yu2021mask,park2022matteformer} partially mitigate this issue by leveraging auxiliary inputs, such as trimaps, to provide boundary conditions for the unknown regions. However, this method of one-off prediction has intrinsic limitations. It assumes a static relationship between the auxiliary input and the image content, which can lead to inaccuracies due to oversimplified assumptions about the spatial distribution of opacity within the unknown regions.

To overcome these limitations, we introduce a reconceptualized framework, DiffusionMat, that embraces a sequential refinement learning strategy. Unlike one-off prediction models that generate a single alpha matte in a direct manner, our approach progressively refines the matte through a series of informed iterations. This methodology aligns with the stochastic nature of diffusion processes, which have demonstrated exceptional capability in capturing complex data distributions in recent studies~\cite{austin2021structured,gu2022vector,rombach2022high,kingma2021variational,dhariwal2021diffusion}.

DiffusionMat operates on the premise that the matting of unknown regions can be enhanced incrementally, benefiting from each iteration's feedback to correct and refine the predictions. This is in stark contrast to one-off predictions that lack the flexibility to revise and improve upon initial estimations. Our correction module plays a pivotal role in this iterative process, fine-tuning the matte at each step to ensure fidelity to the input image's structural nuances. Particularly, our method commences with the training of an unconditional diffusion model on a comprehensive dataset of alpha mattes, preparing it to understand and generate the distribution of matting details. We then inject random noise into a trimap in a controlled manner to mimic real-world unpredictability, enabling our model to explore a diverse set of potential alpha matte solutions. The noised trimap is fed into the diffusion model, which denoises it step by step, iteratively refining towards a high-quality alpha matte ({see in Fig.~\ref{fig:teaser}}). This stochastic process naturally incorporates variability, which we harness rather than constrain, to produce a range of plausible matte solutions.

To achieve a deterministic result aligned with the input image $I$, our framework incorporates a deterministic denoising process using Denoising Diffusion Implicit Models (DDIM) inversion~\cite{song2021denoising}. This process inverts a ground truth alpha matte through the diffusion model to establish a reference trajectory for the denoising steps. A correction module, augmented with an image encoder, ensures that the intermediate outputs adhere to this trajectory, focusing particularly on the unknown regions as identified by the trimap. We further propose an Alpha Reliability Propagation (ARP) to regulate the intermediate denoised results by the determined regions. This operation enhances the model's focus and efficiency in learning, allowing for rapid refinement within the most ambiguous regions of the trimap.

In summary, our primary contributions are as follows:
\begin{itemize}[leftmargin=*,nosep]
    \item We present the first attempt to utilize diffusion models for image matting. This marks a novel step in the context of matting tasks, demonstrating the potential of diffusion models in this domain.
    \item We reconceptualize image matting as a process of transforming trimap guidance into precise alpha mattes, employing a novel correction strategy to align random denoising trajectories with the GT ones.
    \item We propose an Alpha Reliability Propagation module to accelerate and regulate the sequential learning process.
    \item Our extensive experiments on both portrait and non-portrait matting datasets validates that DiffusionMat sets a new benchmark for state-of-the-art performance.
\end{itemize}

Through this approach, we demonstrate that the proposed DiffusionMat not only addresses the core challenges of image matting but also opens up a new avenue to apply sequential learning to complex vision tasks.

\section{Related Works}
\label{sec:related}

\topparaheading{Natural Image Matting}
Image matting techniques can be broadly divided into sampling-based and affinity-based methods. Sampling-based methods~\cite{he2011global,chuang2001bayesian,shahrian2013improving,wang2007optimized} estimate alpha values by drawing color samples from foreground and background regions. In contrast, affinity-based methods~\cite{aksoy2017designing,levin2007closed,chen2013knn} extrapolate alpha values into unknown regions based on the relational affinities between pixels. Recent advancements have seen the application of deep learning to image matting. Xu~\etal~\cite{xu2017deep} were pivotal in introducing a dedicated image-matting dataset alongside an encoder-decoder network tailored for this task. Lutz~\etal~\cite{lutz2018alphagan} further enhanced matting frameworks by integrating generative adversarial networks, while Lu~\etal~\cite{lu2019indices} developed IndexNet, which dynamically adapts to feature maps for index generation. Li~\etal~\cite{li2020natural} introduced the concept of guided contextual attention for the effective global communication of opacity information. Building upon the transformer architecture, Park~\etal~\cite{park2022matteformer} implemented a trimap prior token to leverage the self-attention mechanism for improved matting. Cai~\etal~\cite{cai2022transmatting} focused on the challenges presented by transparent objects, employing a design with an expansive receptive field. Furthermore, Yu~\etal~\cite{yu2021mask} proposed a concatenate network that enhances the prediction of the unknown region. Diverging from the conventional end-to-end matting paradigm, our work redefines the process as a transformation from noisy trimap to alpha matte, capitalizing on the generative capacities of diffusion models. This approach not only acknowledges but actively engages with the generative aspects of diffusion to inform the matting process.

\topparaheading{Diffusion Models}
Diffusion models have recently risen to prominence for their superior generative performance in various domains~\cite{austin2021structured,gu2022vector,rombach2022high,kingma2021variational,dhariwal2021diffusion}. These models function by iteratively refining from random noise to create coherent samples, showcasing impressive capabilities in image synthesis and restoration. Notable implementations include GDP~\cite{fei2023generative}, which uses pre-trained diffusion models for image restoration and enhancement, and SDEdit~\cite{meng2021sdedit}, which synthesizes realistic images through stochastic differential denoising. Stable Diffusion further extends this utility to diverse applications such as image and video manipulation, language-to-image translation, and synthetic media generation~\cite{zhang2023adding,qi2023fatezero,yang2023rerender}. In this paper, we harness the generative capabilities of pre-trained diffusion models to address image matting, a fundamental perceptual task.

\topparaheading{Diffusion Models for Perception Tasks}
The versatility of diffusion models, traditionally celebrated for generative tasks, is now being explored for perceptual challenges. Segdiff~\cite{amit2021segdiff} marked an initial foray into diffusion-driven segmentation, while Chen~\etal~\cite{chen2022generalist} expanded the utility with the Bit Diffusion model~\cite{chen2022analog}, adept at panoptic segmentation across still images and videos. In a similar vein, Boah~\etal~\cite{kim2022diffusion} combined diffusion models with adversarial training for enhanced vascular segmentation accuracy. Baranchuk~\etal~\cite{baranchuk2021label} investigated the potential of diffusion model activations in capturing the semantic segmentation of images. Building on this, DiffusionDet~\cite{chen2022diffusiondet} introduced a paradigm shift in object detection by conceptualizing it as a denoising diffusion sequence. The recent VPD model by Zhao~\etal~\cite{zhao2023unleashing} exploits a pre-trained text-to-image diffusion model's semantic prowess for a range of visual perception tasks, and DDP by Ji~\etal~\cite{ji2023ddp} applies it to various dense visual prediction scenarios. Our work pioneers the application of diffusion models for image matting, leveraging their generative strengths to advance this intricate perceptual task.

\section{Approach}
transforming trimap guidance into precise alpha mattes
In this section, we present~DiffusionMat, a novel diffusion-based framework for image matting. Our key idea is to transform the trimap guidance into precise alpha matte by a novel correction strategy. We first present the background of diffusion models, and then describe how we correct the denoised results via deterministic denoising and our alpha reliability propagation.

\subsection{Preliminaries: Diffusion Models}

Diffusion models are classes of likelihood-based models that learn the distribution space in a gradual denoising process~\cite{austin2021structured,gu2022vector,rombach2022high,kingma2021variational,dhariwal2021diffusion}. A diffusion model consists of a noising process and a reverse denoise sampling process. In the forward process, the diffusion model adds the noise to the data gradually via a Markov chain. Each forward step can be represented as:
\[
q({x}_t | {x}_{t-1})=\mathcal{N}\left(\sqrt{1-\beta_t} {x}_{t-1}, \beta_t \mathbf{I}\right),
\]
where $\{\beta_t\}_{t=0}^T$ are variance schedule. The latent variable ${x}_t$ can be denoted as:
\[
{x}_t=\sqrt{{\boldsymbol{\alpha}}_t}{x}_0 + \sqrt{1-{\boldsymbol{\alpha}}_t}{\epsilon},{\epsilon} \sim \mathcal{N}(\mathbf{0},\mathbf{I}),
\]
where ${\boldsymbol{\alpha}}_t:=1-\beta_t$~\footnote{Note that the bold version $\boldsymbol{\alpha}$ is not the alpha matte $\alpha$ (regular version).}. During training, the diffusion model ${\epsilon}_\theta$ is used for predicting ${\epsilon}$ from ${x}_t$ with following equation.
\[
\mathcal{L}_{\mathrm{DM}}=\| {\epsilon}-{\epsilon}_\theta ({x}_t, t) \|^2_2.
\]

During the inference, the data can be sampled using the following Denoising Diffusion Probabilistic Models (DDPM)~\cite{ho2020denoising} reverse diffusion process:

\[
{x}_{t-1}=\frac{1}{\sqrt{1-\beta_t}}\left({x}_t-\frac{\beta_t}{\sqrt{1-\boldsymbol{\alpha}_t}} {\epsilon}_\theta\left({x}_t, t\right)\right)+\sigma_t {z},
\]
where ${z} \sim \mathcal{N}(\mathbf{0},\mathbf{I})$. Moreover, Song~\etal~propose the DDIM~\cite{song2021denoising} denoising process with fewer steps. The DDIM sampling is a deterministic process, which allows us to fully invert the real data to its latent variables.



\begin{figure}
\centering
\includegraphics[width=0.35\textwidth]{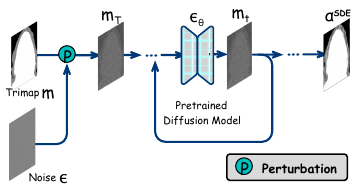}
\vspace{-3mm}
\caption{Pipeline of SDEdit~\cite{meng2021sdedit} for synthesizing alpha matte under the trimap guidance.}
\label{fig:sdeit}
\vspace{-5mm}
\end{figure}

\begin{figure*}
\centering
\vspace{-5mm}
\includegraphics[width=0.99\textwidth]{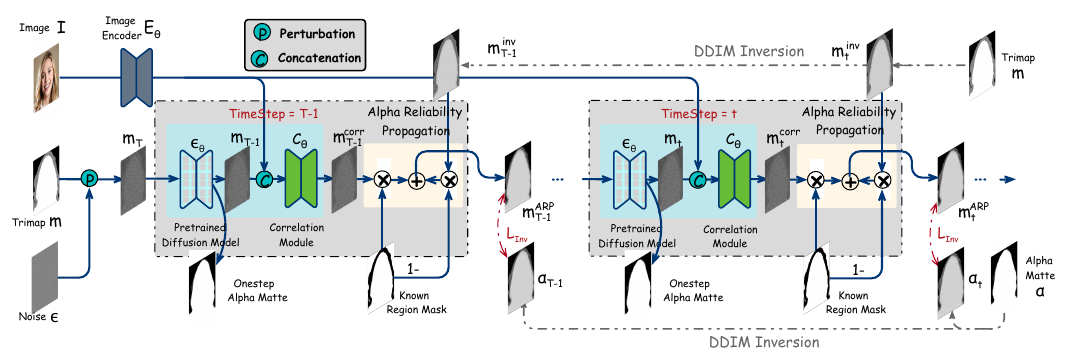}
\vspace{-3mm}
\caption{Pipeline of our~DiffusionMat. Given a trimap guidance $m$, we first perturb it with random noise $\epsilon$ to a noised mask $m_T$ at time step $T$. In each denoising timestep $t$, we propose the deterministic denoising that corrects the immediate results $m_t$ with an image encoder $\texttt{E}_{\theta}$ and correction module $\texttt{C}_{\theta}$. We further propose an Alpha Reliability Propagation that regulates the intermediate denoised results by the determined unknown regions. Then the output result ${m}^{ARP}_{t}$ is aligned with the corresponding GT inverted results $\alpha_t$. We also predict a clean alpha matte $\tilde{\alpha}^0_{t}$ in each step and minimize its distance with the GT $\alpha$. Note that loss $\mathcal{L}_{\mathrm{\alpha}}$ and $\mathcal{L}_{\mathrm{Comp}}$ are omitted for simplicity.}
\label{fig:framework}
\vspace{-5mm}
\end{figure*}

\subsection{Procedure}

Given an input image $I$ along with its trimap guidance $m$, our objective is to derive the corresponding alpha matte $\alpha$. Diverging from existing one-off prediction approaches, we introduce a sequential refinement learning strategy that transforms noised trimap to clean alpha matte.

The trimap-to-alpha transformation can be seen as a conditional image generation process, SDEdit~\cite{meng2021sdedit} provides a simple pipeline for this generation, it synthesizes output images under given guidance by iteratively denoising through a stochastic differential equation. Here, we present the pipeline that applies SDEdit to synthesize an alpha matte under trimap guidance. As illustrated in Fig.~\ref{fig:sdeit}, SDEdit begins with the trimap guidance $m$, which is initially perturbed with random noise to create a noised mask $m_T$ at time step $T$. Subsequently, this corrupted trimap $m_T$ undergoes denoising in the diffusion model's sampling process through iterative application of Eq.~\ref{denoise}, leading to the final alpha matte denoted as $\alpha^{\texttt{SDE}}$. This process can be represented as follows:






\begin{equation}
{{m}_T=\sqrt{{\boldsymbol{\alpha}}_T}m + (1-{\boldsymbol{\alpha}}_T){\epsilon},{\epsilon} \sim \mathcal{N}(\mathbf{0},\mathbf{I})},
\label{addnoise}
\end{equation}
\begin{equation}
\begin{aligned}
{m}_{T-1}=&\sqrt{\boldsymbol{\alpha}_{T-1}} \underbrace{(\frac{{m}_T-\sqrt{1-\boldsymbol{\alpha}_t} \epsilon_\theta({m}_T;T)}{\sqrt{\boldsymbol{\alpha}_T}})}_{\text {``predicted clean alpha matte in one step ''} } +\\ &\underbrace{\sqrt{1-\boldsymbol{\alpha}_{T-1}} \cdot \epsilon_\theta({m}_T;T)}_{\text {``direction pointing to } {m}_T \text {"}},
\label{denoise}
\end{aligned}
\end{equation}
where ${\epsilon}_{\theta}$ is the unconditional diffusion model pre-trained on large scaled alpha matte dataset, and the first term is the clear alpha matte $\tilde\alpha^0_t$ predicted in one step, and the second term is the direction pointing to $\boldsymbol{m}_T$.


\subsubsection{Deterministic Denoising}

In the SDEdit, a noised guidance trimap can be denoised to arbitrary alpha matte due to the randomness brought by the random noise. However, as a perception task, image matting has one only deterministic alpha matte. For obtaining the precisely alpha matte, we correct the intermediate denoised results using the GT inverted guidance~\cite{mokady2023null}. Particularly, given the GT alpha matte $\alpha$, we invert it to the pre-trained diffusion model via DDIM inversion, and obtain the deterministic inversion trajectory, which can be used as the supervision during correction.

In the case of SDEdit, a noised guidance trimap has the potential to be denoised into an arbitrary alpha matte due to the randomness introduced by the random noise. However, image matting, being a classic perception task, necessitates a single deterministic alpha matte. To achieve this precise alpha matte, we correct the denoised results under the supervision of the GT inverted guidance. Different from~\cite{mokady2023null}, we use a learnable correction module for the revision.


Specifically, starting with the GT alpha matte $\alpha$, we perform an inversion process to map it onto the pre-trained diffusion model via DDIM inversion, and yields a deterministic inversion trajectory.



In each denoised timestep $t$, we correct the intermediate denoised result $m_t$ using an image encoder $\texttt{E}_{\theta}$ and a correction module $\texttt{C}_{\theta}$. As illustrated in Fig.~\ref{fig:framework}, we initiate the process by encoding the image $I$ with the image encoder $e_{\theta}$, resulting in the image feature $f_{I}$. Subsequently, we concatenate $f_{I}$ with ${m}_{t}$ and feed this combined input to the correction module $c_{\theta}$, generating the corrected denoising result ${m}^{corr}_{t}$. These procedures can be expressed as follows:
\[
f_{I} = \texttt{E}_{\theta}(I),
\]
\[
{m}^{corr}_{t}  = \texttt{C}_{\theta}(\texttt{cat}(f_{I};{m}_{t}); t),
\]
where $\texttt{cat}(\cdot;\cdot)$ denotes the concatenation operator. 


\subsubsection{Alpha Reliability Propagation}

Learning such a correction for all pixels is challenging and unnecessary since the difference between the trimap and alpha matte only exists in the unknown region. To alleviate the learning complexity and concentrate the module's efforts on the unknown regions, we introduce the ``Alpha Reliability Propagation'' (ARP) module that regulates the intermediate denoised results by the known regions. 


Particularly, we first invert the trimap $m$ to diffusion model via DDIM inversion and get the inversion trajectory $m^{\texttt{inv}}_{t}$, then we replace the known regions' values on ${m}^{corr}_{t}$ with those in $m^{\texttt{ddim}}_t$ according the known mask $m_\texttt{known}$, that is:
\[
{m}^{ARP}_{t}  = {m}^{corr}_{t} \times (1-m_\texttt{known}) + m^{\texttt{inv}}_{t} \times m_\texttt{known}.
\]

Then we denoise ${m}^{ARP}_{t}$ to its next timestep following Eq.~\ref{denoise}. Note that the first term in Eq.~\ref{denoise} can be regard as the alpha matte $\tilde{\alpha}^0_{t}$ that predicted in one step directly.

\begin{table*}[!h]
     \caption{Quantitative evaluation on portrait matting with state-of-the-art methods on P3M-10K dataset. ``Trimap$^*$'' denotes the GT trimap. The lower the better for all metrics, and the best results are marked in \textbf{Bold}.}
     \vspace{-5mm}
     \begin{center}
     \setlength{\tabcolsep}{0.275cm}{
     \begin{tabular}{c|c|c|c|c|c|c|c|c|c}
         \toprule
         \multirow{2}{*}{\diagbox{Methods}{Metrics}}   &\multirow{2}{*}{Guidance}   &\multicolumn{4}{c|}{P3M-500-P}      &\multicolumn{4}{c}{P3M-500-NP}\\
         \cline{3-10}   & &SAD$\downarrow$    &MSE$\downarrow$     &Grad$\downarrow$ &Conn$\downarrow$  &SAD$\downarrow$ &MSE$\downarrow$  &Grad$\downarrow$ &Conn$\downarrow$\\
         \hline

        \rowcolor{gray!20}LF~\cite{zhang2019late}    &None     & 42.95  &0.0191   &42.19  &18.80 &32.59  &0.0131    &31.93 &19.50\\
        HATT~\cite{qiao2020attention}  &None      & 25.99  &0.0054  &14.91  &25.29 &30.53  &0.0072   &19.88 &27.42\\
        \rowcolor{gray!20}SHM~\cite{chen2018semantic}    &None    & 21.56  &0.0100   &21.24  &17.53 &20.77  &0.0093    &20.30 &17.09\\
        GFM~\cite{li2020end}     &None    & 13.20  &0.0050   &12.58  &17.75 &15.50  &0.0056    &14.82 &18.03\\
        \rowcolor{gray!20}P3M-Net~\cite{li2021privacy}  &None   & 8.73   &0.0026   &8.22   &13.88 &11.23  &0.0035    &10.35 &12.51\\
        MODNet~\cite{MODNet}    &None                &13.31    &0.0038         &16.50    &10.88    &16.70     &0.0051        &15.29  &13.81\\
        \hline
        \rowcolor{gray!20}DIM~\cite{xu2017deep} &Trimap$^*$  &4.89   &0.0009  &\textbf{4.48}   &9.68  &5.32   &0.0009    &\textbf{4.70}  &7.70\\
        AlphaGAN~\cite{lutz2018alphagan}  &Trimap               &5.27     &0.0112          &- &- &5.24 &0.0112 &- &-  \\
        \rowcolor{gray!20}GCA~\cite{li2020natural}       &Trimap               &4.36     &0.0088          &10.04 &5.03 &\textbf{4.35} &0.0089 &8.27 &5.26  \\
        IndexNet~\cite{lu2019indices}  &Trimap               &4.69    &\textbf{0.0007}         &8.94 &4.20 &5.36 &0.0007&7.19 &4.71  \\
        \rowcolor{gray!20}MG~\cite{yu2021mask}        &Mask               &5.60            &0.0012  &11.27 &5.16 &6.23 &0.0012 &9.54 &5.59 \\
        MatteFormer~\cite{park2022matteformer}       &Trimap               &\textbf{4.54}             &0.0008  &9.32 &4.08 &{4.91} &\textbf{0.0007} &7.19 &4.27 \\
        \hline
        \rowcolor{gray!20}{DiffusionMat} &Trimap & 4.58   &\textbf{0.0007}  &{8.67}  &\textbf{3.89}  &5.03   &\textbf{0.0007}   &{6.29}  &\textbf{4.15}\\
        \bottomrule
     \end{tabular}
     }
     \end{center}
     \vspace{-0.8cm}
     \label{table:p3m}
\end{table*}

\subsubsection{Loss Functions}

In this section, we provide the loss functions for training~DiffusionMat, the first one is the DDIM inversion loss $\mathcal{L}_{\mathrm{Inv}}$ for correcting the intermediate denoised results. Let $\{\alpha_T, ...,\alpha_t,... \alpha_0\}$ denotes the deterministic inversion trajectory of GT alpha matte $\alpha$, in each timestep $t$, we define the DDIM inversion loss $\mathcal{L}_{\mathrm{Inv}}$ as:
\begin{equation}
\mathcal{L}_{\mathrm{Inv}}=\| \alpha_{t}- {m}^{ARP}_{t}  \|^2_2.
\end{equation}

Moreover, we also develop another loss term $\mathcal{L}_{\mathrm{\alpha}}$ by aligning the distance between one step alpha matte $\tilde\alpha_t^0$ with GT alpha matte $\alpha$:
\begin{equation}
\mathcal{L}_{\mathrm{\alpha}}=\| \alpha-\tilde\alpha^0_t \|_1.
\end{equation}

For those matting datasets that have GT foreground and background, we also also introduce the composition loss that wild used in many image matting works~\cite{li2020natural,yu2021mask,park2022matteformer,xu2017deep}. We first composite an image $I^{comp}_t$ using $\tilde\alpha_t^0$ following Eq.~\ref{eq:matting}, then we minimize the absolute difference between the composited image and the GT image $I$:
\begin{equation}
\mathcal{L}_{\mathrm{Comp}}=\| {I} -   I^{comp}_t \|_1.
\end{equation}

Now we define the final loss $\mathcal{L}_{\mathrm{Final}}$ for training the DiffusionMat:
\begin{equation}
\mathcal{L}_{\mathrm{Final}}= \lambda_1\mathcal{L}_{\mathrm{Inv}} + \lambda_2\mathcal{L}_{\mathrm{\alpha}} + \lambda_3\mathcal{L}_{\mathrm{Comp}},
\label{eqfinal}
\end{equation}
where $\{\lambda_i\}$ denote the weight factors for balancing loss terms. It is noticed that $\mathcal{L}_{\mathrm{Final}}$ can be applied on arbitrary timesteps.

Different from classical diffusion models trained and inference in two different pipelines, our DiffusionMat is trained in the inference process of diffusion. Once trained, we can predict the alpha matte with the same pipeline.



\section{Experiments}

\subsection{Experimental Settings}
\label{sec:exps}

\textbf{Datasets.} To evaluating the effectiveness of our~DiffusionMat, we conduct experiment on following image matting datasets. 
\begin{itemize}[leftmargin=*,nosep]
\item {P3M-10K} dataset~\cite{li2021privacy} is the largest privacy-preserving portrait matting dataset, it contains 10,421 high-resolution real-word portrait images and the corresponding alpha matte. The dataset is divided into one training set and two test datasets. The training set contains 9,421 face-blurred portrait-matte pairs. The first test set has 500 face-blurred portrait-matte pairs, which are denoted as {P3M-500-P}. Another test set also contains 500 portrait-matte pairs, but the portraits are not blurred faces, and it is denoted as {P3M-500-NP}.
\item {Human-2K} dataset~\cite{liu2021tripartite} provides 2,100 portrait foreground image and matte pairs. Then the foreground images are composited with background images from MS COCO~\cite{lin2014microsoft} and Pascal VOC~\cite{everingham2010pascal} datasets, which result in 43,100 training samples and 2,000 testing samples.
\item {Composition-1k} dataset~\cite{xu2017deep}. Beyond above portrait matting datasets, we also conduct experiment on Composition-1k dataset, which contains not only portrait but also object images. It contains 431,000 training image-matting pairs which composited with background from Pascal VOC~\cite{everingham2010pascal} datasets. The test dataset contains 1,000 composited images which composited with background image from MS COCO~\cite{lin2014microsoft} dataset. 
\end{itemize}

\textbf{Evaluation Metrics.}
We use four metrics to evaluate the alpha matting results, namely, the mean square error (MSE), the sum of absolute differences (SAD), the gradient error (Grad), and connectivity errors (Conn)~\cite{rhemann2009perceptually}. Particularly, MSE and SAD measure the statistical differences between predicted and GT alpha mattes, Conn evaluates the disconnected foreground objects, and Grad focuses on the over-smoothed or erroneous discontinuities in the alpha matte. Compared with MSE and SAD, Conn and Grad are more related to human perception.

\subsection{Implementation Details}
\label{sec:imp}

We implement the proposed framework in Pytorch on a PC with Nvidia GeForce RTX 3090. We first follow~\cite{ho2020denoising} that uses U-Net architecture~\cite{ronneberger2015u} as the diffusion model's structure. Since the resolution of alpha matte is relatively high, we remove the ``\texttt{att\_block}'' of U-Net to save GPU memory. For two portrait matting datasets, we use the diffusion model that trains on their mixed alpha mattes. The diffusion model fine-tuned on the Composition-1k dataset is utilized on this object matting dataset. We use the Swin-Unet as the image encoder~\cite{cao2022swin}, and the correction network also has the same U-Net structure as the diffusion model. The framework is optimized by the Adam optimizer~\cite{kingma2014adam} with the learning rate of $1e^{-4}$. We empirically set the balancing weights in Eq.~\ref{eqfinal} as $\lambda_{1}=2$, $\lambda_{2}=1$, and $\lambda_{3}=1$. We crop the input images and trimaps to the resolution of 512$\times$512 for training the framework. During the inference, we predict the alpha matte of the full-resolution images.

\begin{table}[t]
     \caption{Quantitative evaluation on portrait matting with state-of-the-art methods on Human-2K dataset. ``Trimap$^*$'' denotes the GT trimap. The lower the better for all metrics. The best results are marked in \textbf{Bold}.}
     \vspace{-5mm}
     \begin{center}
     \setlength{\tabcolsep}{0.08cm}{
     \begin{tabular}{c|c|c|c|c|c}
        \toprule
        Methods & Guidance &SAD$\downarrow$         &MSE$\downarrow$  &Grad$\downarrow$ &Conn$\downarrow$  \\
        \hline
        DIM~\cite{xu2017deep}       &Trimap$^*$ &7.53  &0.008  &6.40  &6.70 \\
        \rowcolor{gray!20}IndexNet~\cite{lu2019indices}   &Trimap  &6.55  &0.006  &4.50  &5.50 \\
        GCA~\cite{li2020natural}      &Trimap    &5.18  &0.004  &3.00  &4.00 \\
        \rowcolor{gray!20}TIMI-Net~\cite{liu2021tripartite} &Trimap &4.20  &0.0026 &2.06 &2.95 \\
        MODNet~\cite{MODNet}    &None   &7.80 &0.0080 &7.20 &7.40 \\
        \rowcolor{gray!20}MG~\cite{yu2021mask} &Mask &4.40 &0.0040 &2.50 &3.20 \\
        SPGM~\cite{xu2022situational}    &Mask     &\textbf{4.00} &\textbf{0.0020} &2.00 &2.80 \\
        \hline
        \rowcolor{gray!20}DiffusionMat  &Trimap      &4.04  &\textbf{0.0020} &\textbf{1.66} &\textbf{2.66}  \\
        \bottomrule
     \end{tabular}
     }
     \end{center}
     \vspace{-0.8cm}
     \label{table:h2k}
\end{table}




\subsection{Evaluations}

\begin{figure*}[!t]
    \centering
    \captionsetup[subfloat]{labelformat=empty,justification=centering}
    \subfloat[Image]{
     \begin{minipage}{0.095\linewidth}
     \includegraphics[width=\linewidth]{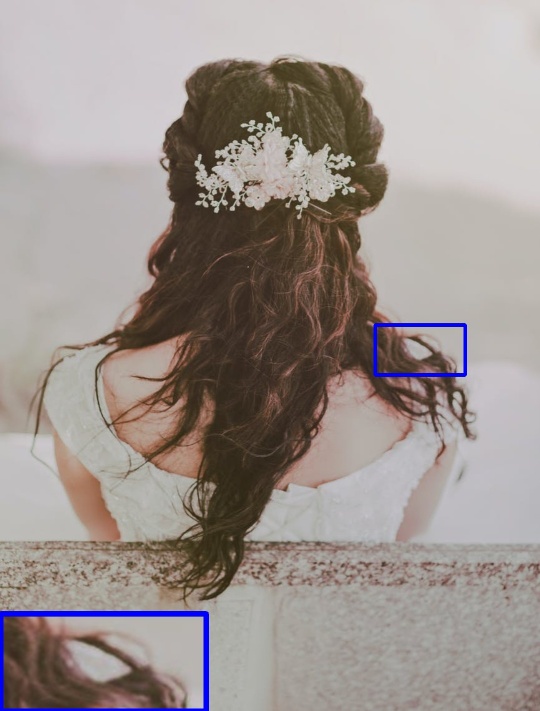}
     \includegraphics[width=\linewidth]{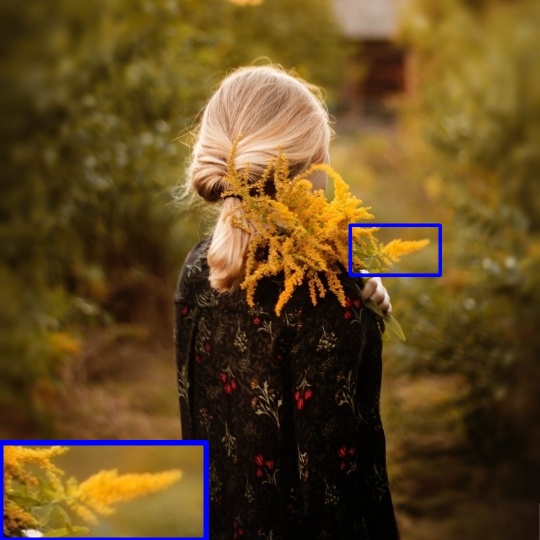}
     \includegraphics[width=\linewidth]{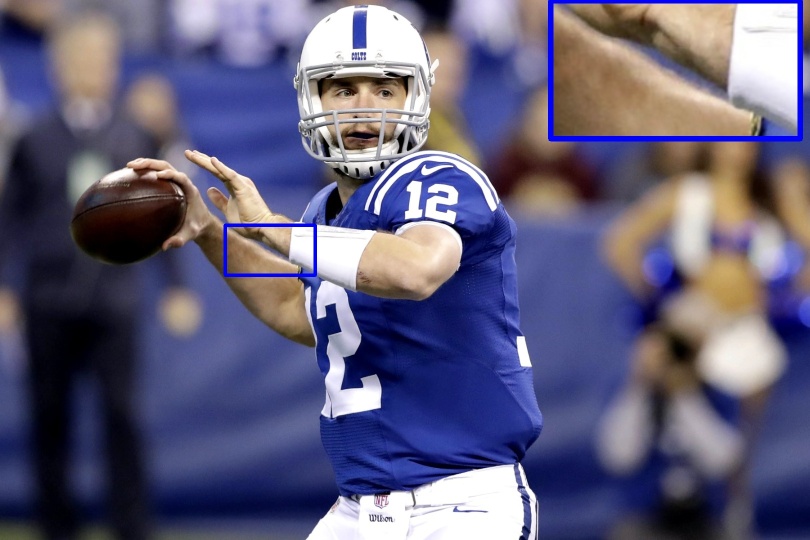}
     \includegraphics[width=\linewidth]{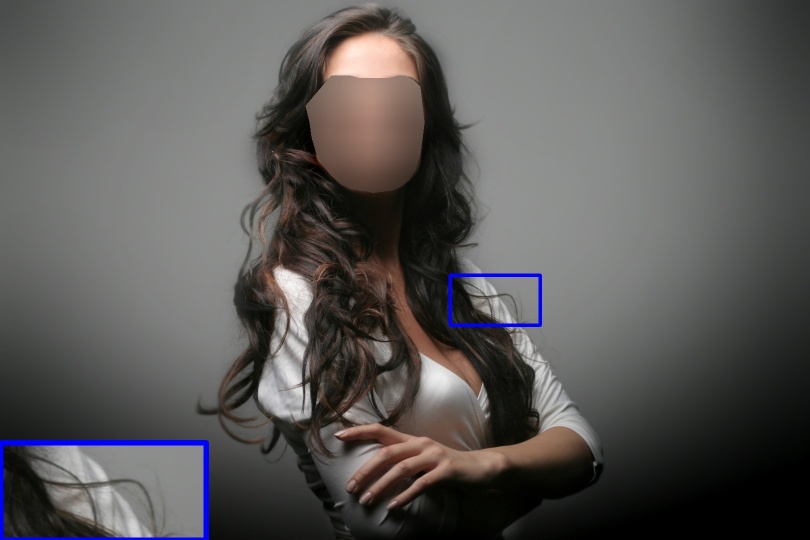}
     \includegraphics[width=\linewidth]{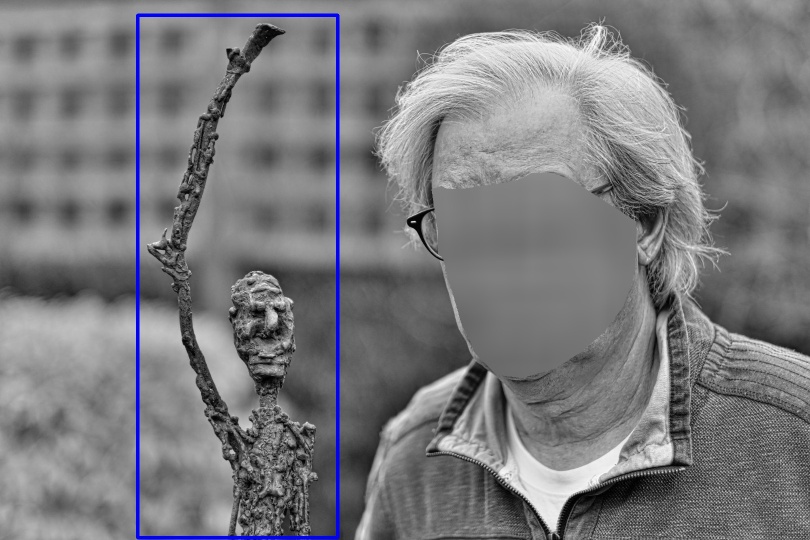}
     \includegraphics[width=\linewidth]{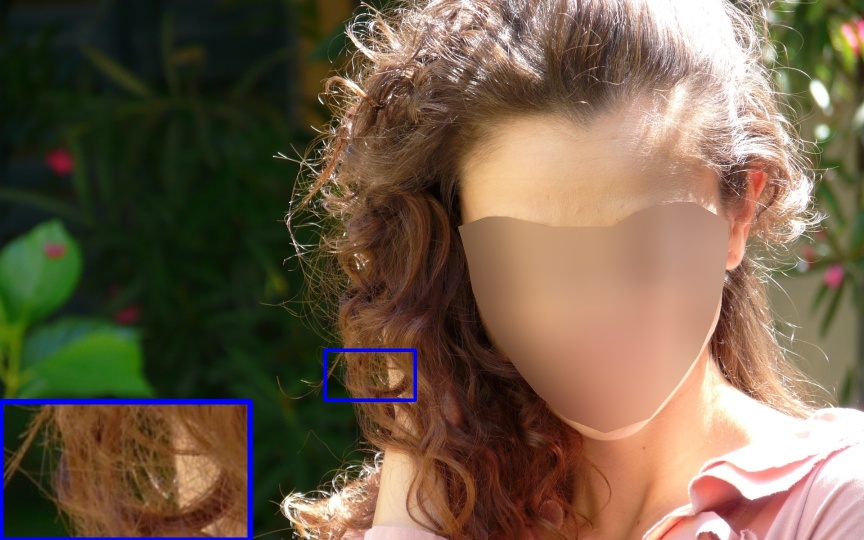}
     \end{minipage}
     }
    \hspace{-2.5mm}
    \subfloat[Trimap]{
     \begin{minipage}{0.095\linewidth}
     \includegraphics[width=\linewidth]{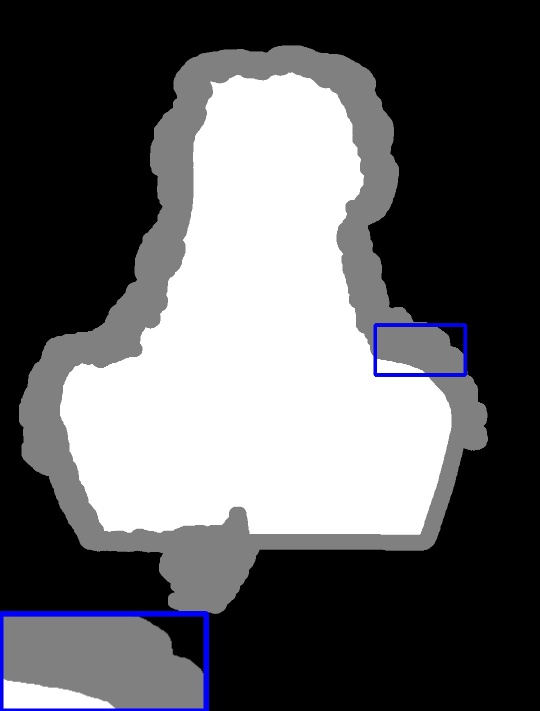}
     \includegraphics[width=\linewidth]{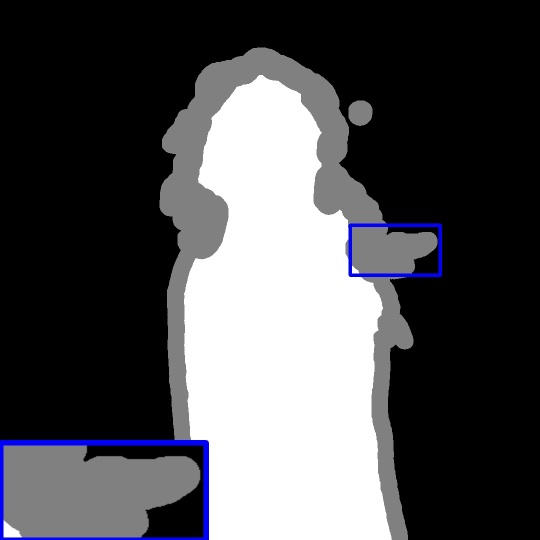}
     \includegraphics[width=\linewidth]{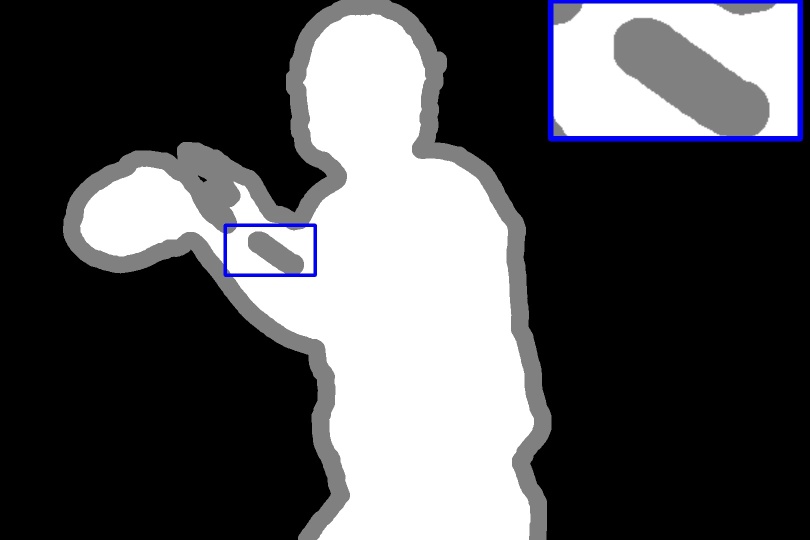}
     \includegraphics[width=\linewidth]{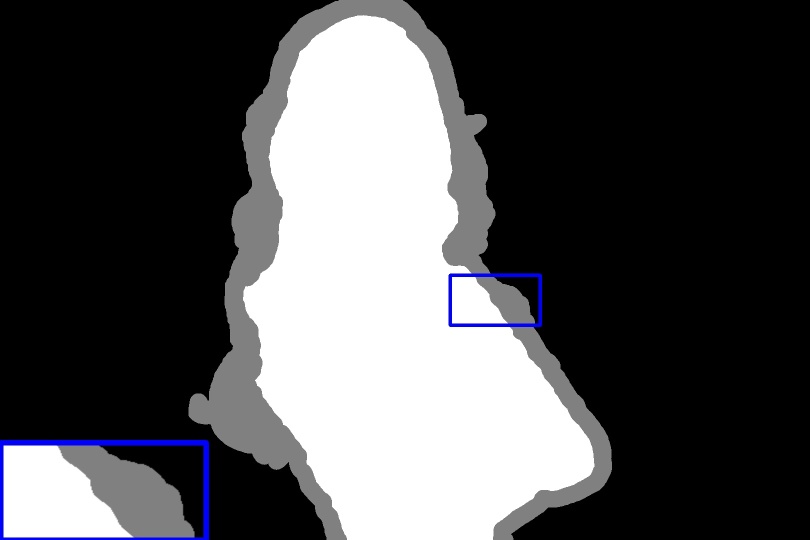}
     \includegraphics[width=\linewidth]{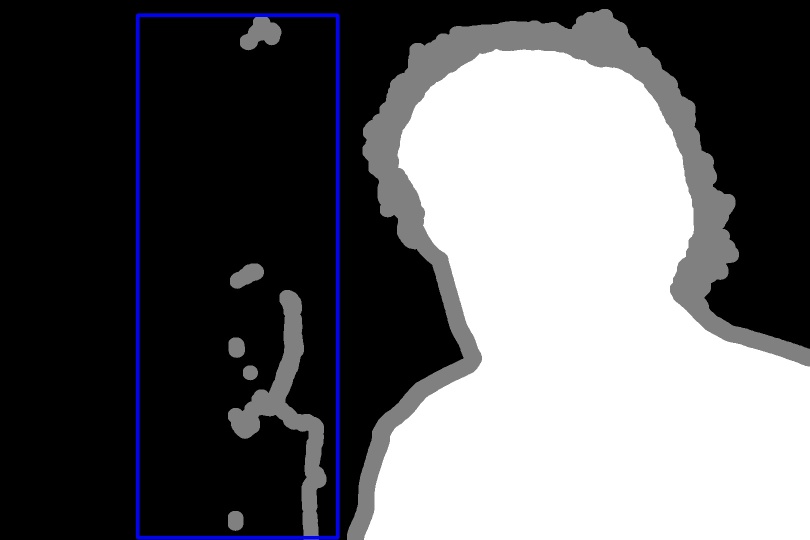}
     \includegraphics[width=\linewidth]{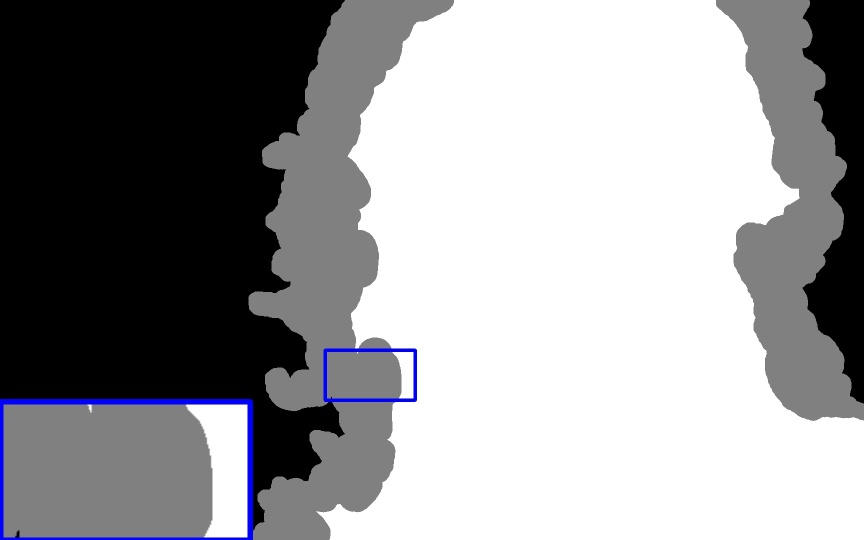}
     \end{minipage}
     }
    \hspace{-2.5mm}
    \subfloat[MODNet~\cite{MODNet}]{
     \begin{minipage}{0.095\linewidth}
     \includegraphics[width=\linewidth]{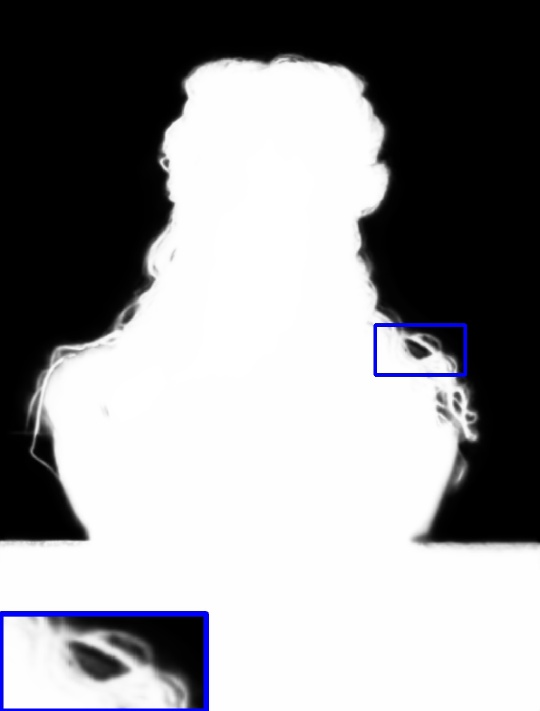}
     \includegraphics[width=\linewidth]{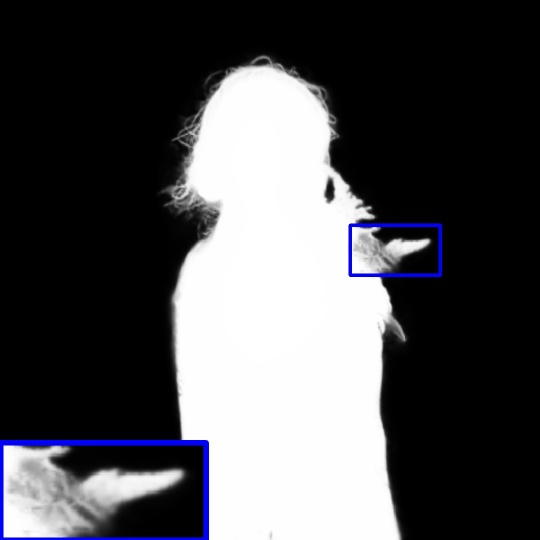}
     \includegraphics[width=\linewidth]{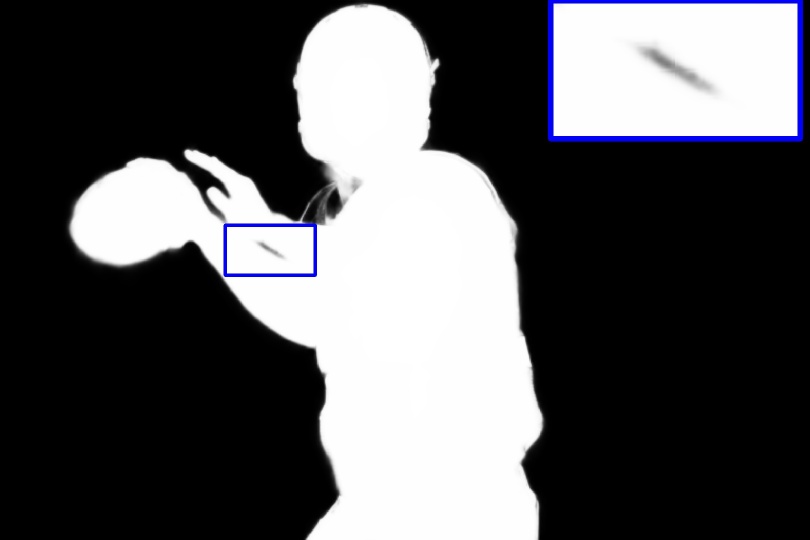}
     \includegraphics[width=\linewidth]{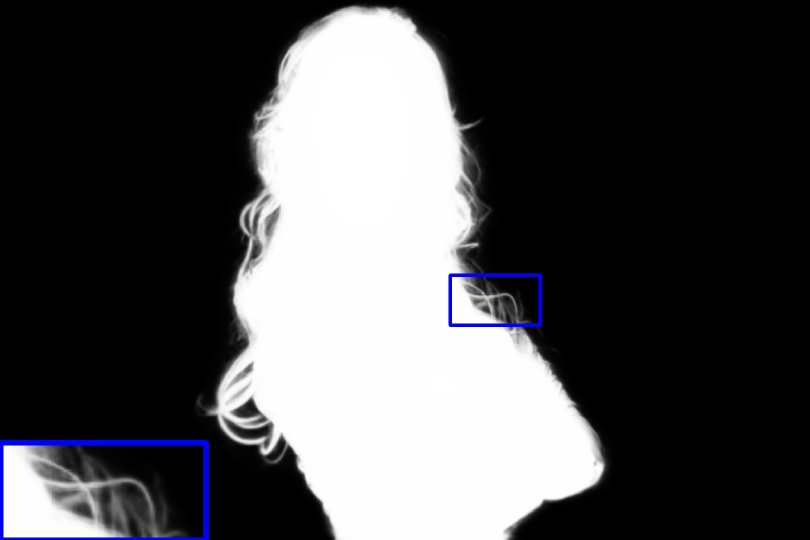}
     \includegraphics[width=\linewidth]{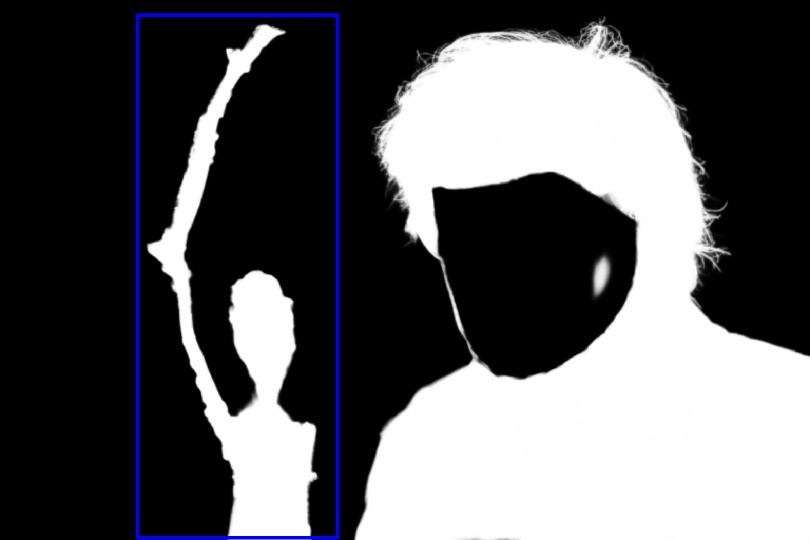}
     \includegraphics[width=\linewidth]{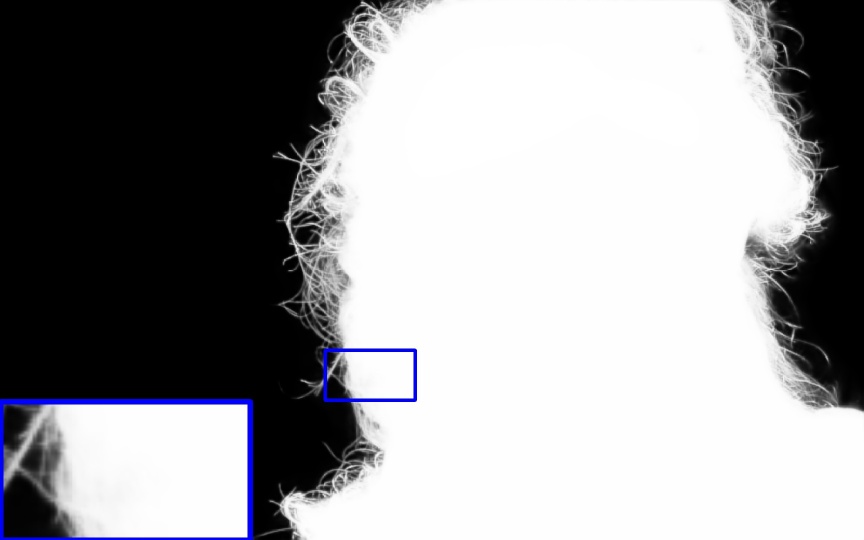}
     \end{minipage}
     }
    \hspace{-2.5mm}
    \subfloat[P3M-Net~\cite{li2021privacy}]{
     \begin{minipage}{0.095\linewidth}
     \includegraphics[width=\linewidth]{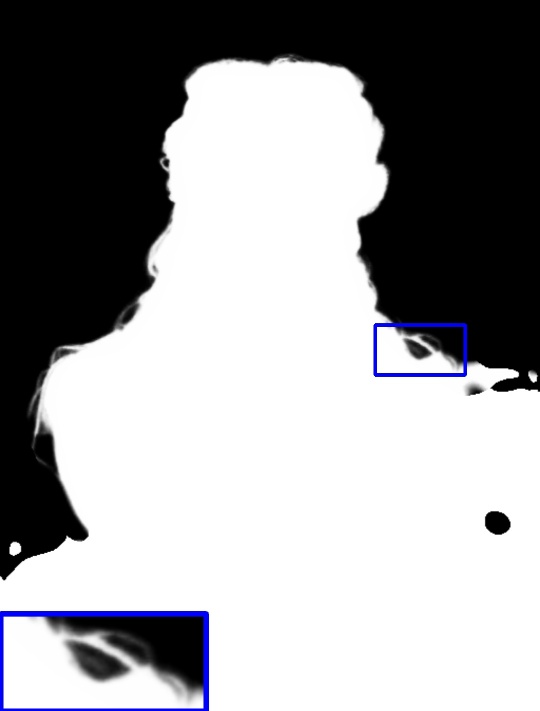}
     \includegraphics[width=\linewidth]{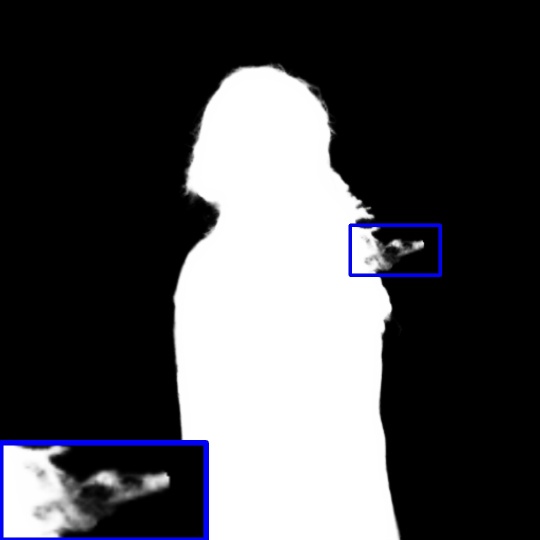}
     \includegraphics[width=\linewidth]{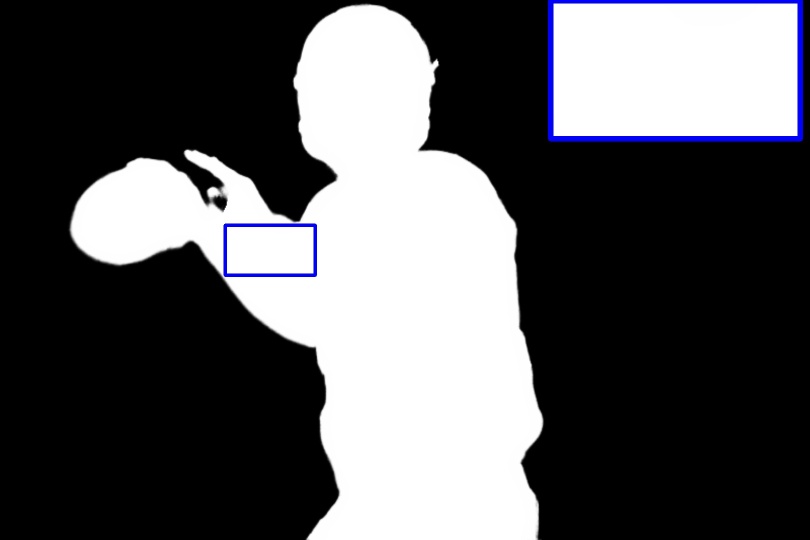}
     \includegraphics[width=\linewidth]{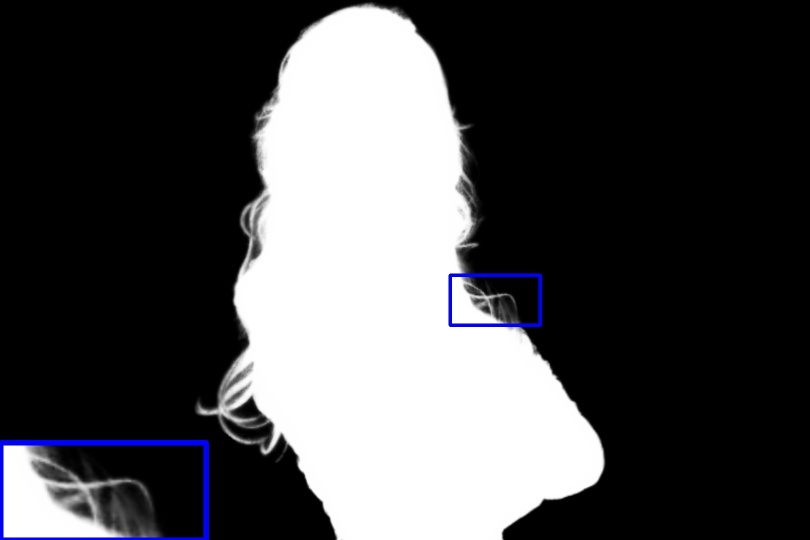}
     \includegraphics[width=\linewidth]{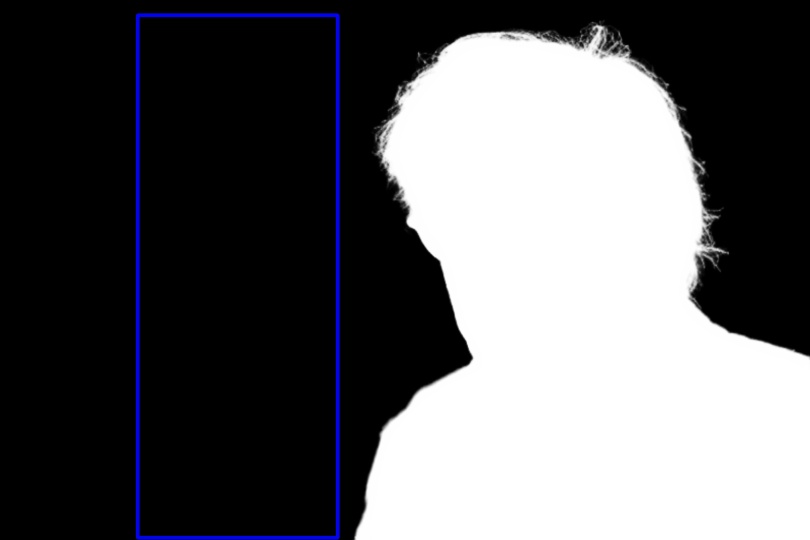}
     \includegraphics[width=\linewidth]{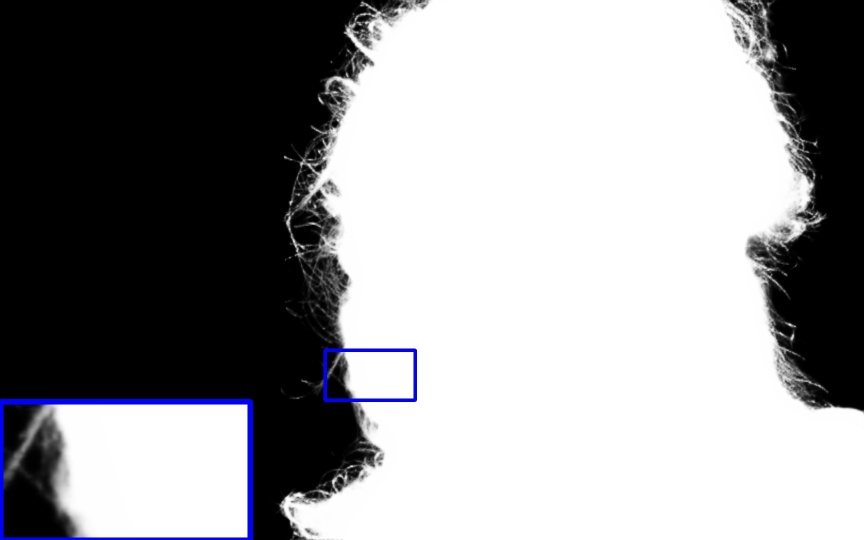}
     \end{minipage}
     }
    \hspace{-2.5mm}
    \subfloat[IndexNet~\cite{lu2019indices}]{
     \begin{minipage}{0.095\linewidth}
     \includegraphics[width=\linewidth]{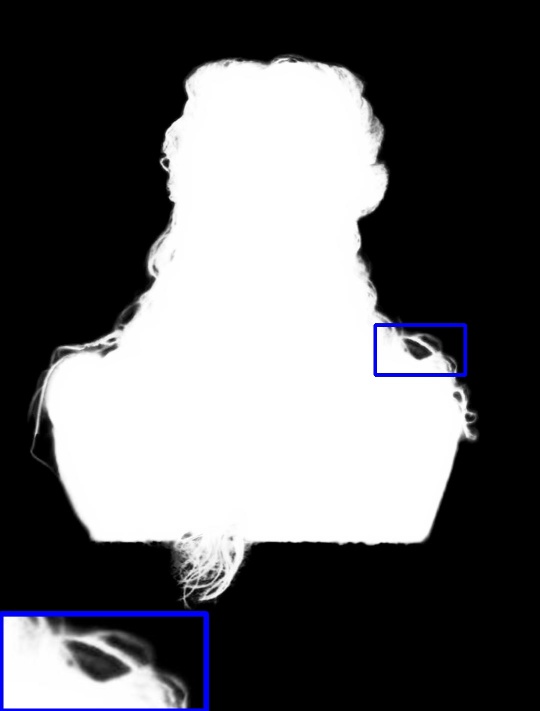}
     \includegraphics[width=\linewidth]{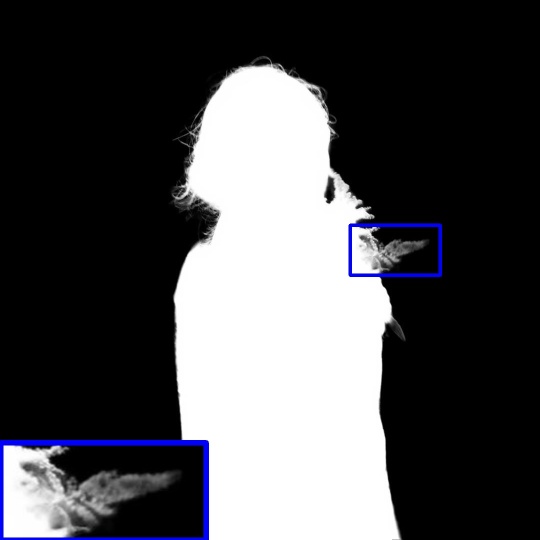}
     \includegraphics[width=\linewidth]{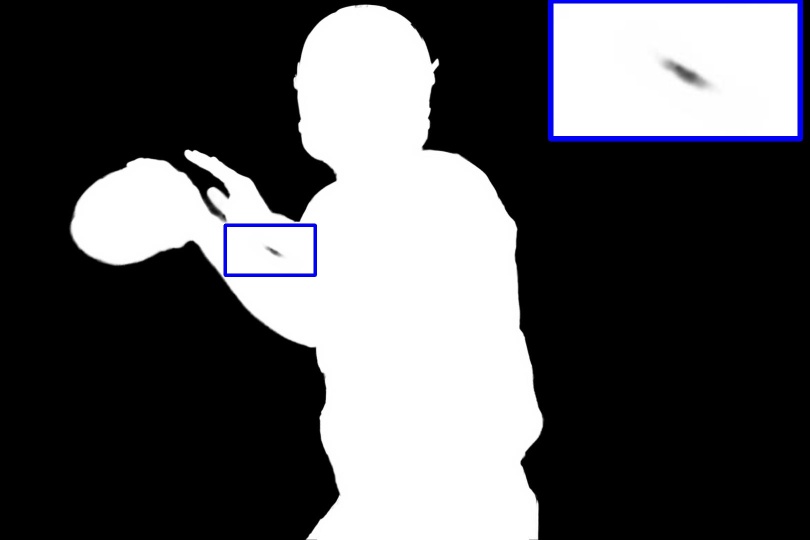}
     \includegraphics[width=\linewidth]{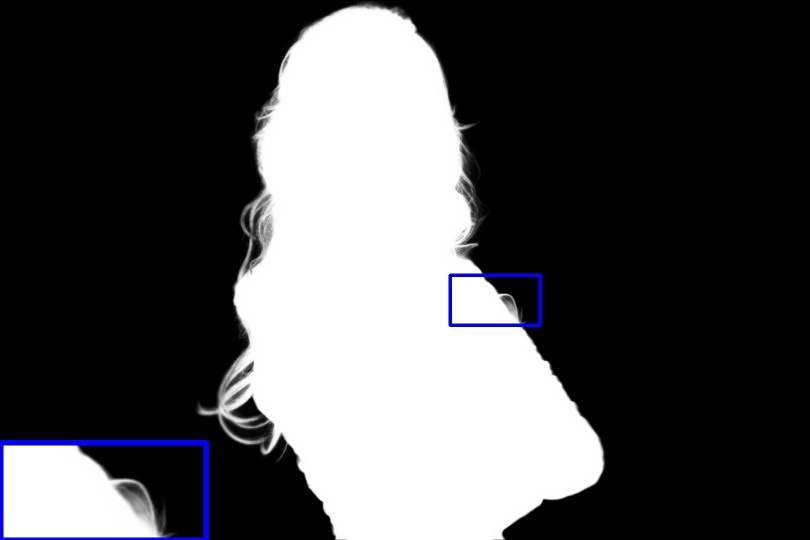}
     \includegraphics[width=\linewidth]{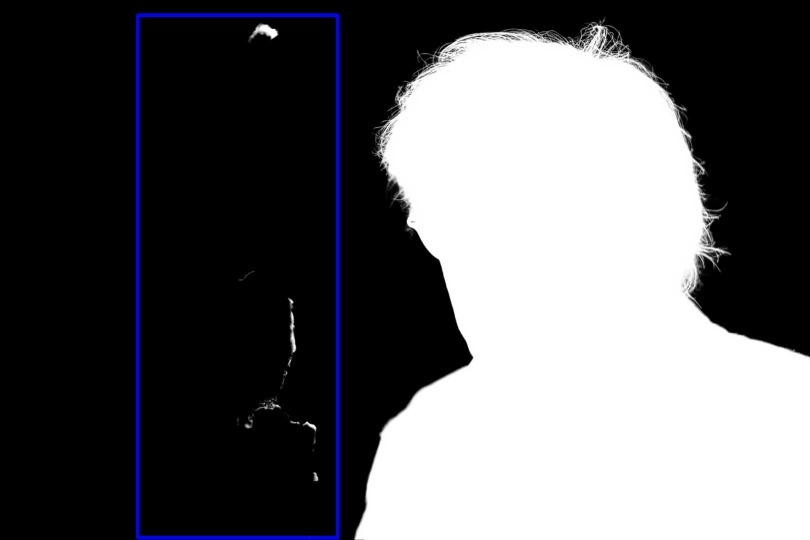}
     \includegraphics[width=\linewidth]{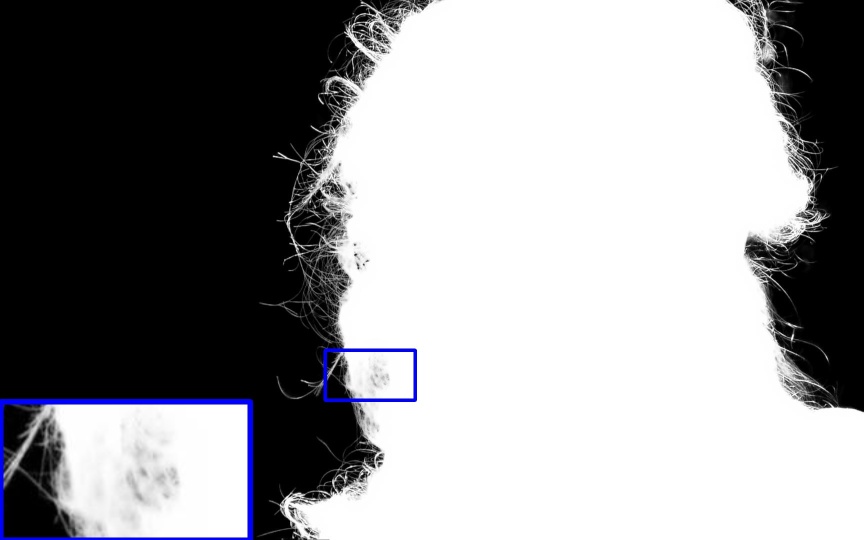}
     \end{minipage}
     }
    \hspace{-2.5mm}
    \subfloat[GCA~\cite{li2020natural} ]{
     \begin{minipage}{0.095\linewidth}
     \includegraphics[width=\linewidth]{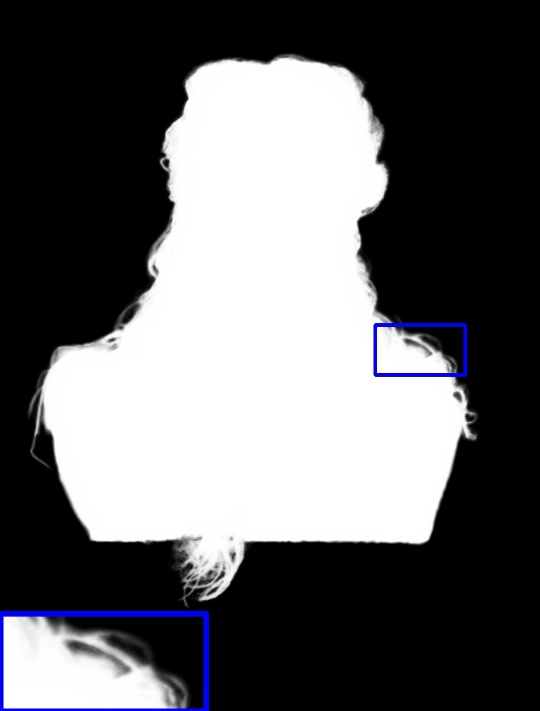}
     \includegraphics[width=\linewidth]{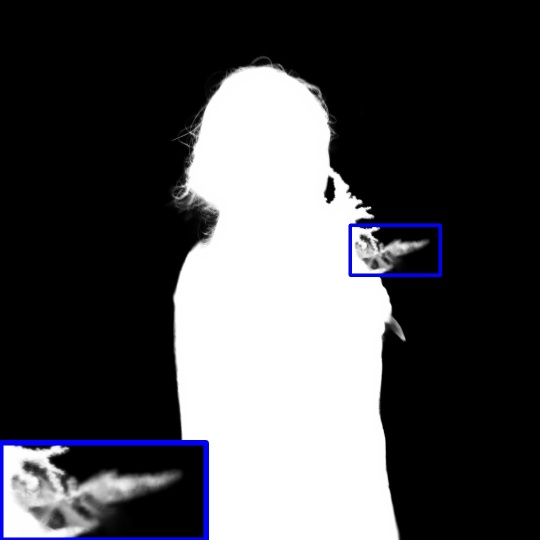}
     \includegraphics[width=\linewidth]{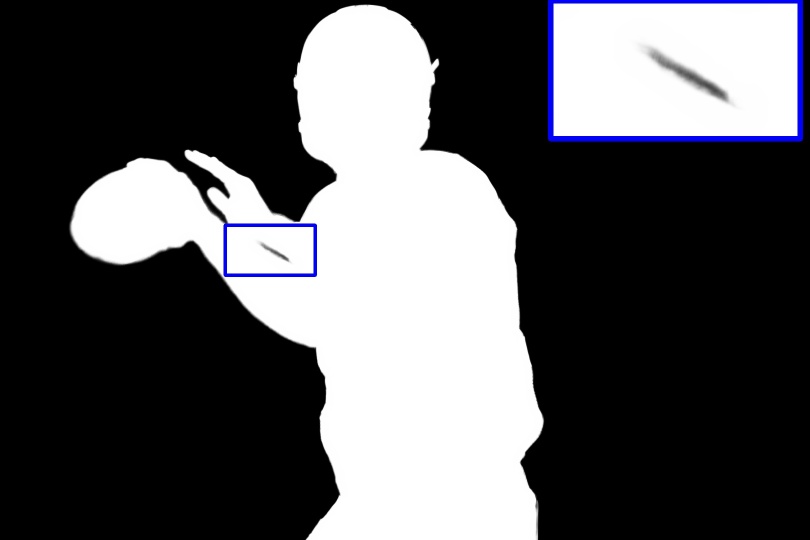}
     \includegraphics[width=\linewidth]{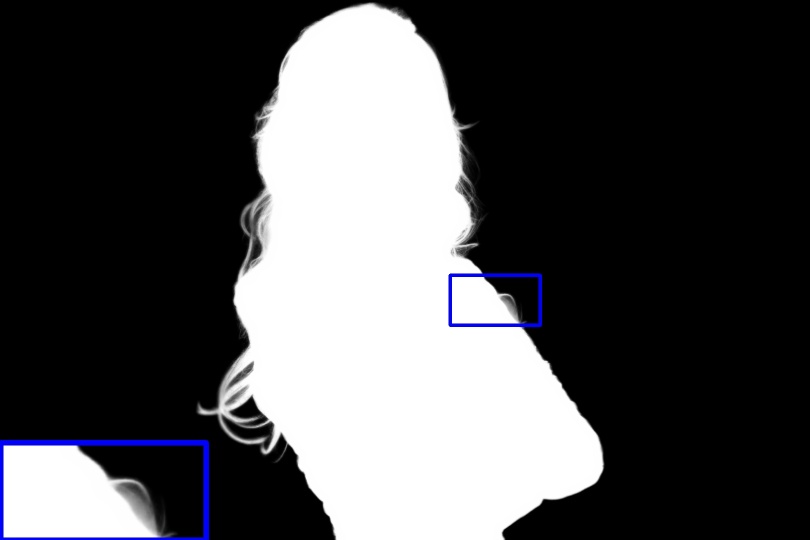}
     \includegraphics[width=\linewidth]{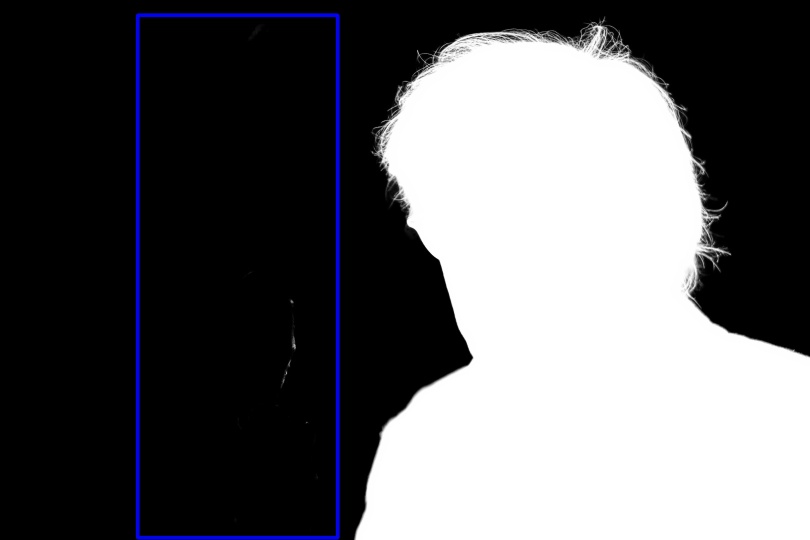}
     \includegraphics[width=\linewidth]{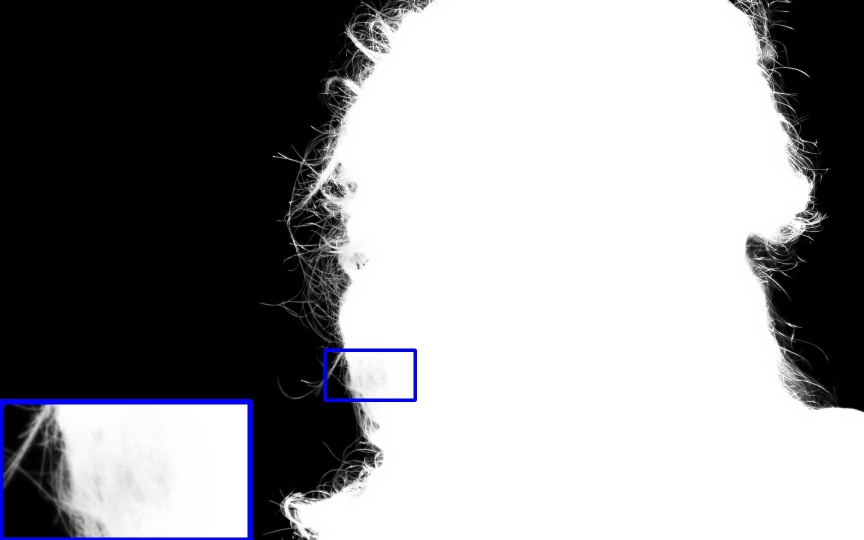}
     \end{minipage}
     }
    \hspace{-2.5mm}
    \subfloat[MG~\cite{yu2021mask}]{
     \begin{minipage}{0.095\linewidth}
     \includegraphics[width=\linewidth]{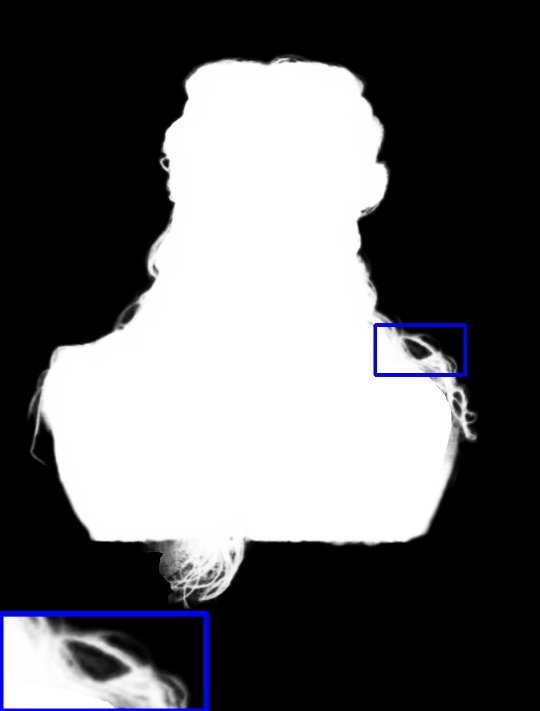}
     \includegraphics[width=\linewidth]{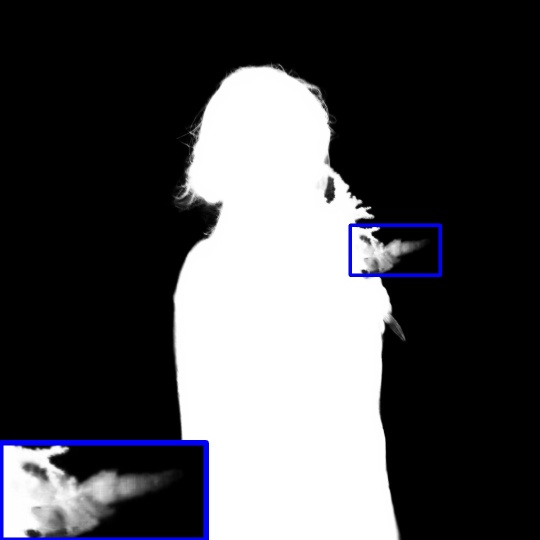}
     \includegraphics[width=\linewidth]{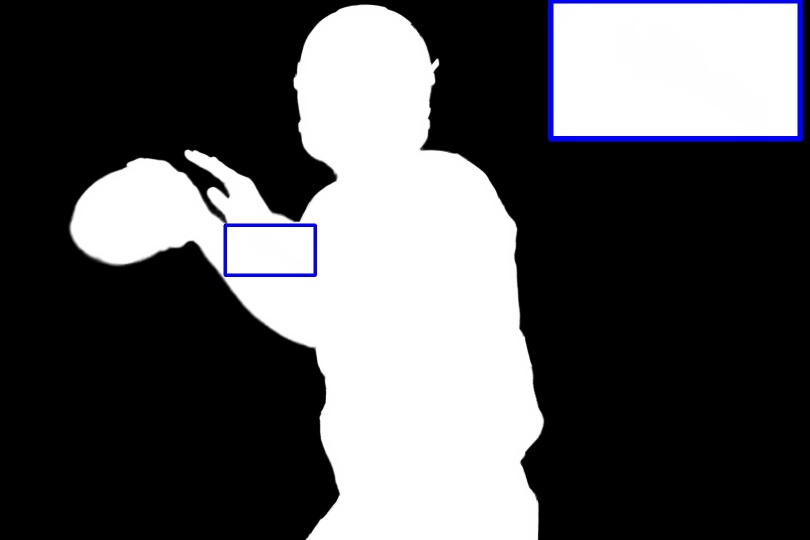}
     \includegraphics[width=\linewidth]{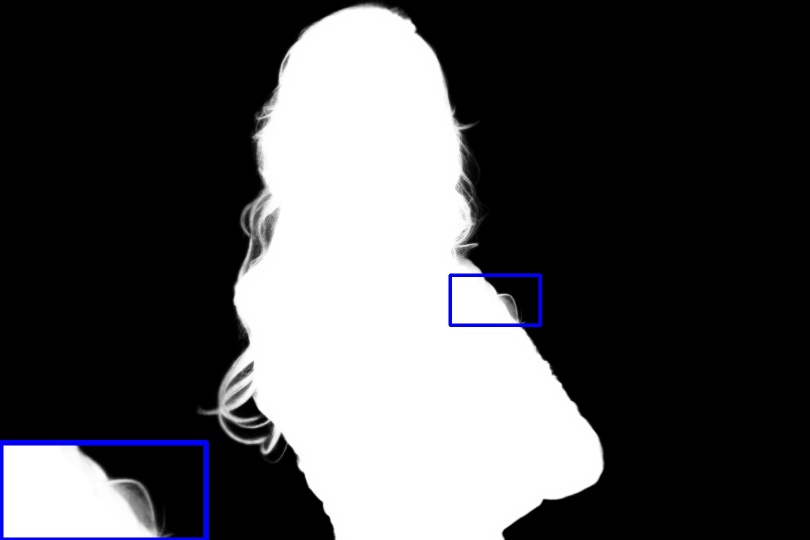}
     \includegraphics[width=\linewidth]{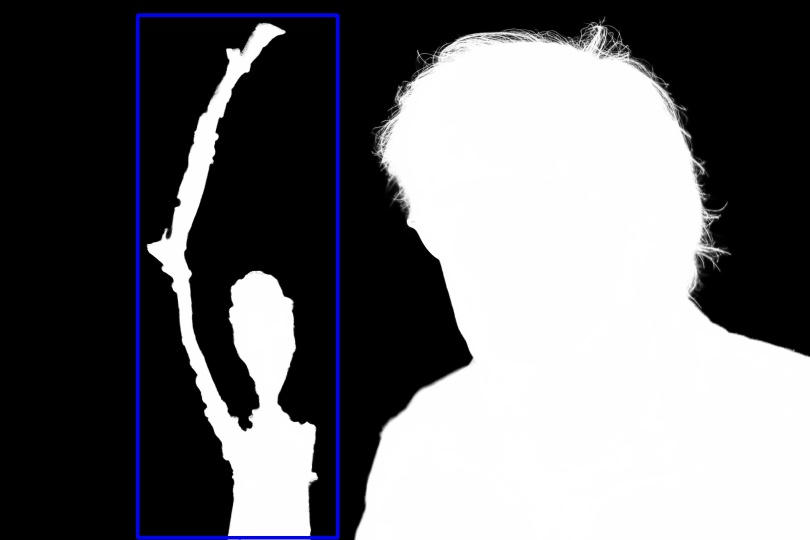}
     \includegraphics[width=\linewidth]{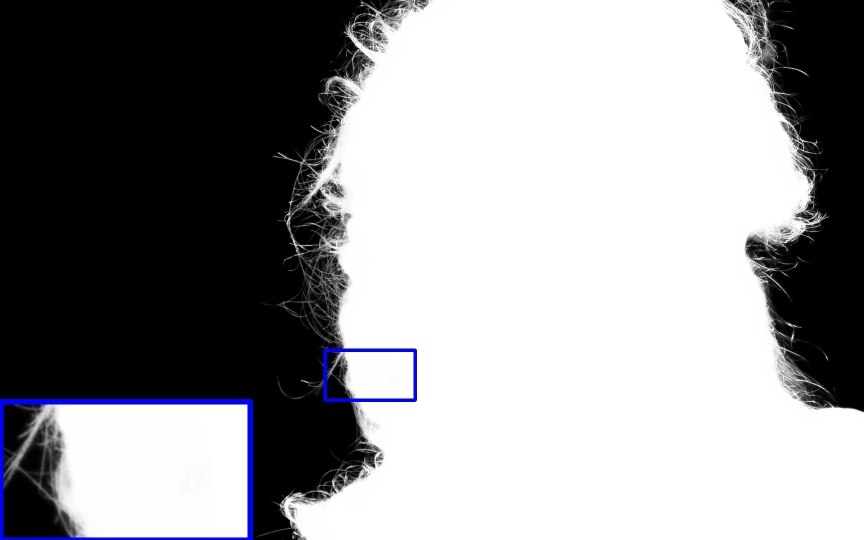}
     \end{minipage}
     }
    \hspace{-2.5mm}
    \subfloat[MFormer~\cite{park2022matteformer}]{
     \begin{minipage}{0.095\linewidth}
     \includegraphics[width=\linewidth]{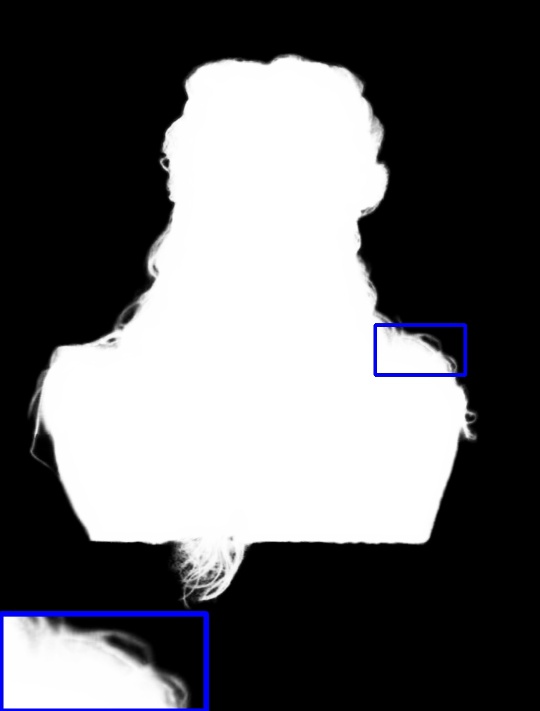}
     \includegraphics[width=\linewidth]{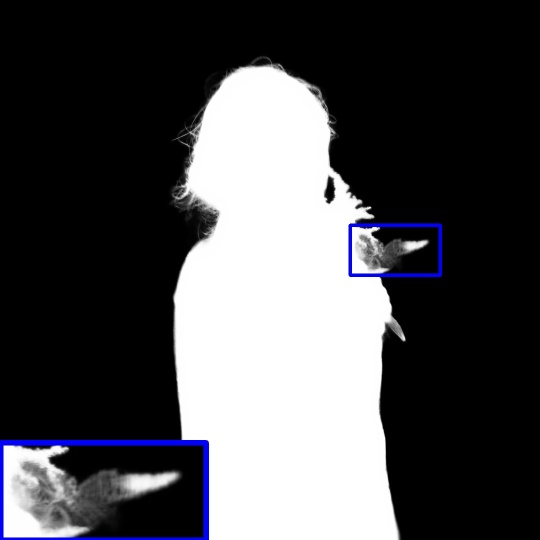}
     \includegraphics[width=\linewidth]{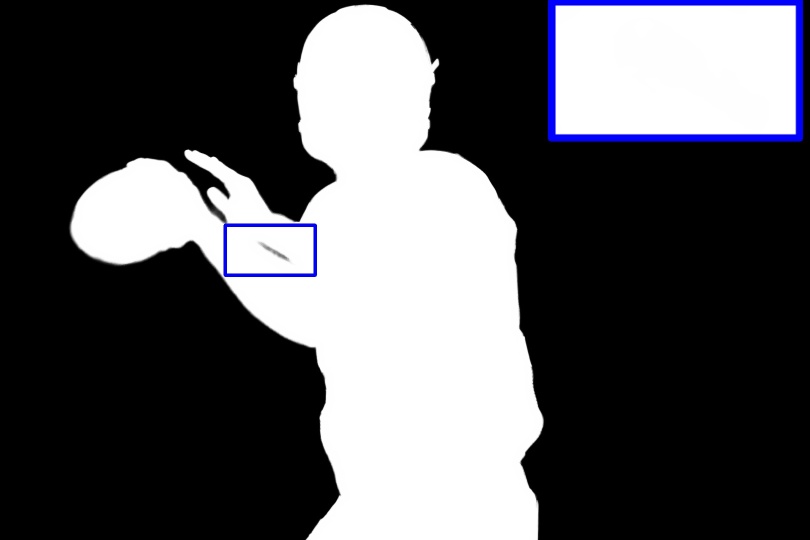}
     \includegraphics[width=\linewidth]{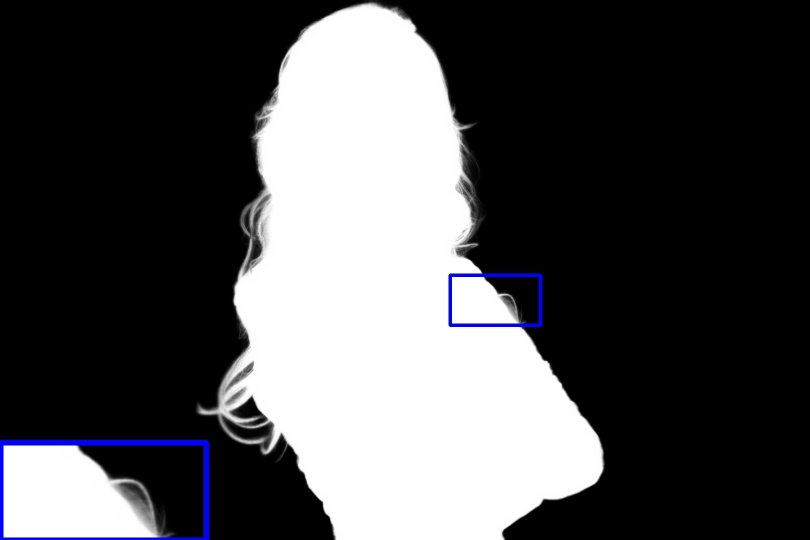}
     \includegraphics[width=\linewidth]{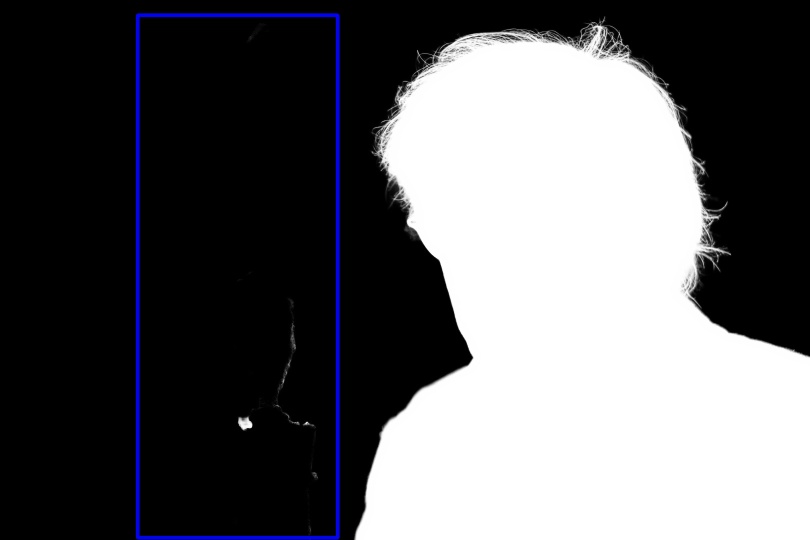}
     \includegraphics[width=\linewidth]{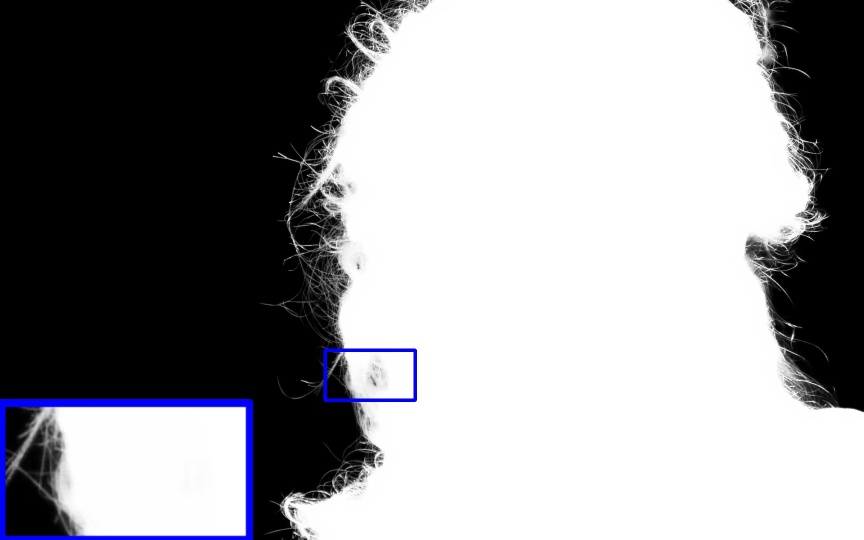}
     \end{minipage}
     }
    \hspace{-2.5mm}
    \subfloat[DiffusionMat]{
     \begin{minipage}{0.095\linewidth}
     \includegraphics[width=\linewidth]{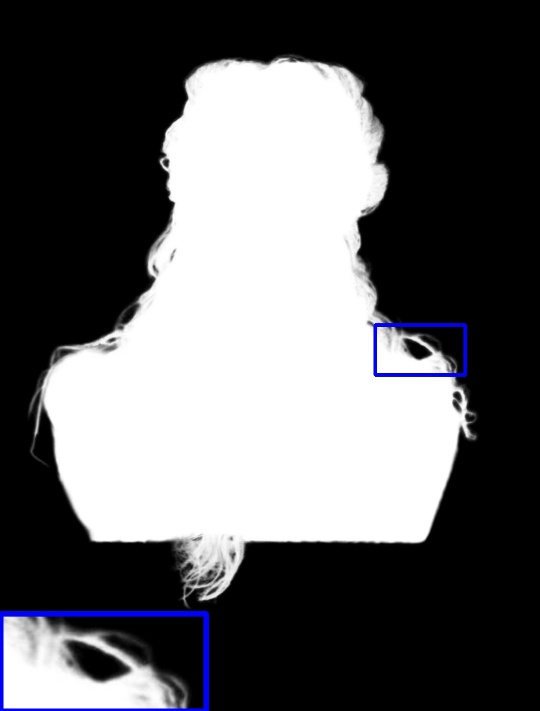}
     \includegraphics[width=\linewidth]{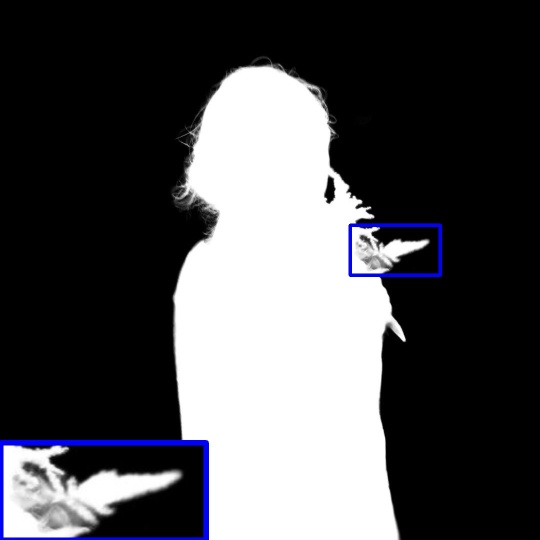}
     \includegraphics[width=\linewidth]{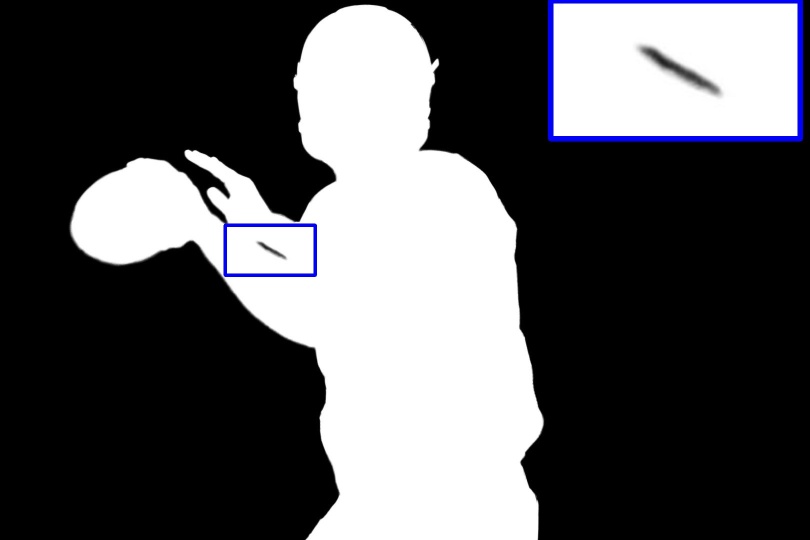}
     \includegraphics[width=\linewidth]{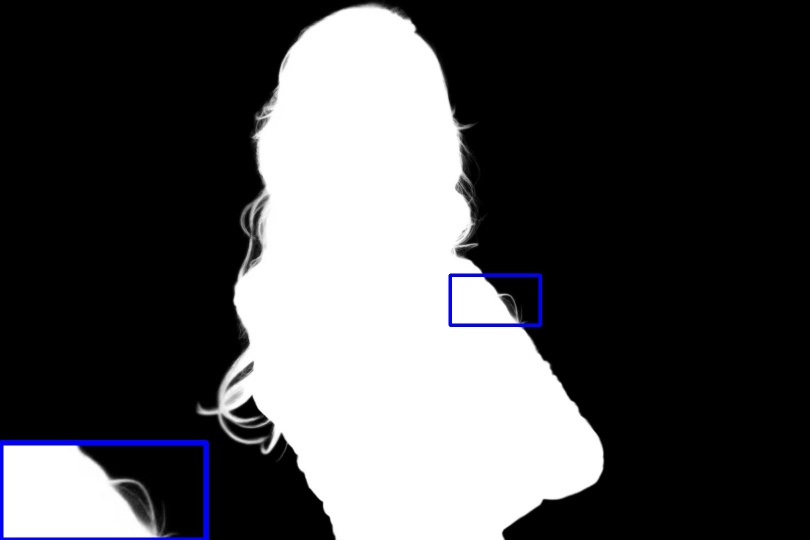}
     \includegraphics[width=\linewidth]{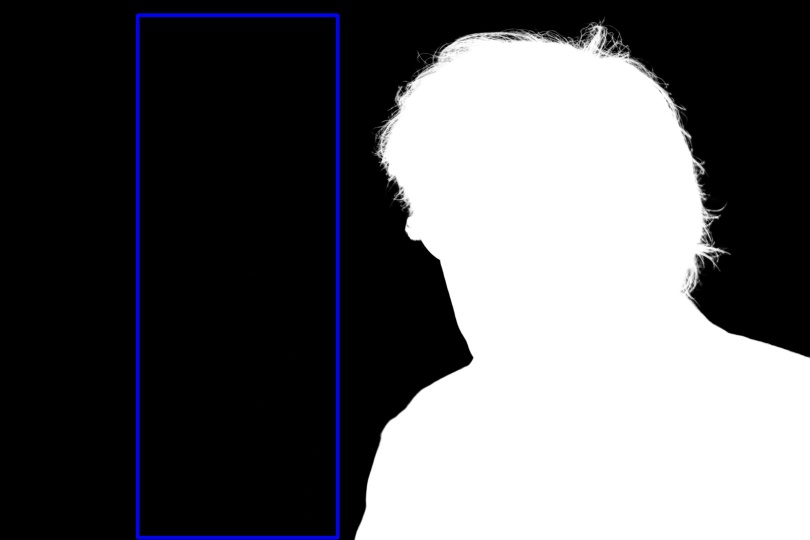}
     \includegraphics[width=\linewidth]{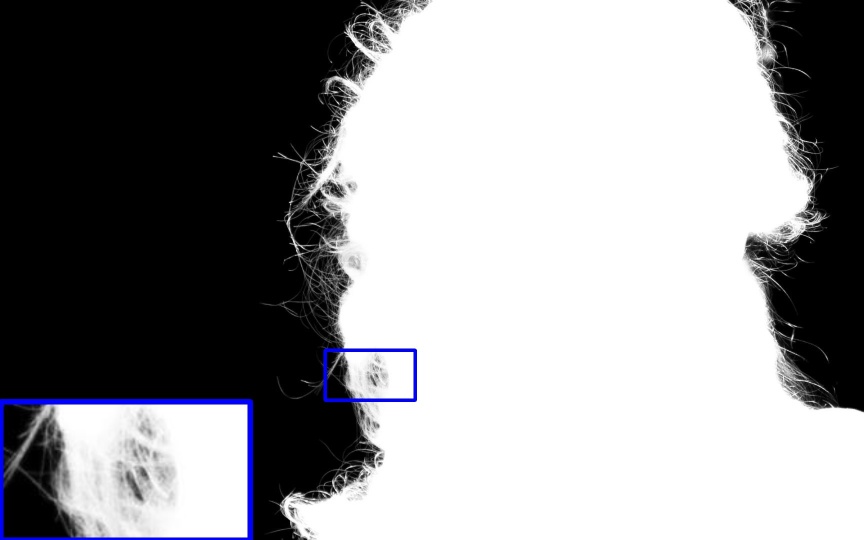}
     \end{minipage}
     }
    \hspace{-2.5mm}
    \subfloat[GT]{
     \begin{minipage}{0.095\linewidth}
     \includegraphics[width=\linewidth]{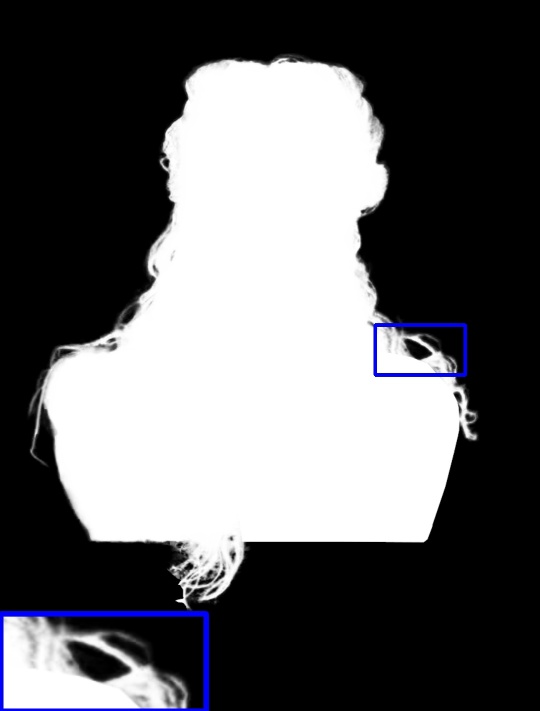}
     \includegraphics[width=\linewidth]{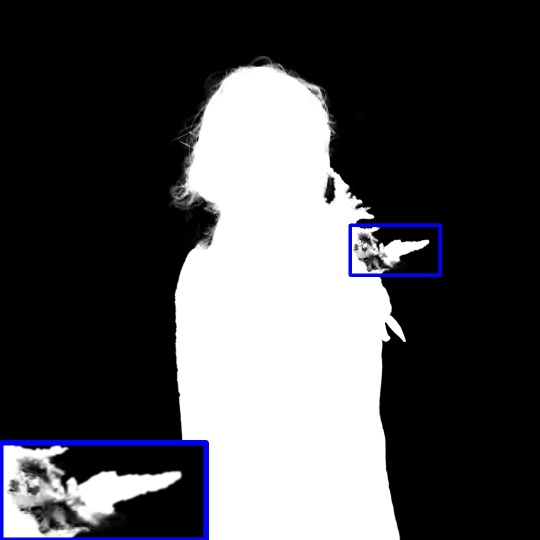}
     \includegraphics[width=\linewidth]{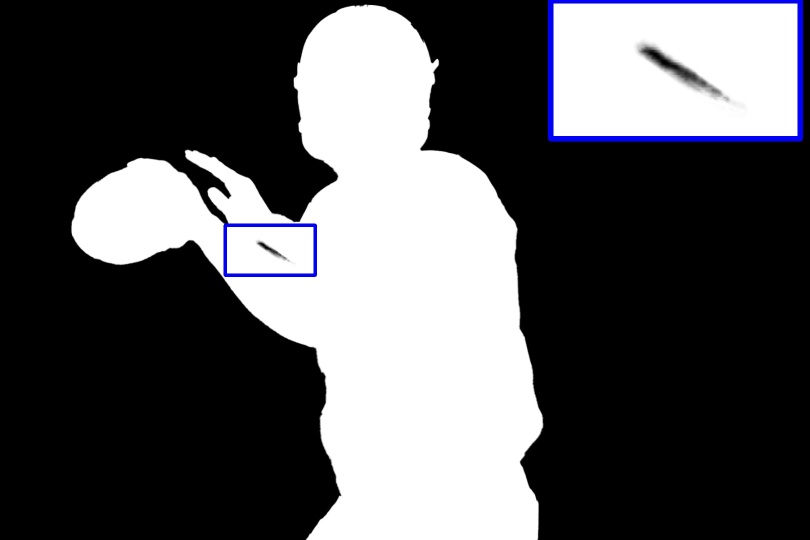}
     \includegraphics[width=\linewidth]{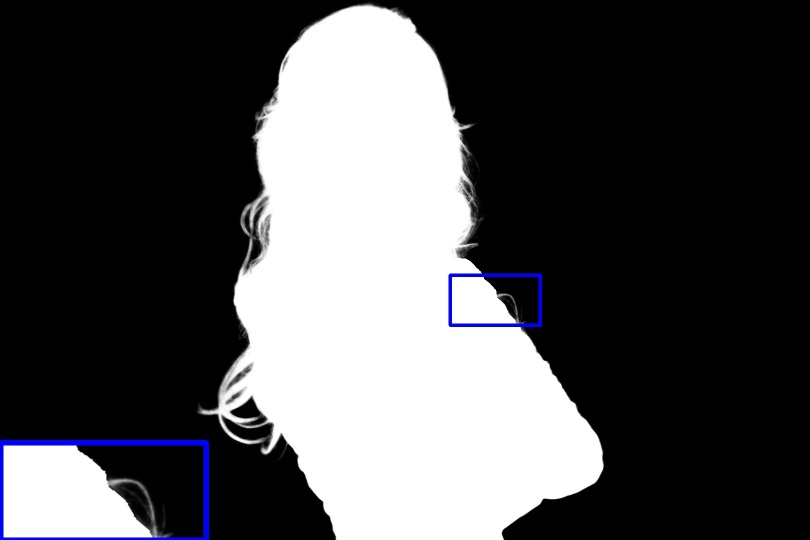}
     \includegraphics[width=\linewidth]{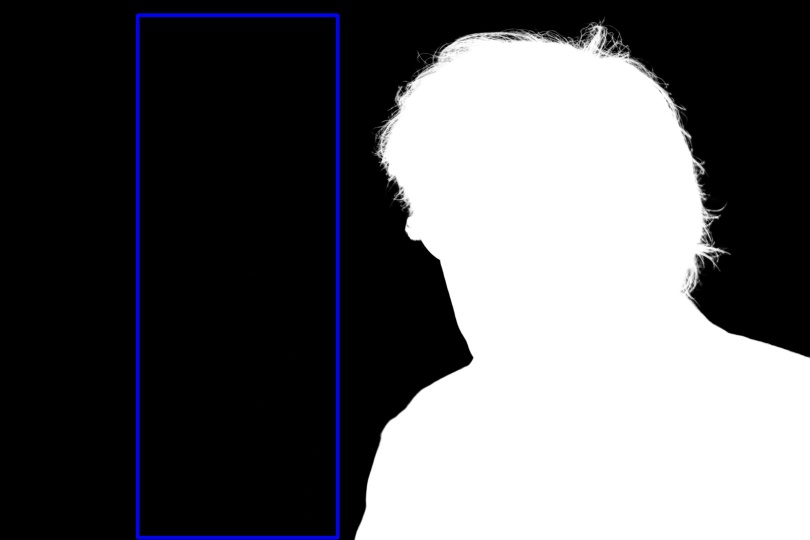}
     \includegraphics[width=\linewidth]{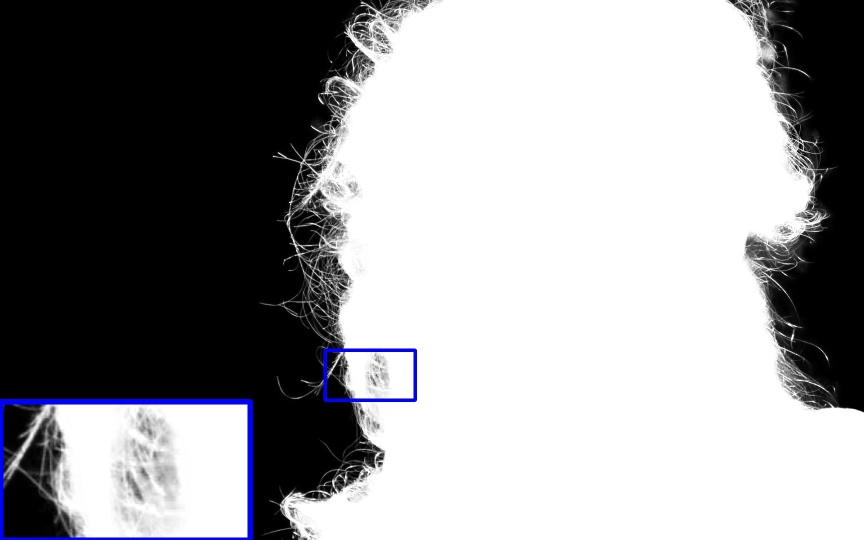}
     \end{minipage}
     }
\vspace{-3mm}
\caption{The qualitative comparison results on P3M dataset, our method produces the most refined alpha mattes and is robustness against inaccurate trimaps. Best viewed by zooming in.}
\label{fig:p3m}
\vspace{-5mm}
\end{figure*}

\textbf{P3M-10K.} We first compare~DiffusionMat with state-of-the-art models on P3M-10K dataset in Tab.~\ref{table:p3m}. Generally speaking, guidance-based approaches tend to outperform guidance-free methods. This is primarily due to the auxiliary guidance effectively reducing learning ambiguity. Moreover, our~DiffusionMat achieves comparable results to its competitors on two test sets. This validates the effectiveness of our approach. In particular, DiffusionMat yields lower Grad and Conn metrics, indicating that our predicted alpha mattes are more perceptually favorable to humans. This can be attributed to our utilization of the generative prior from pre-trained diffusion models, which effectively eliminates discontinuous regions that are rarely encountered in portrait alpha mattes. This further underscores the effectiveness of~DiffusionMat in sequential refinement.


The qualitative comparison results are shown in Fig.~\ref{fig:p3m}. It is evident that DiffusionMat produces the most refined alpha mattes. In the $2_{nd}$ sample, DiffusionMat accurately captures the semi-transparent region around the ``plant'', while other methods failed on capturing it. Notably, our approach also demonstrates robustness against inaccurate trimaps. In the $5_{th}$ sample, where the trimap inaccurately marks regions in the background as unknown, most trimap-based methods erroneously predict alpha mattes around the object. This error even affects the mask-guided method MG~\cite{yu2021mask}. Conversely, DiffusionMat remains unaffected by the imprecise trimap, accurately predicting the alpha matte exclusively around the portrait.~{We attribute this to the strong generative prior of the diffusion model, which has been trained on extensive portrait matte datasets, naturally preventing predictions on background objects.}

\textbf{Human-2K.} We present the quantitative comparison on the Human-2K dataset in Tab.~\ref{table:h2k}. We can see that DiffusionMat also works well on this compositional dataset, which evidences its generalization ability. Same as the P3M-10K dataset, our predicted alpha mattes achieve better performance on Grad and Conn metrics, which is more favorable to human perceptual.


\textbf{Composition-1k.} The quantitative comparison on the Composition-1k dataset is presented in Tab.~\ref{table:adobe}. Our DiffusionMat also outperforms the state-of-the-art methods on this dataset, especially on human perceptual-related Grad and Conn metrics, which evidences that our method is not limited on portrait matting dataset, but also generalize well on the object matting dataset.

\begin{table}[t]
     \caption{Quantitative evaluation on image matting with state-of-the-art methods on Composition-1k dataset. ``Trimap$^*$'' denotes the GT trimap. The lower the better for all metrics. The best results are marked in \textbf{Bold}.}
     \vspace{-5mm}
     \begin{center}
     \setlength{\tabcolsep}{0.05cm}{
     \begin{tabular}{c|c|c|c|c|c}
         \toprule
         Methods &Guidance &SAD$\downarrow$         &MSE$\downarrow$  &Grad$\downarrow$ &Conn$\downarrow$  \\
         \hline
         \rowcolor{gray!20}Learning Based~\cite{zheng2009learning} &None    &113.9   &0.048   &91.6 &122.2 \\
         Closed-Form~\cite{levin2007closed}      &None    &168.1   &0.091   &126.9 &167.9 \\
         \rowcolor{gray!20}KNN~\cite{chen2013knn}                  &None    &175.4   &0.103  &124.1 &176.4 \\
         DIM~\cite{xu2017deep}                   &Trimap$^*$  &50.4&0.014   &31.0 &50.8 \\
         \rowcolor{gray!20}AlphaGAN~\cite{lutz2018alphagan}        &Trimap  &52.4    &0.030   &38.0 &- \\
         IndexNet~\cite{lu2019indices}           &Trimap  &45.8    &0.013   &25.9 &43.7 \\
         \rowcolor{gray!20}HATT~\cite{qiao2020attention}           &Trimap  &44.0    &0.007  &29.3 &46.4 \\
         AdaMatting~\cite{cai2019disentangled}   &Trimap  &41.7    &0.010 &16.8 &- \\
         \rowcolor{gray!20}SampleNet~\cite{tang2019learning}       &Trimap  &40.4    &0.010  &- &- \\
         Fine-Grained~\cite{liu2021towards}      &Trimap  &37.6    &0.009  &18.3 &35.4 \\
         \rowcolor{gray!20}Context-Aware~\cite{hou2019context}     &Trimap  &35.8    &0.008  &17.3 &33.2 \\
         GCA~\cite{li2020natural}                &Trimap  &35.3    &0.009  &16.9 &32.5 \\
         \rowcolor{gray!20}HDMatt~\cite{yu2020high}                &Trimap  &33.5    &0.007  &14.5 &29.9 \\
         MG~\cite{yu2021mask}                    &Mask    &31.5    &0.007  &13.5 &27.3 \\
         \rowcolor{gray!20}TIMINet~\cite{liu2021tripartite}        &Trimap  &29.1    &0.006  &11.5 &25.4 \\
         SIM~\cite{sun2021semantic}              &Trimap  &28.0    &0.006  &10.8 &24.8 \\
         \rowcolor{gray!20}TransMatting~\cite{cai2022transmatting}    &Trimap  &25.0    &0.005  &9.7 &20.2 \\
         MatteFormer~\cite{park2022matteformer}  &Trimap  &23.8    &\textbf{0.004}  &8.7 &{18.9} \\

     \hline
     \rowcolor{gray!20}DiffusionMat   &Trimap  &\textbf{22.8}    &\textbf{0.004} &\textbf{6.8} &\textbf{18.4}  \\
     \bottomrule
     \end{tabular}
     }
     \end{center}
     \vspace{-0.8cm}
     \label{table:adobe}
\end{table}



\subsection{Ablation Studies}

In this section, we perform ablation studies to evaluate DiffusionMat on the Human-2K dataset.  Then we develop various variants with different settings and the modification of loss functions.


\textbf{$\bm{w/}$ \vs $\bm{w/o}$ Diffusion.}
To demonstrate the effectiveness of introducing the pre-trained diffusion model, we propose a vanilla variant ($w/o$ Diffusion) by removing the diffusion model. In this variant, we concatenate the feature $f_{I}$ with the trimap along channels and feed them directly to the correction network for alpha matte prediction. The comparison between $w/$ Diffusion and $w/o$ Diffusion can be found in Tab.~\ref{table:ablation_wio}. The $w/o$ Diffusion variant exhibits poor performance on the Human-2K dataset, while $w/$ Diffusion (DiffusionMat) achieves significantly better results. This suggests that our performance gain does not originate from the image encoder or correction network but rather from our sequential refinement trimap-to-alpha framework.

\begin{figure}[t]
    \centering
    \captionsetup[subfloat]{labelformat=empty,justification=centering}
    \subfloat[Image]{
     \begin{minipage}{0.195\linewidth}
     \includegraphics[width=\linewidth]{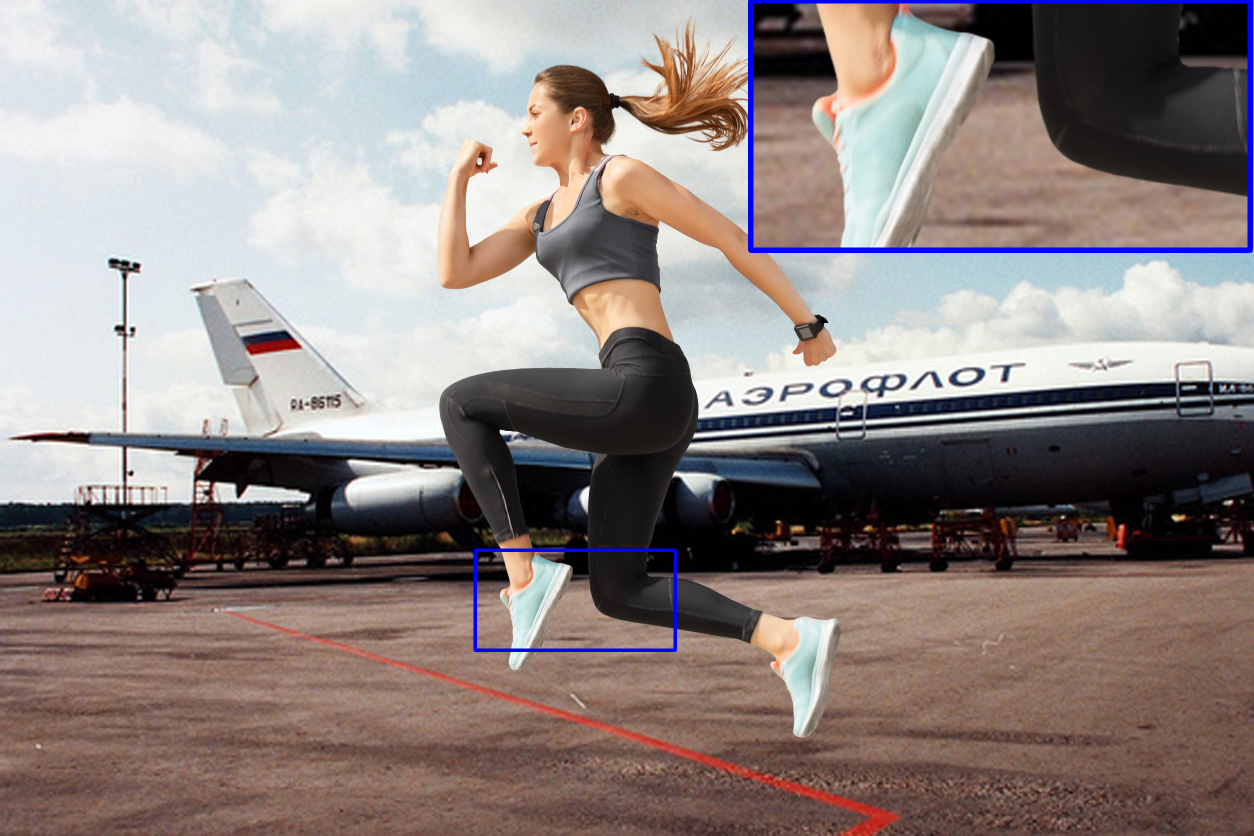}
     \includegraphics[width=\linewidth]{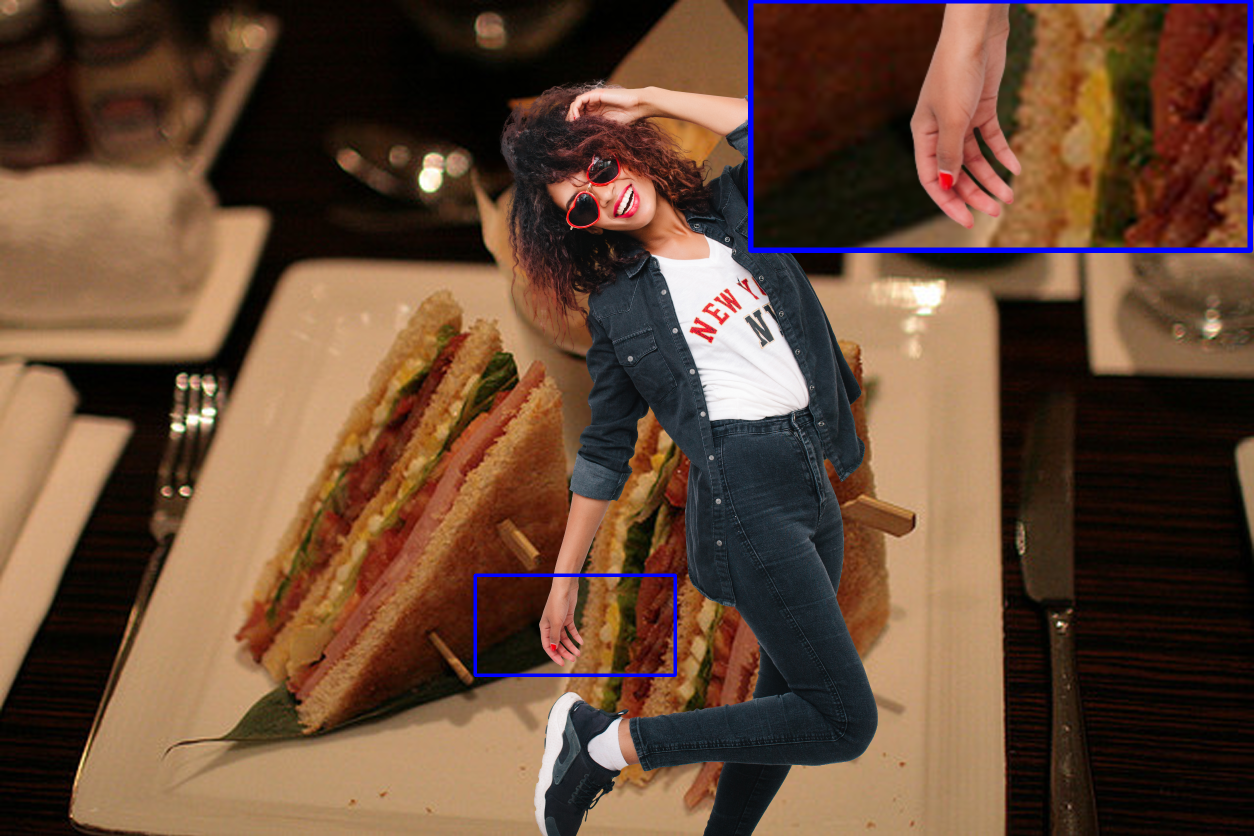}
     \end{minipage}
     }
    \hspace{-2.5mm}
    \subfloat[Trimap]{
     \begin{minipage}{0.195\linewidth}
     \includegraphics[width=\linewidth]{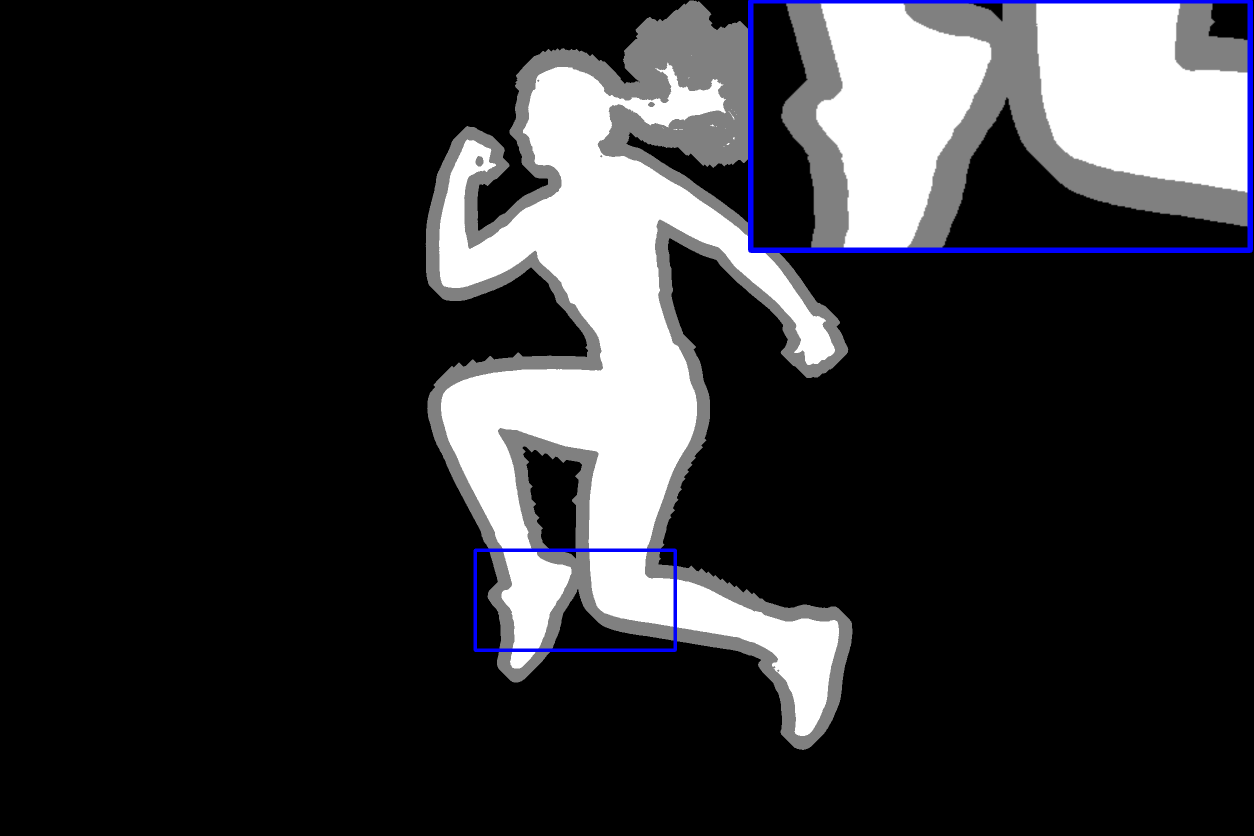}
     \includegraphics[width=\linewidth]{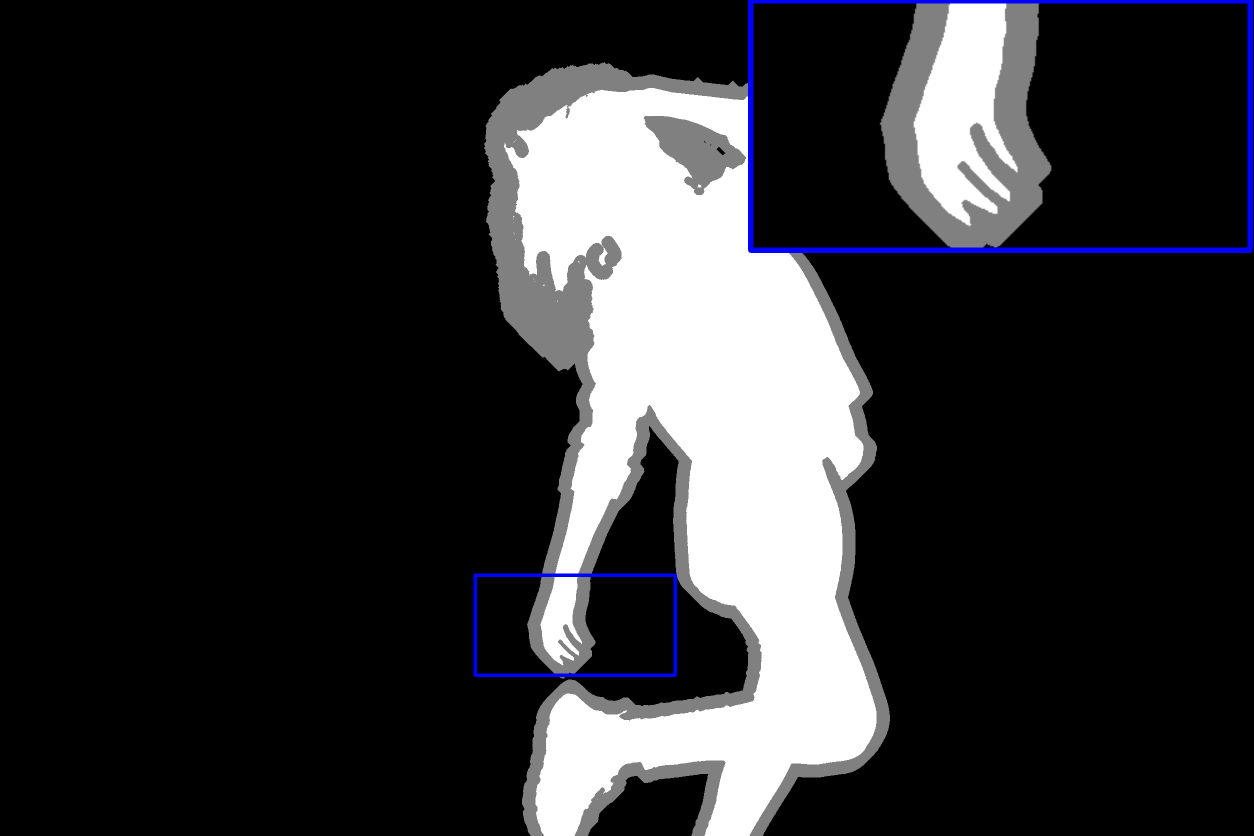}
     \end{minipage}
     }
    \hspace{-2.5mm}    
    \subfloat[$w/o$ Diffusion]{
     \begin{minipage}{0.195\linewidth}
     \includegraphics[width=\linewidth]{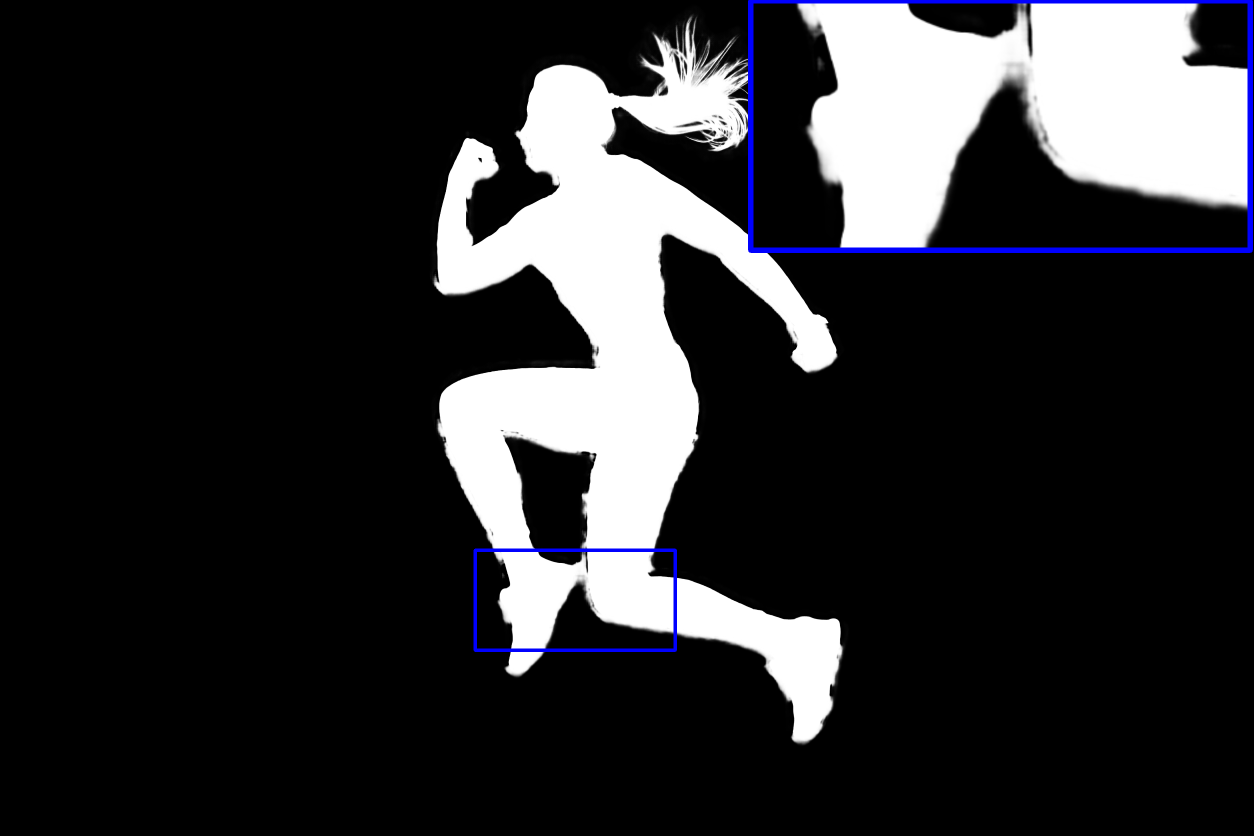}
     \includegraphics[width=\linewidth]{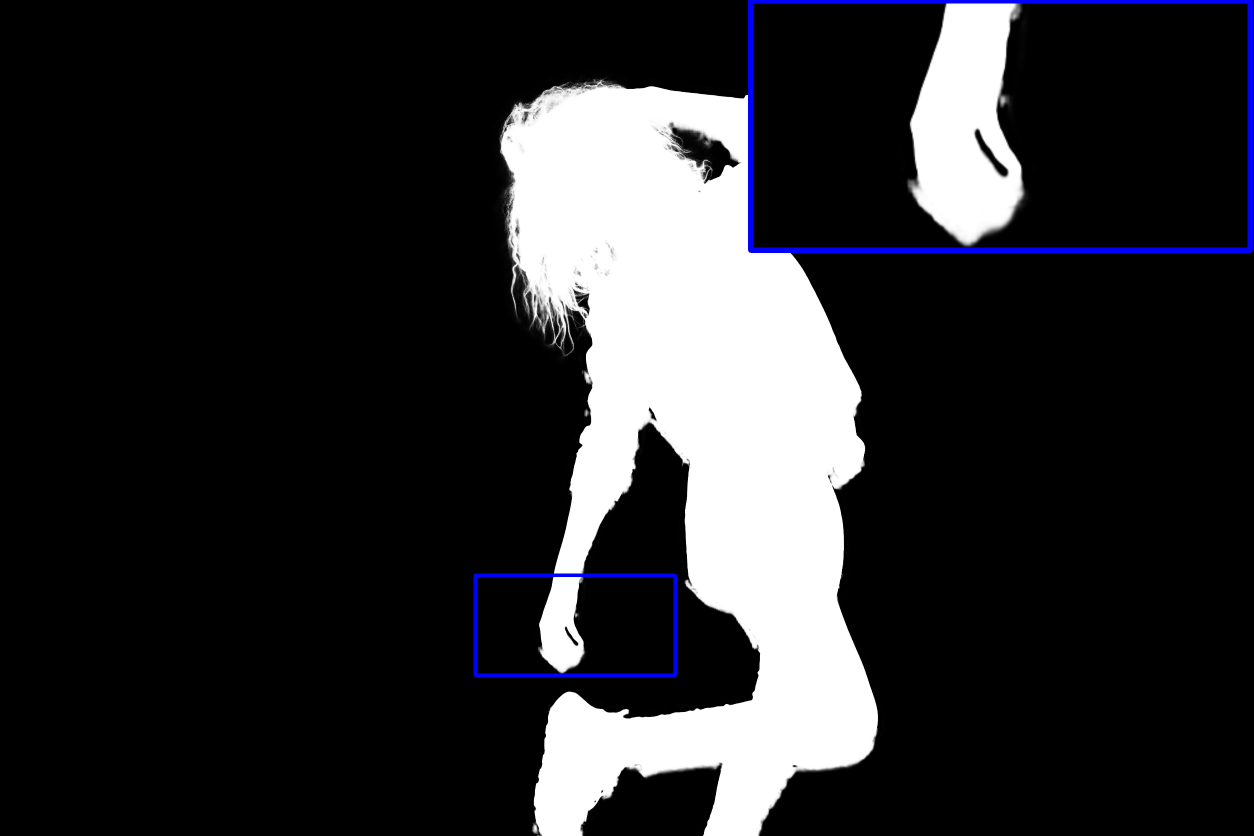}
     \end{minipage}
     }
    \hspace{-2.5mm}    
    \subfloat[$w/$ Diffusion]{
     \begin{minipage}{0.195\linewidth}
     \includegraphics[width=\linewidth]{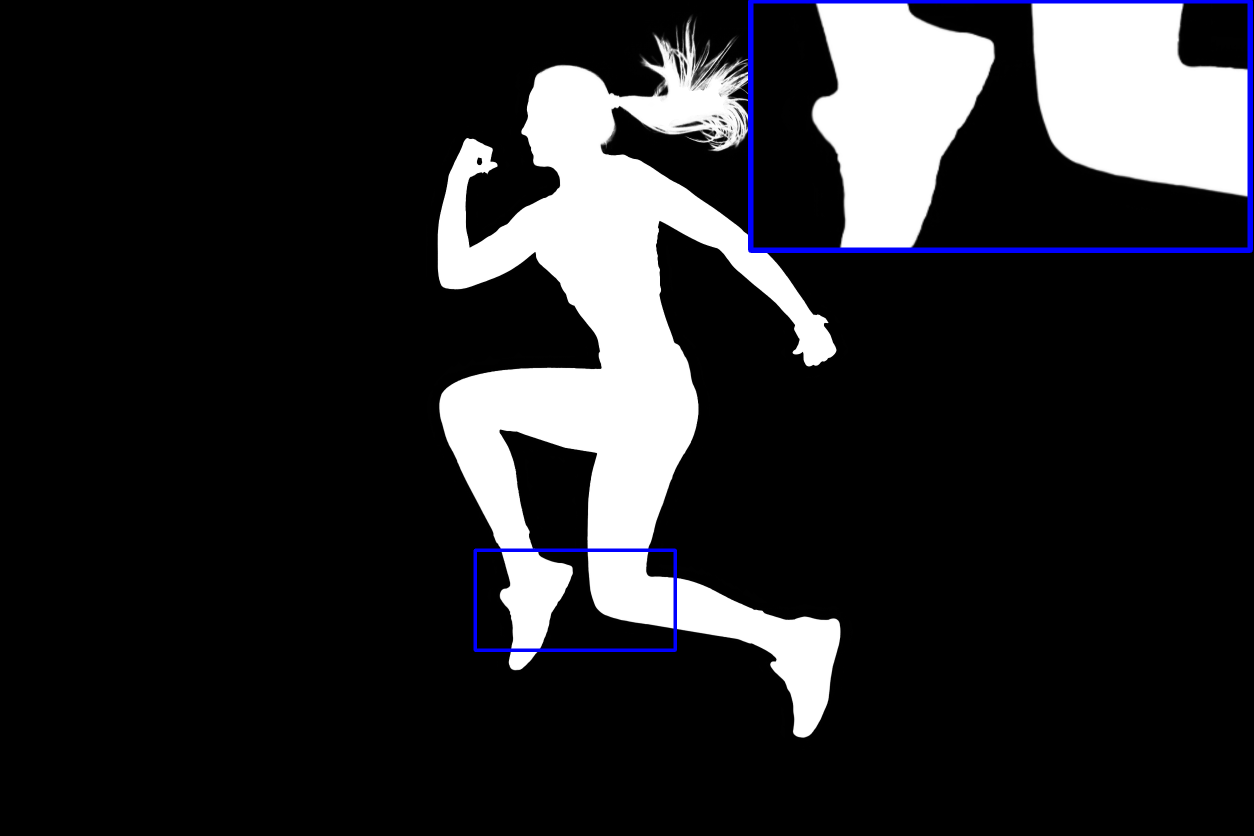}
     \includegraphics[width=\linewidth]{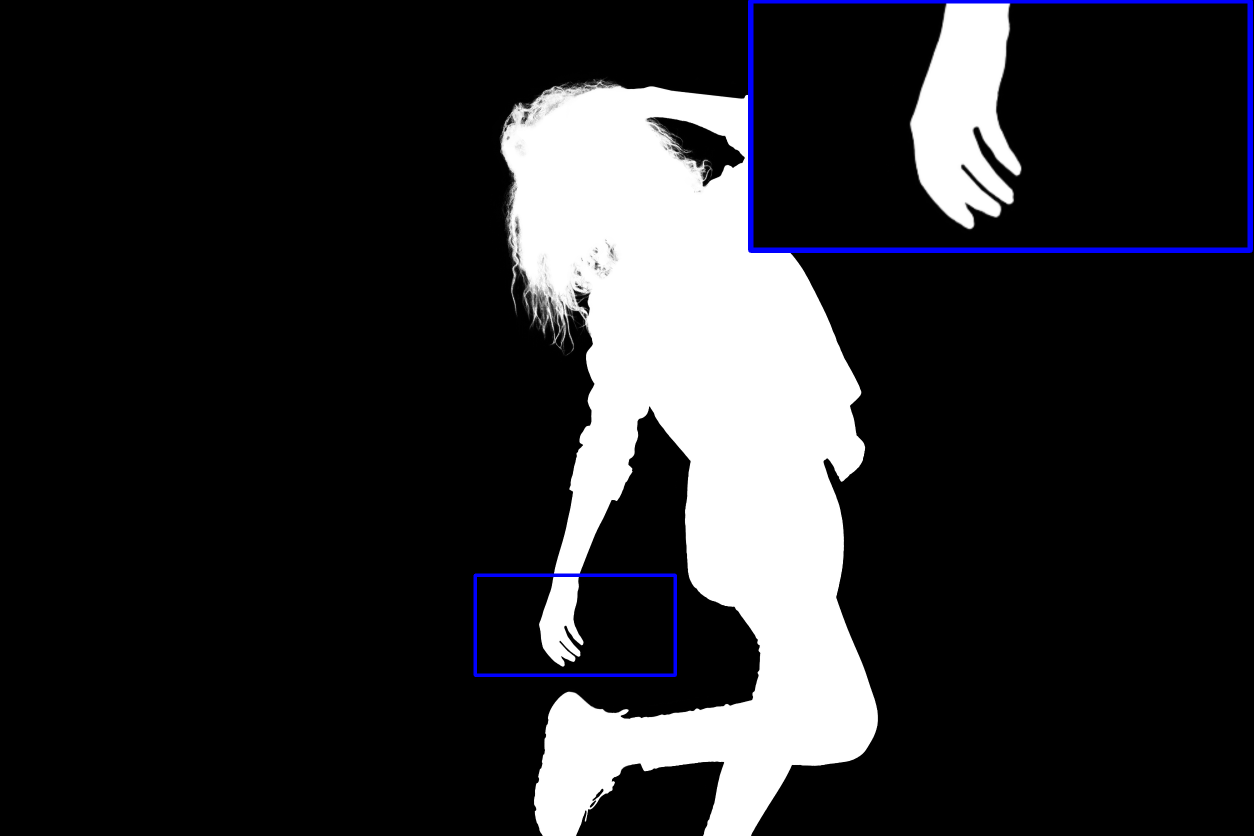}
     \end{minipage}
     }
    \hspace{-2.5mm}
    \subfloat[GT]{
     \begin{minipage}{0.195\linewidth}
     \includegraphics[width=\linewidth]{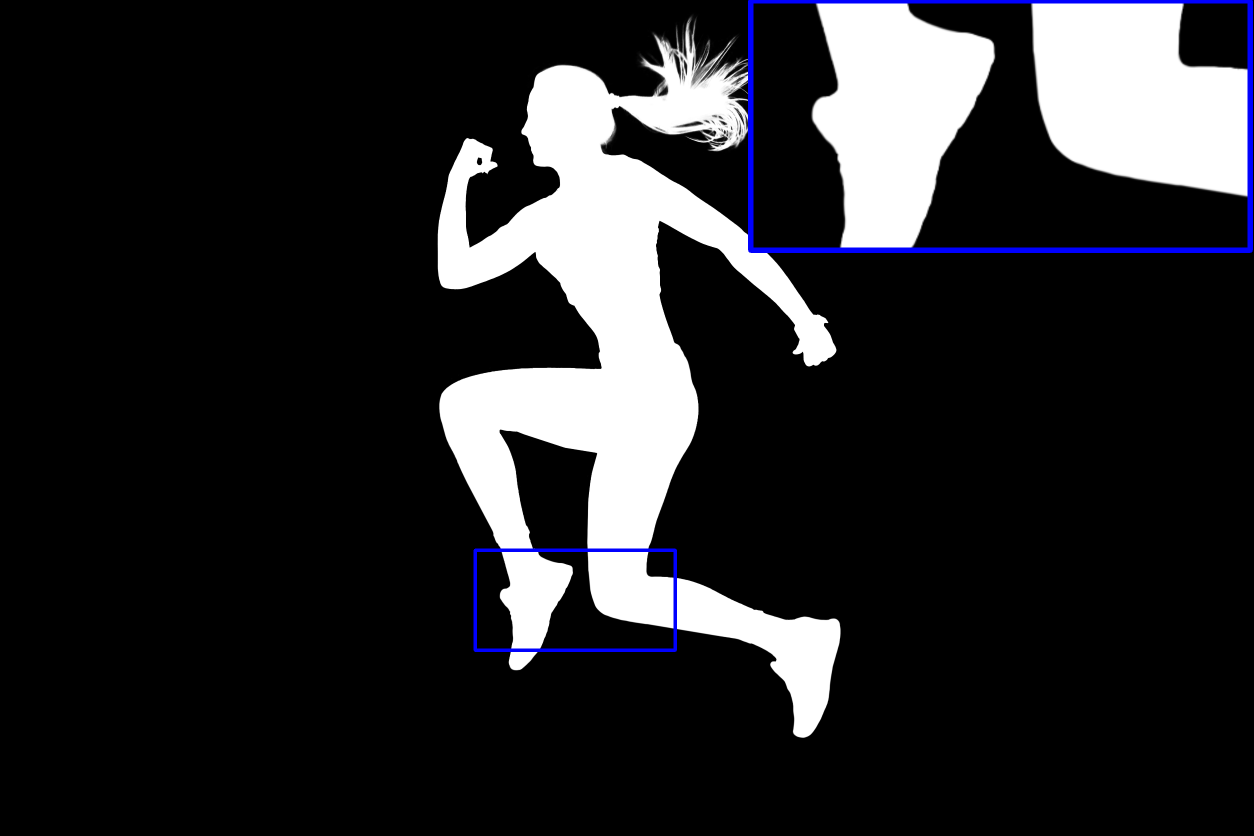}
     \includegraphics[width=\linewidth]{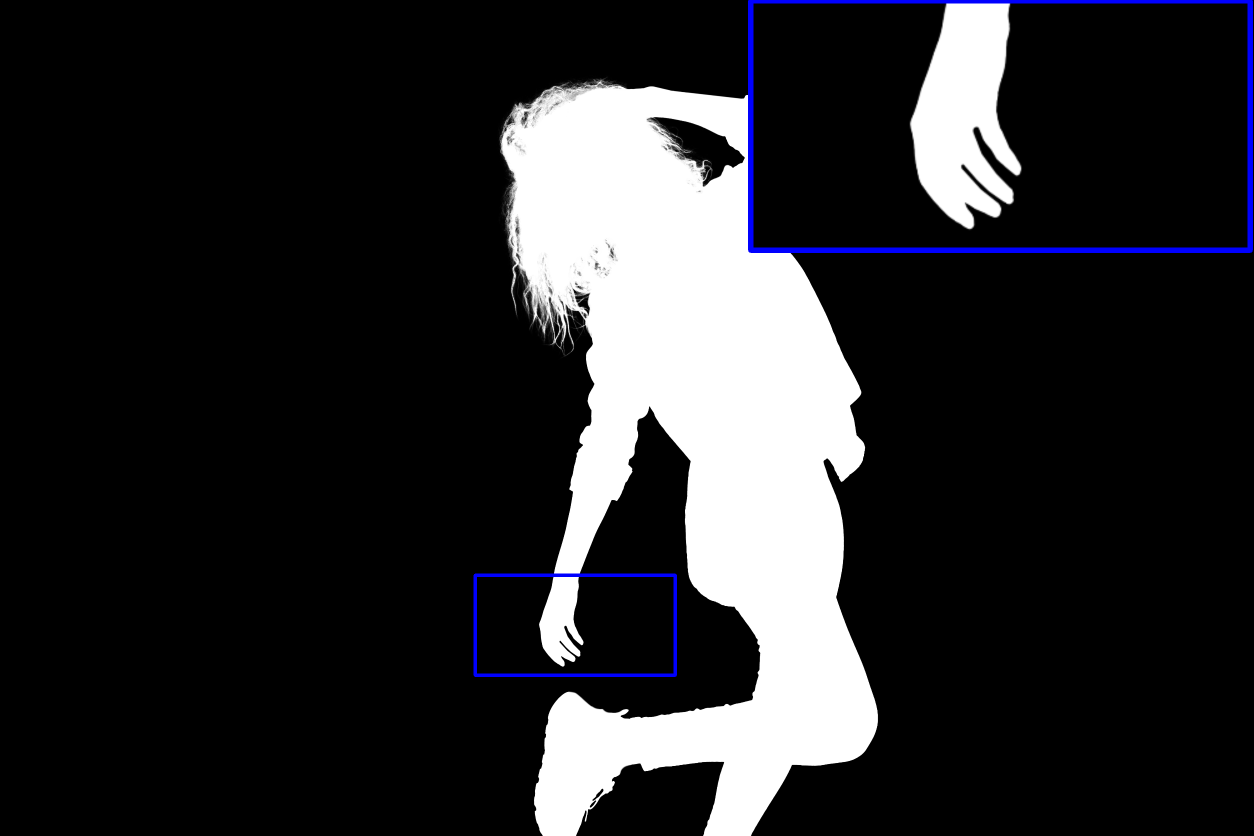}
     \end{minipage}
     }
\vspace{-3mm}
\caption{Without the generative prior of pre-trained diffusion, the vanilla variant $w/o$ Diffusion cannot predict the finest alpha matte.}
\label{fig:ablation_wo}
\vspace{-5mm}
\end{figure}

Quantitative comparisons are provided in Fig.~\ref{fig:ablation_wo}, where the $w/o$ Diffusion variant only produces a coarse alpha matte with significant detail loss. Conversely, $w/$ Diffusion captures fine details effectively. 


\begin{table}[t]
     \caption{Ablation study on $w/$ Diffusion \vs $w/o$ Diffusion model.}
     \vspace{-5mm}
     \begin{center}
     \setlength{\tabcolsep}{0.2cm}{
     \begin{tabular}{c|c|c|c|c}
        \toprule
        Variants &SAD$\downarrow$     &MSE$\downarrow$  &Grad$\downarrow$ &Conn$\downarrow$  \\
        \hline
		{$w/o$} Diffusion   &17.34  &0.0709 &51.33 &16.38  \\
		\hline
      {$w/$} Diffusion      &\textbf{4.04}  &\textbf{0.0020} &\textbf{1.66} &\textbf{2.66} \\
		\bottomrule
     \end{tabular}
     }
     \end{center}
     \vspace{-0.25cm}
     \label{table:ablation_wio}
\end{table}

\textbf{$\bm{w/}$ \vs $\bm{w/o}$ ARP.}
To highlight the effectiveness of our Alpha Reliability Propagation, we trained DiffusionMat by removing this module. The comparison is presented in Tab.~\ref{table:ablation_wioarp}, where the variant ${w/o}$ ARP exhibits inferior performance across all four metrics. ARP module plays a vital role in propagating reliable knowledge from the trimap. It not only eases the learning process but also ensures that DiffusionMat focuses on the unknown regions during alpha matte prediction, resulting in more accurate predictions. 
\begin{table}[t]
     \caption{Ablation study of $w/$ \vs $w/o$ Alpha Reliability Propagation module.}
     \vspace{-5mm}
     \begin{center}
     \setlength{\tabcolsep}{0.2cm}{
     \begin{tabular}{c|c|c|c|c}
        \toprule
        Variants &SAD$\downarrow$     &MSE$\downarrow$  &Grad$\downarrow$ &Conn$\downarrow$  \\
        \hline
      {$w/o$} ARP   &7.07  &0.0062 &3.82 &5.57  \\
      \hline
      {$w/$} ARP   &\textbf{4.04}   &\textbf{0.0020} &\textbf{1.66} &\textbf{2.66} \\
		\bottomrule
     \end{tabular}
     }
     \end{center}
     \vspace{-0.5cm}
     \label{table:ablation_wioarp}
\end{table}






\textbf{Choice of Perturbation Timestep $\bm{T}$.}
We further conduct an ablation study to evaluate the impact of the perturbation timestep ($T$), which controls the noise level added to the trimap and subsequent denoising. We set different values of $T$ with 3 sampling steps in this ablation study and the comparison is shown in Tab.~\ref{table:ablation_t}. Variant $T=1,000$ performs the worst performance, which indicates a higher level of trimap perturbation, providing less useful guidance to the denoising process. Conversely, the variant with $T=100$ also showed suboptimal performance compared to $T=250$, indicates fewer perturbation cannot transform the trimap to an alpha matte successfully. Notably, the $T=250$ variant yields the best performance across all metrics and we use it as our default setting.




\begin{table}[t]
     \caption{Ablation study on corrupt timestep $T$ with fixed 3 sampling steps.}
     \vspace{-5mm}
     \begin{center}
     \setlength{\tabcolsep}{0.2cm}{
     \begin{tabular}{l|c|c|c|c}
        \toprule
        Variants &SAD$\downarrow$     &MSE$\downarrow$  &Grad$\downarrow$ &Conn$\downarrow$  \\
        \hline
       $T=100$ &5.52    &0.0045  &2.96  &4.16   \\
       $T=250$   &5.34 &\textbf{0.0044}  &\textbf{2.71}  &\textbf{3.99} \\
       $T=500$ &\textbf{5.33}    &0.0050  &2.84  &3.74   \\
       $T=750$ &5.46    &0.0047  &2.77  &4.03  \\
       $T=1,000$ &9.40   &0.0126  &7.15  &8.04    \\

      \bottomrule
     \end{tabular}
     }
     \end{center}
     \vspace{-0.5cm}
     \label{table:ablation_t}
\end{table}

\textbf{Choice of Sampling Steps.}
We evaluated the performance and computational costs of DiffusionMat with varying numbers of sampling steps with the input resolution of 512$\times$512. As shown in Tab.~\ref{table:ablation_steps}, the increased sampling steps results in a more accurate alpha matte. However, larger sampling steps also incur higher computational costs, leading to elevated FLOPs values. Notably, compared to the variant with 5 steps, using 10 steps does not yield a significant improvement in accuracy, yet it doubles the computational resources cost. Considering the trade-off between accuracy and efficiency, we set sampling steps as 5.


\begin{table}[t]
     \caption{Ablation study on different sampling steps with the perturbation timestep $T=250$.}
     \vspace{-5mm}
     \begin{center}
     \setlength{\tabcolsep}{0.125cm}{
     \begin{tabular}{l|c|c|c|c|c}
        \toprule
      Variants &SAD$\downarrow$     &MSE$\downarrow$  &Grad$\downarrow$ &Conn$\downarrow$ &FLOPs$\downarrow$  \\
        \hline
        $step=1$   &6.03  &0.0050 &3.41 &4.78  &\textbf{517.96G}\\
        $step=3$   &5.34 &0.0044  &2.71  &3.99 &{1455.54G} \\
        $step=5$   &4.04  &\textbf{0.0020} &1.66 &2.66  &2393.12G\\
        $step=10$  &\textbf{3.99}  &\textbf{0.0020} &\textbf{1.59} &\textbf{2.57} &4737.07G \\

		\bottomrule
     \end{tabular}
     }
     \end{center}
     \vspace{-0.25cm}
     \label{table:ablation_steps}
\end{table} 







\textbf{Ablation Study on Different Losses.}
To evaluate the impact of different loss terms, we developed three variants: 1) $w/o$ $\mathcal{L}_{\mathrm{Inv}}$ by removing the DDIM inversion loss, 2) $w/o$ $\mathcal{L}_{\mathrm{\alpha}}$ by omitting the alignment between one-step alpha matte and GT alpha matte, and 3) $w/o$ $\mathcal{L}_{\mathrm{Comp}}$ by eliminating the composition loss. The comparison is detailed in Tab.~\ref{table:ablation_loss}. It is evident that the variant $w/o$ $\mathcal{L}_{\mathrm{Inv}}$ presents inferior performance across all metrics. This highlights the crucial role played by the DDIM inversion loss in DiffusionMat, as it corrects the denoised results with the GT inverted trajectory. Additionally, both $\mathcal{L}_{\mathrm{Comp}}$ and $\mathcal{L}_{\mathrm{\alpha}}$ contributed to improved performance.


\begin{table}[t]
     \caption{Ablation study on the modification of loss functions.}
     \vspace{-5mm}
     \begin{center}
     \setlength{\tabcolsep}{0.2cm}{
     \begin{tabular}{c|c|c|c|c}
        \toprule
      Variants &SAD$\downarrow$     &MSE$\downarrow$  &Grad$\downarrow$ &Conn$\downarrow$  \\
        \hline
		$w/o$ $\mathcal{L}_{\mathrm{Inv}}$     &4.96  &0.0034  &2.51  &3.44  \\
        $w/o$ $\mathcal{L}_{\mathrm{\alpha}}$  &4.34  &0.0024  &1.84  &2.91  \\
        $w/o$ $\mathcal{L}_{\mathrm{Comp}}$    &4.77  &0.0027  &1.98  &3.19  \\
        \hline
        DiffusionMat    &\textbf{4.04}  &\textbf{0.0020} &\textbf{1.66} &\textbf{2.66}  \\
        \bottomrule
     \end{tabular}
     }
     \end{center}
     \vspace{-0.5cm}
     \label{table:ablation_loss}
\end{table}





\begin{figure}[!t]
    \centering
    \captionsetup[subfloat]{labelformat=empty,justification=centering}
    \includegraphics[width=0.89\linewidth]{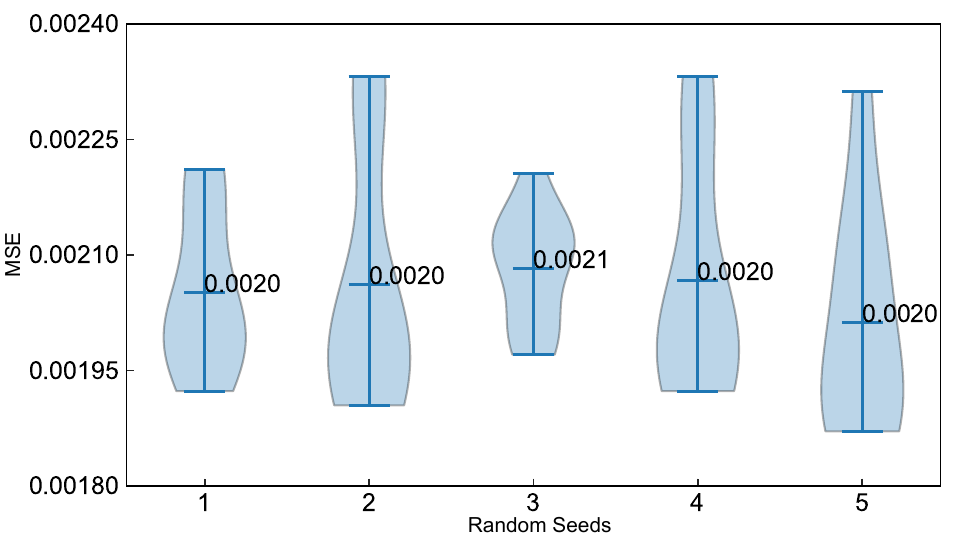}
\vspace{-3mm}
\caption{We train DiffusionMat with 5 different random seeds and evaluate each one 10 times. They get the very similar mean values on MSE metric.}
\vspace{-1mm}
\label{fig:random}
\end{figure}

\textbf{Random seeds.}
To evaluating the stability of DiffusionMat under varying random seeds, we follow DiffusionDet~\cite{chen2022diffusiondet} that trains 5 models with different initial seeds and evaluate their performance with 10 different random seeds. As illustrated in Fig.~\ref{fig:random}, DiffusionMat demonstrates consistent mean values on the MSE metric, signifying its robustness to different sources of noise. This stability can be attributed to our deterministic denoising approach, which effectively reduces randomness during the denoising process. 


\begin{figure}[t]
    \centering
    \captionsetup[subfloat]{labelformat=empty,justification=centering}
    \subfloat[\scriptsize{Trimap/Image}]{
     \begin{minipage}{0.1575\linewidth}
     \includegraphics[width=\linewidth]{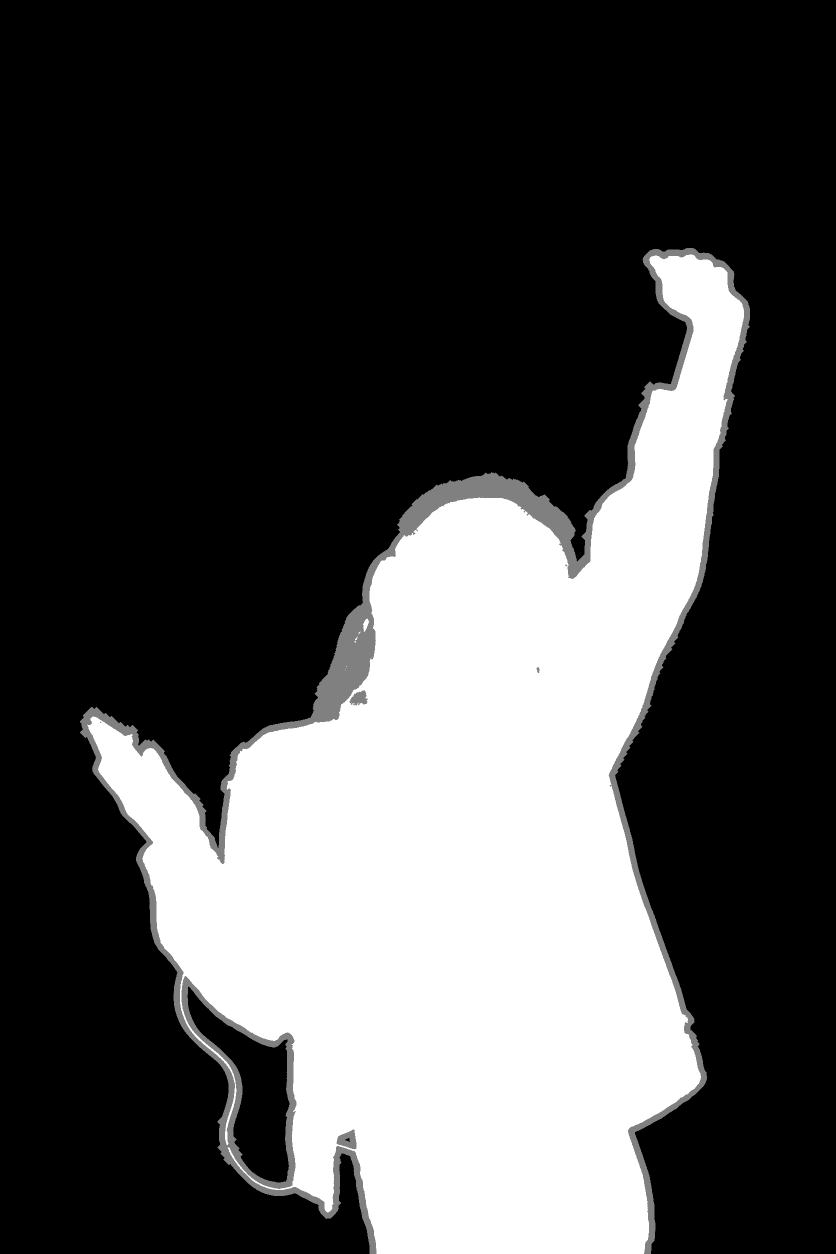}
     \includegraphics[width=\linewidth]{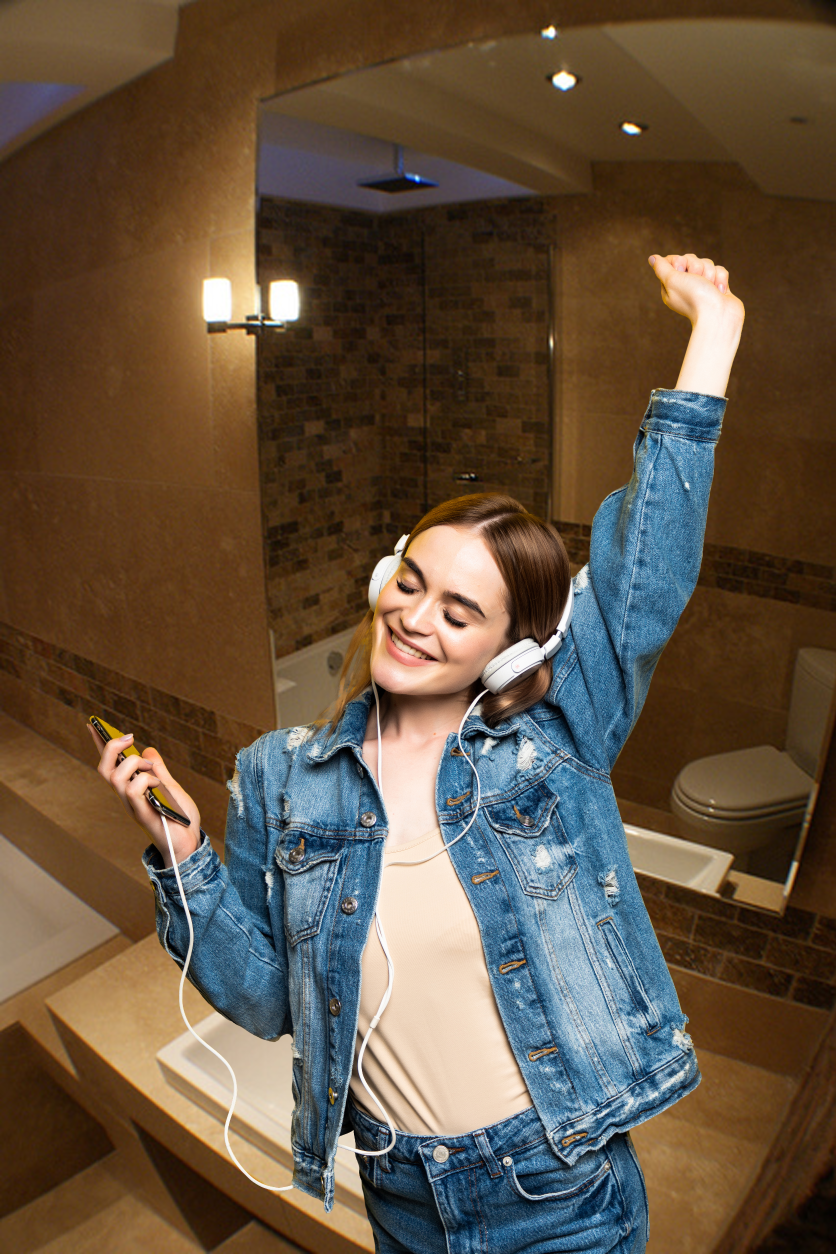}
     \end{minipage}
     }
    \hspace{-2.5mm}
    \subfloat[\scriptsize{$t=200$}]{
     \begin{minipage}{0.1575\linewidth}
     \includegraphics[width=\linewidth]{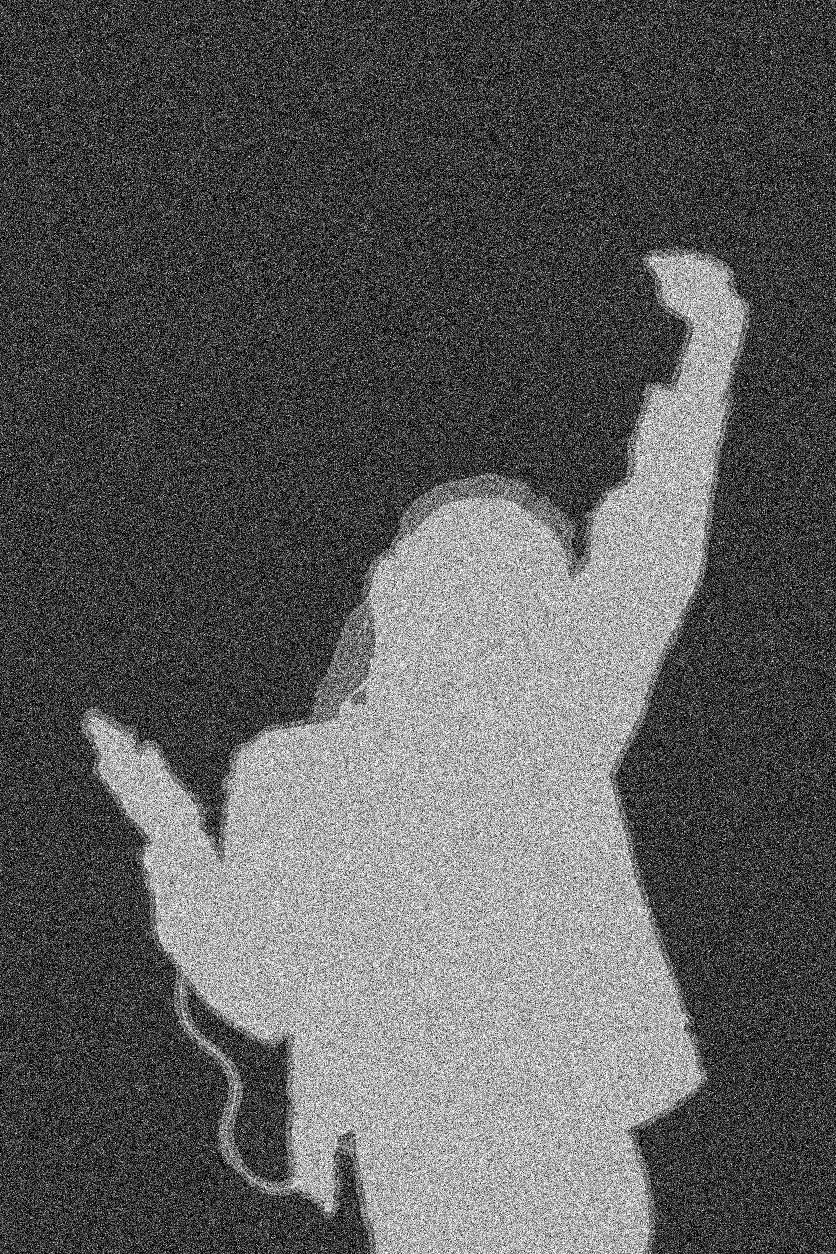}
     \includegraphics[width=\linewidth]{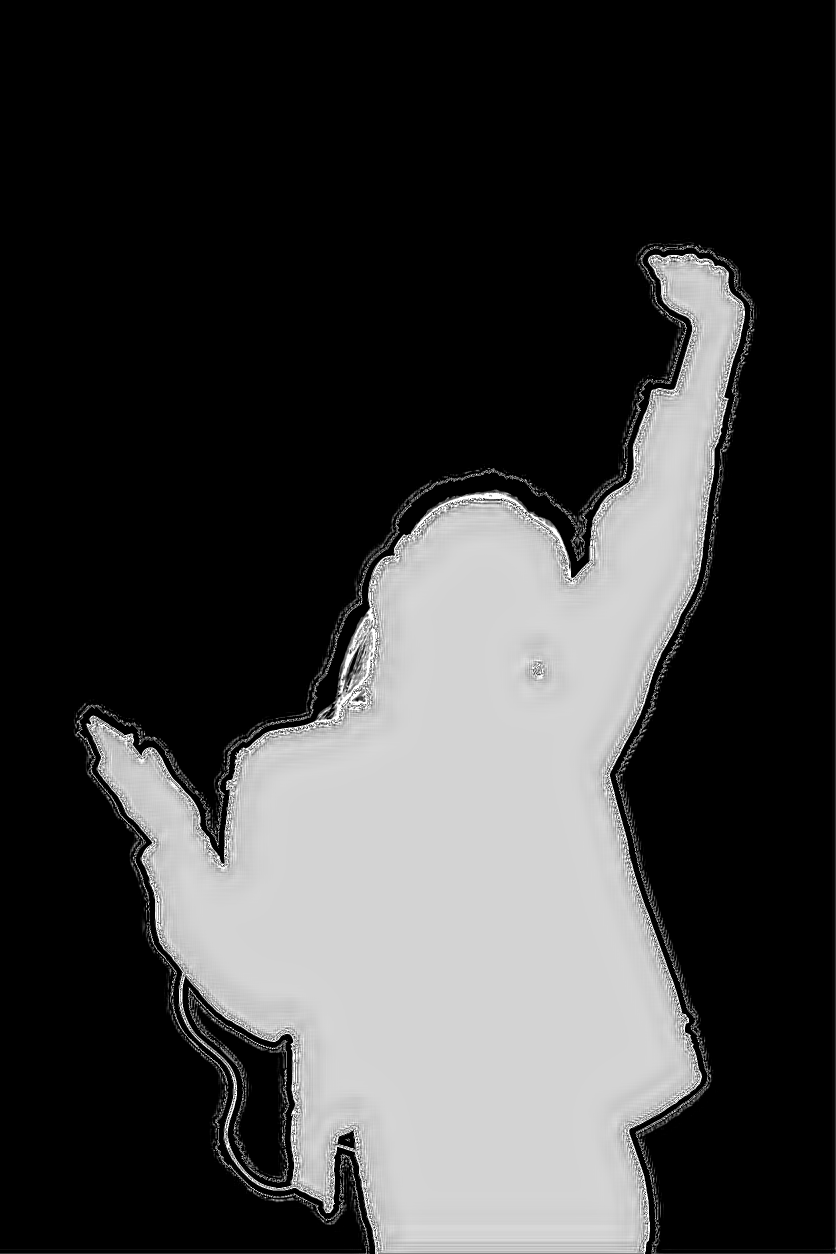}
     \end{minipage}
     }
    \hspace{-2.5mm}    
    \subfloat[\scriptsize{$t=150$}]{
     \begin{minipage}{0.1575\linewidth}
     \includegraphics[width=\linewidth]{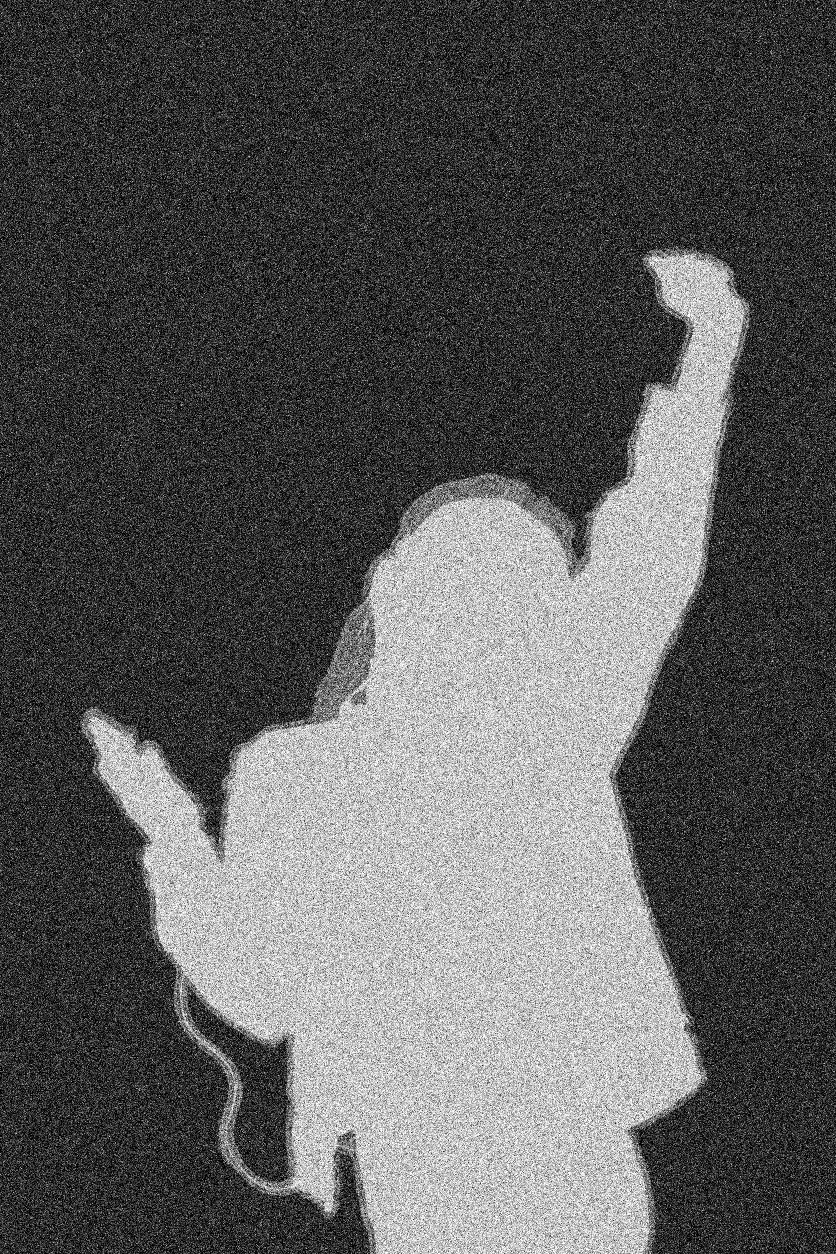}
     \includegraphics[width=\linewidth]{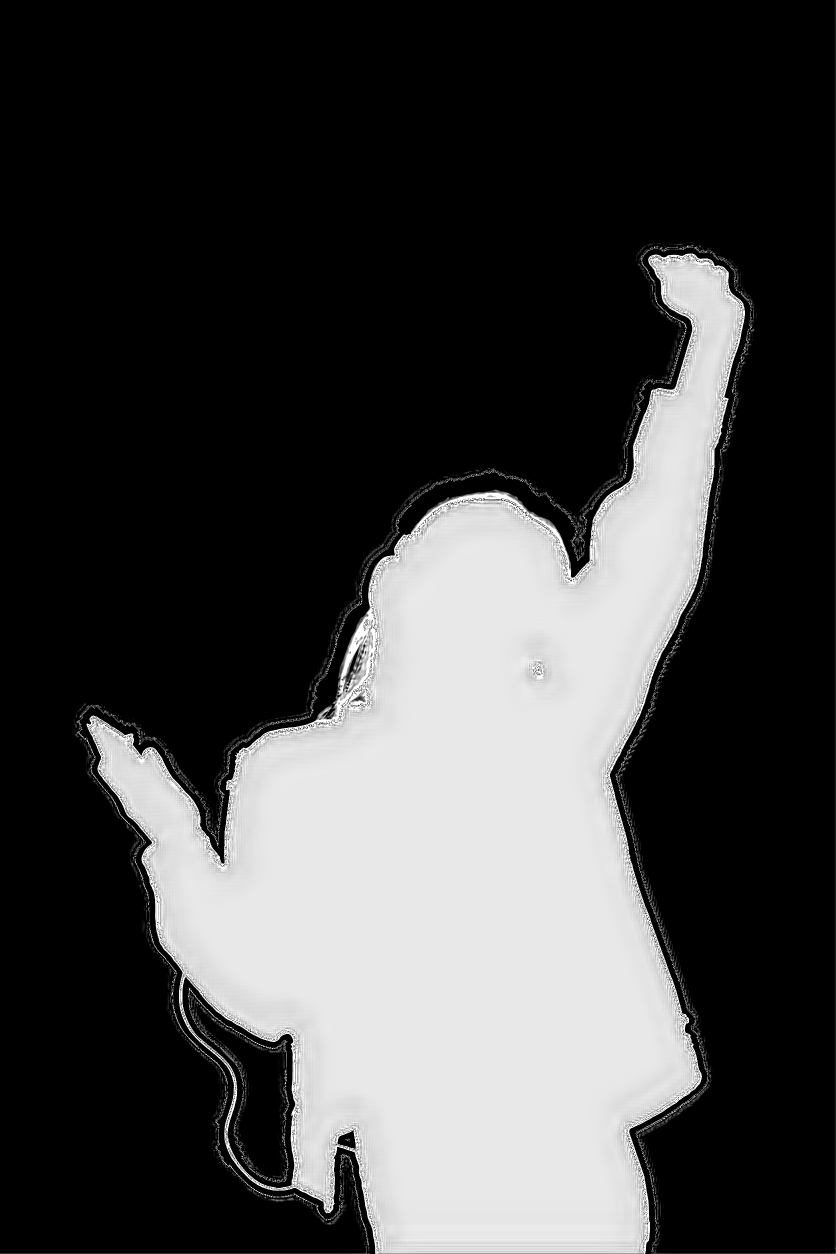}
     \end{minipage}
     }
    \hspace{-2.5mm}    
    \subfloat[\scriptsize{$t=100$}]{
     \begin{minipage}{0.1575\linewidth}
     \includegraphics[width=\linewidth]{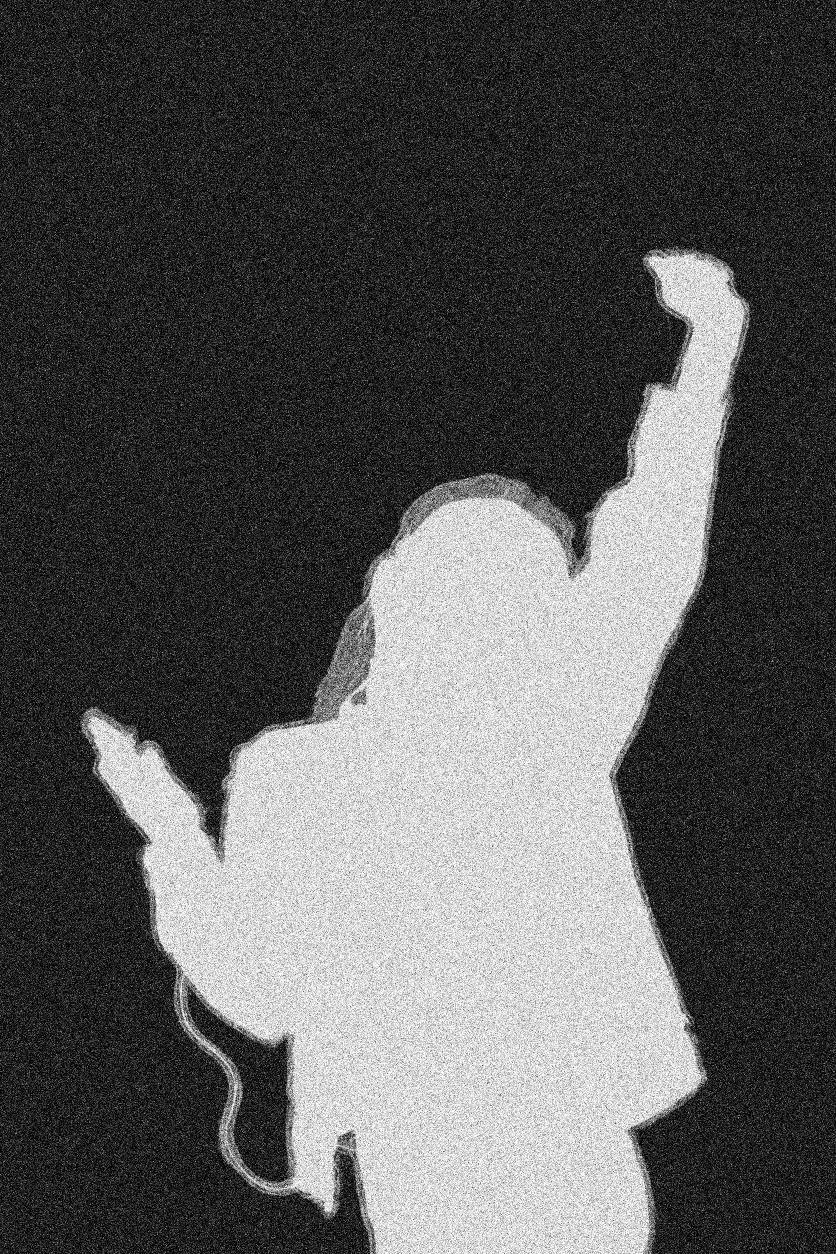}
     \includegraphics[width=\linewidth]{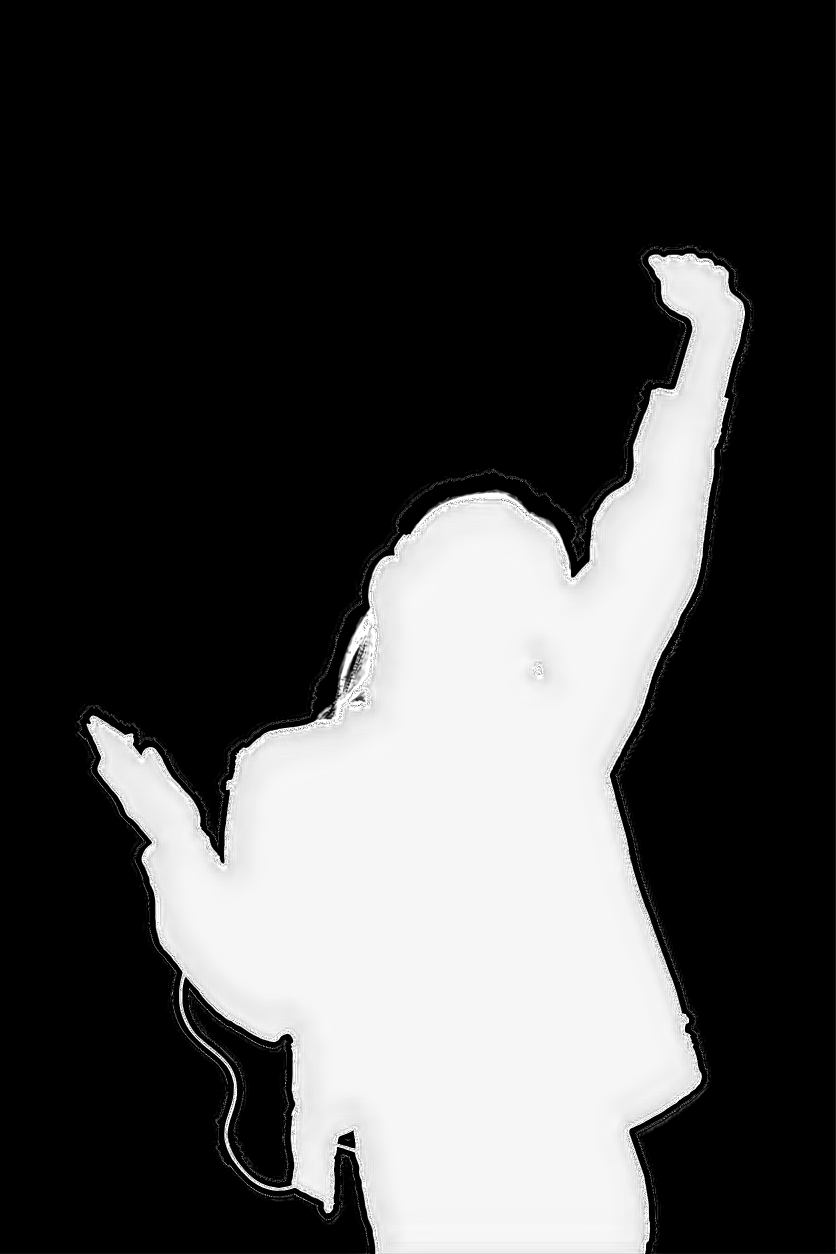}
    \end{minipage}
     }
    \hspace{-2.5mm}
    \subfloat[\scriptsize{$t=50$}]{
     \begin{minipage}{0.1575\linewidth}
     \includegraphics[width=\linewidth]{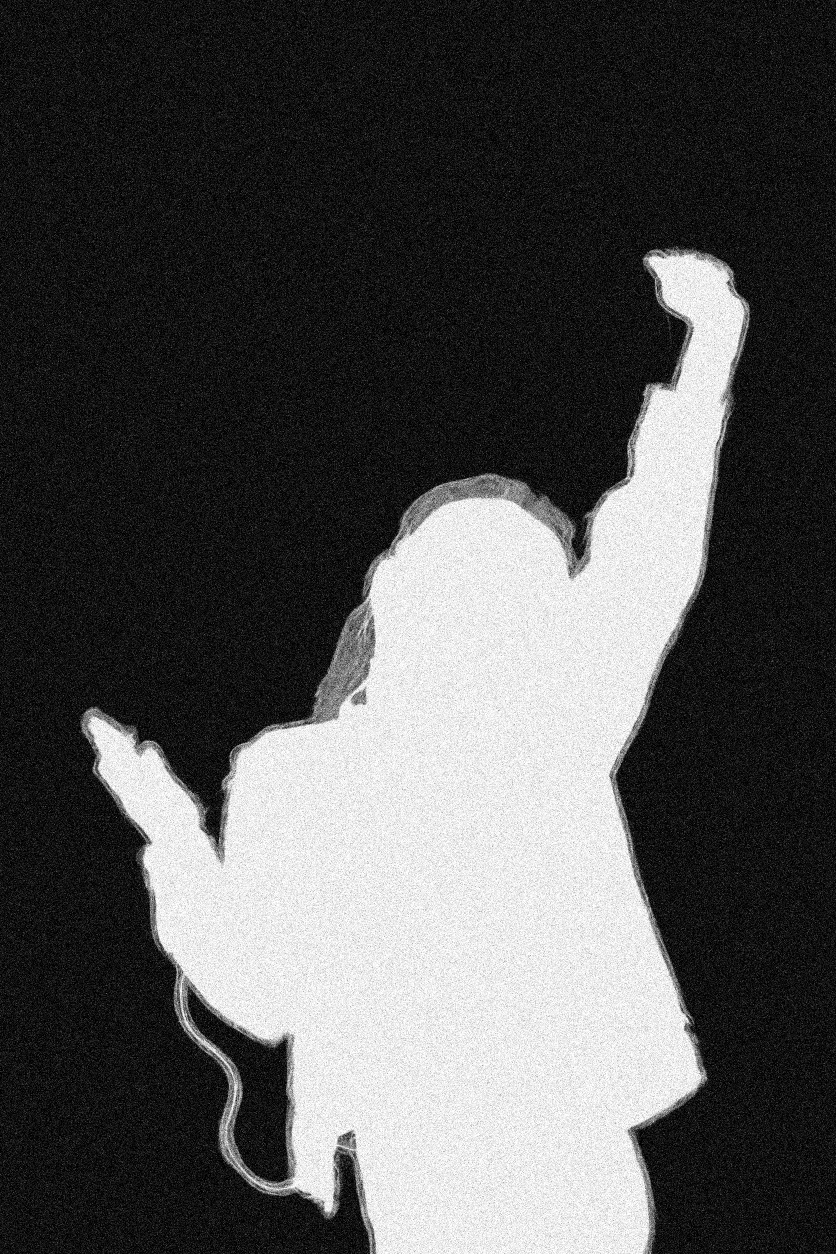}
     \includegraphics[width=\linewidth]{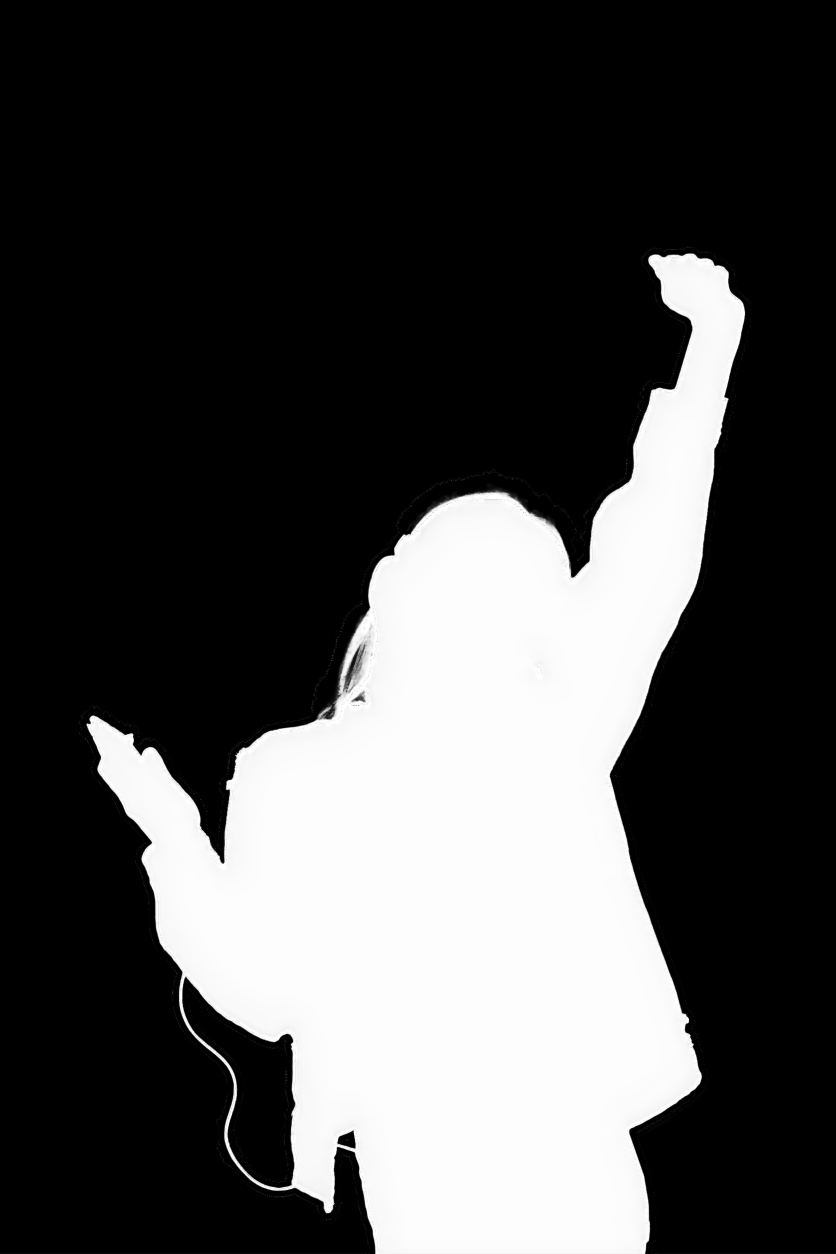}
    \end{minipage}
     }   
    \hspace{-2.5mm}
    \subfloat[\scriptsize{$t=0$}]{
     \begin{minipage}{0.1575\linewidth}
     \includegraphics[width=\linewidth]{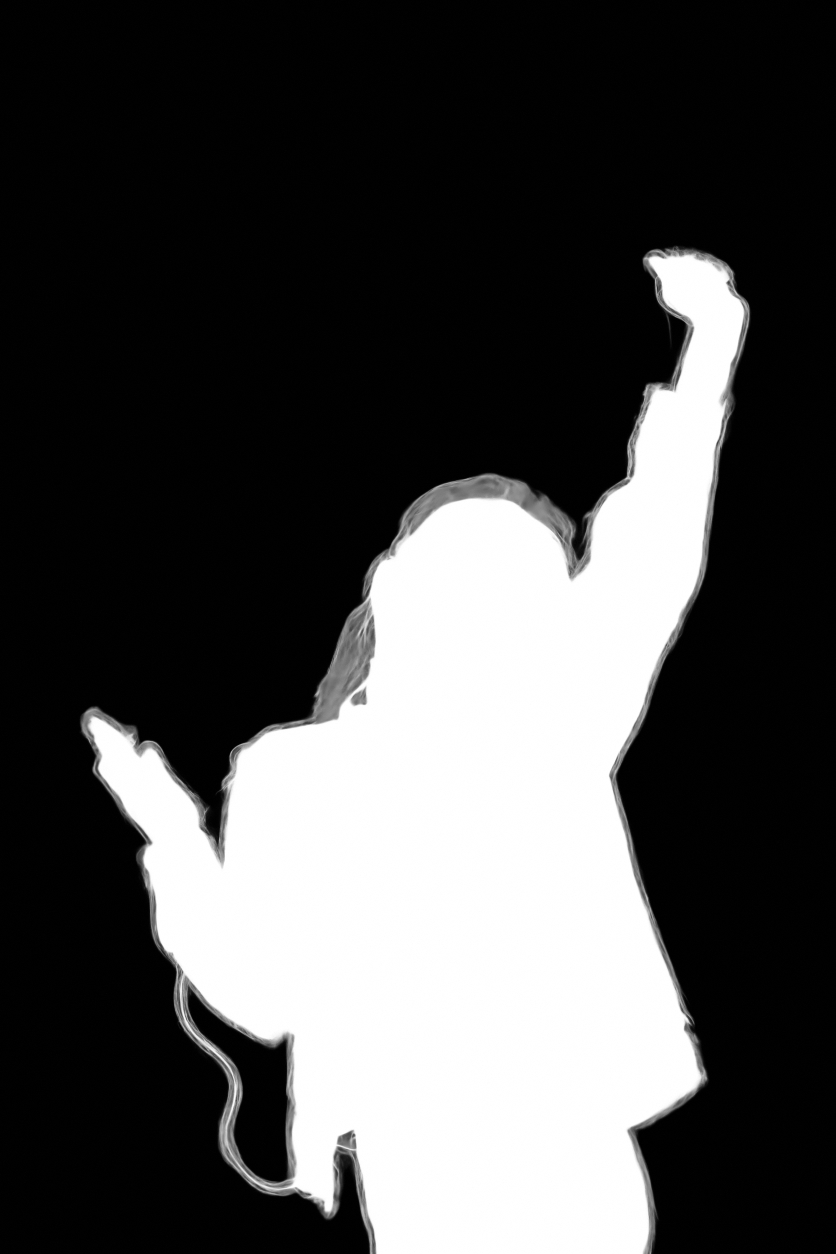}
     \includegraphics[width=\linewidth]{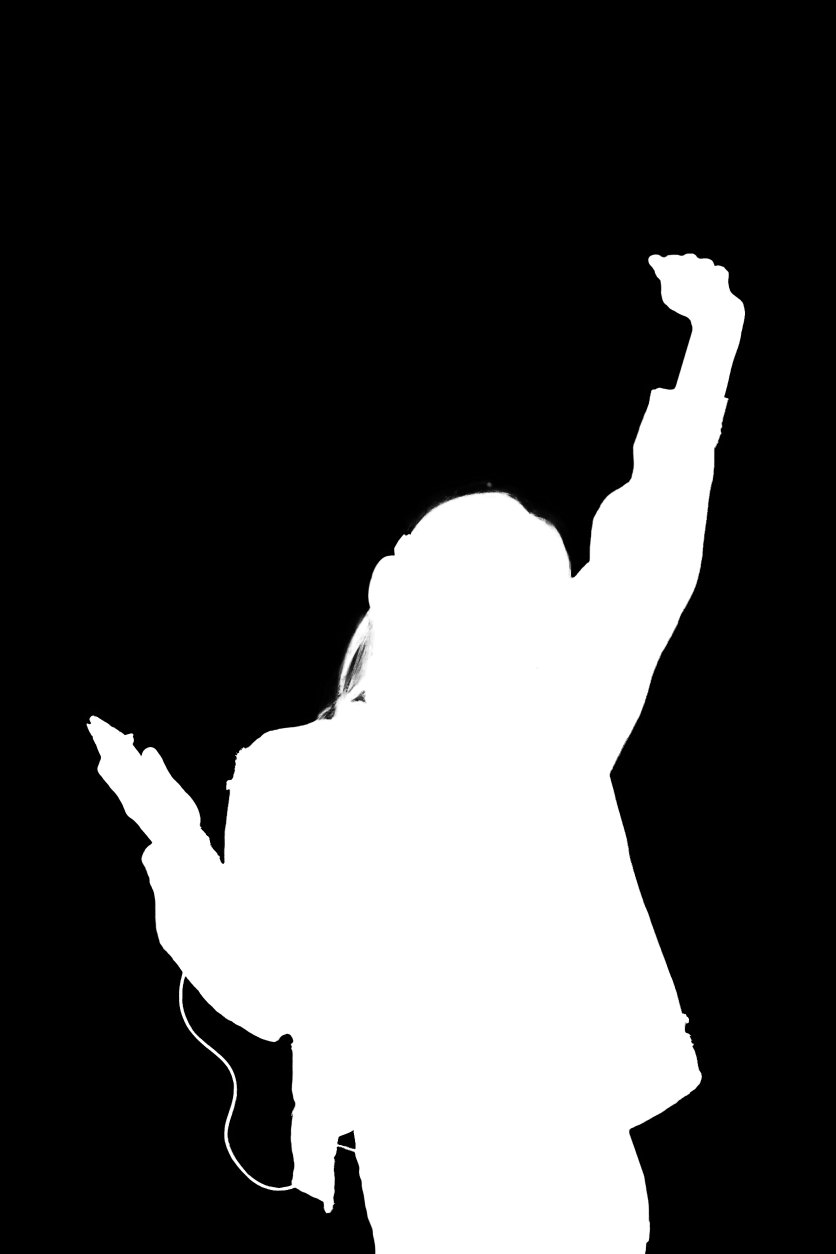}
    \end{minipage}
     }
    \rotatebox[origin=c]{270}{ \hspace{-40mm} \scriptsize{\texttt{Original}}  \hspace{6mm}\scriptsize{\texttt{Corrected}}}\\
\vspace{-3mm}
\caption{Visual comparison between original denoised results (obtained by SDEdit~\cite{meng2021sdedit}) and corrected results on different denoising timesteps.}
\vspace{-5mm}
\label{fig:progress}
\end{figure}

\textbf{Visualization of Denoised Results.}
We provide visualizations of the original and corrected denoised results at different timesteps in Fig.~\ref{fig:progress}. The original denoised results primarily preserve global semantic structures but fail to capture local details. In contrast, the corrected denoised results effectively capture these local details, resulting in an accurate alpha matte. This highlights the effectiveness of our sequential refinement learning strategy.

\section{Conclusion and Discussions}
\label{sec:conclusion}


In this paper, we present DiffusionMat, a framework that transforms a noise-injected trimap into a clean alpha matte by treating image matting as a sequential refinement learning process. Initially, we perturb the trimap with noise and subsequently denoise it using a pre-trained diffusion model. We introduce a correction module designed to adjust each intermediate denoised result. Furthermore, we propose the Alpha Reliability Propagation module, which enhances the efficacy of the correction module, particularly in unknown regions. Our experiments across various image matting datasets demonstrate DiffusionMat's superior performance.

The primary limitation of our DiffusionMat lies in its computational efficiency. It requires more processing time compared to traditional single-pass image matting methods, taking approximately 0.6 seconds for an input resolution of $512\times512$. However, these drawbacks could potentially be mitigated through the development of more efficient diffusion models, which is a direction we aim to explore in future research.

\section{Appendix}
We provide the pseudo code for training and inference our DiffusionMat in Alg.~\ref{alg:train} and Alg.~\ref{alg:sample}.

\begin{figure*}[!h]
    \centering
    \captionsetup[subfloat]{labelformat=empty,justification=centering}
    \subfloat[Image]{
     \begin{minipage}{0.12\linewidth}
     \includegraphics[width=\linewidth]{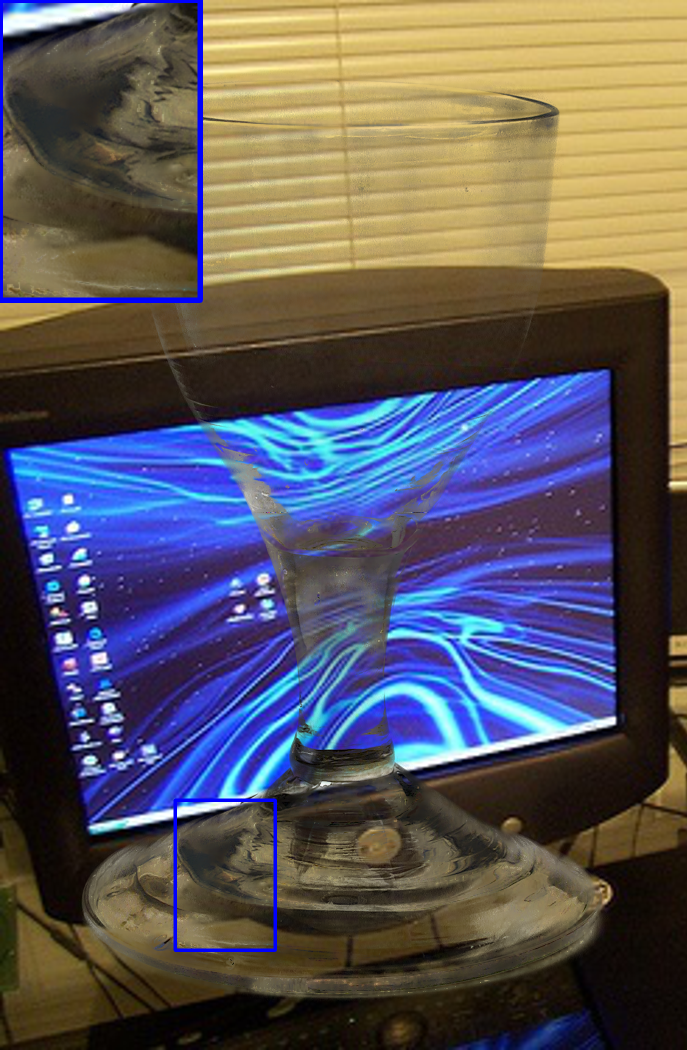}
     \includegraphics[width=\linewidth]{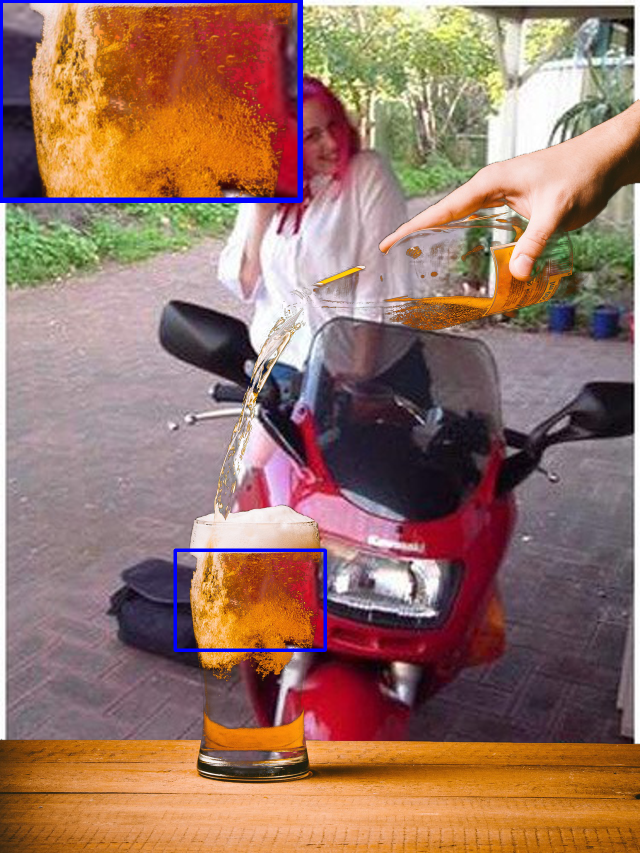}
     \includegraphics[width=\linewidth]{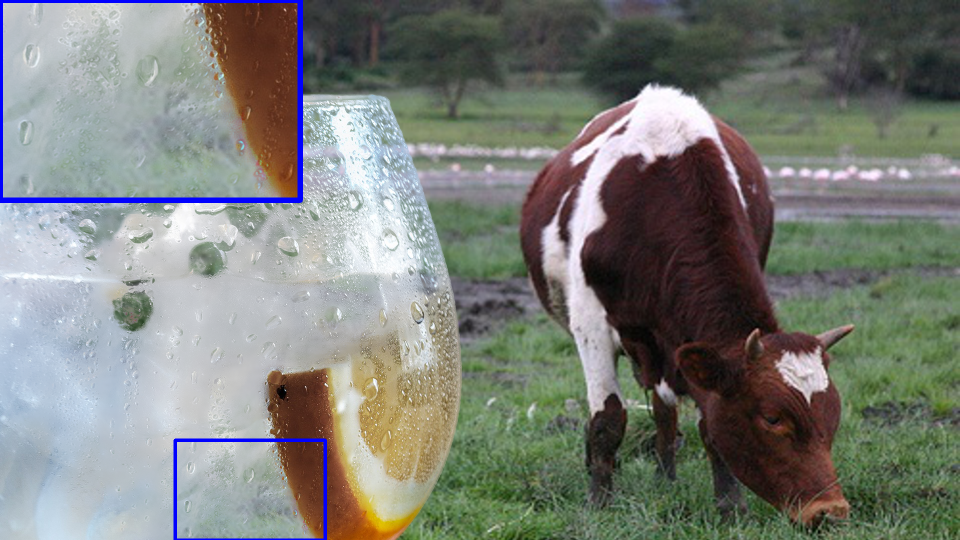}
     \includegraphics[width=\linewidth]{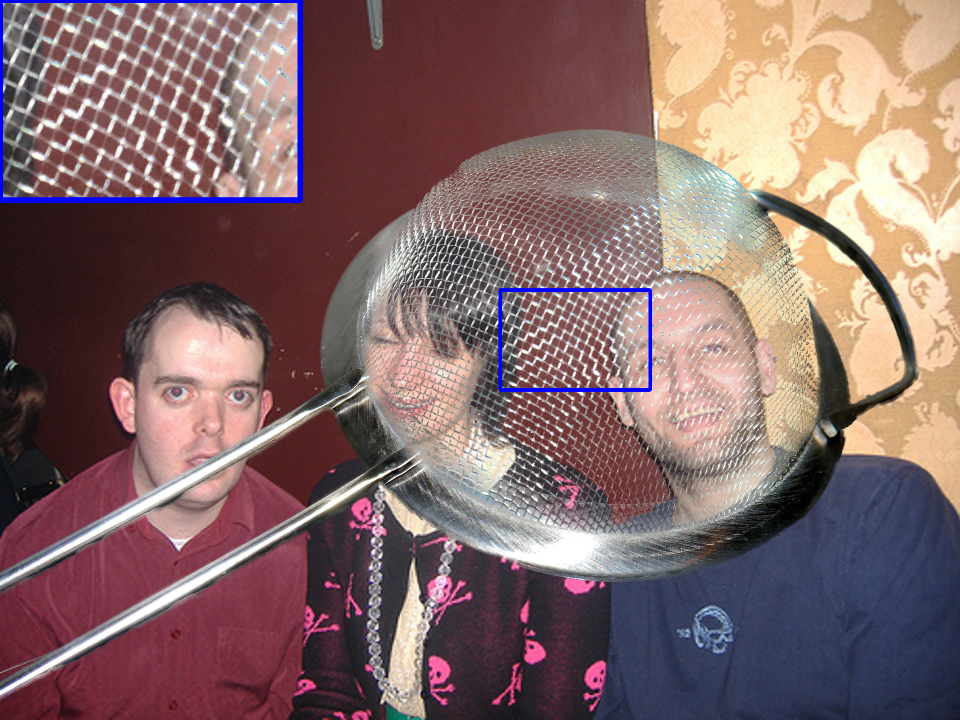}
     \includegraphics[width=\linewidth]{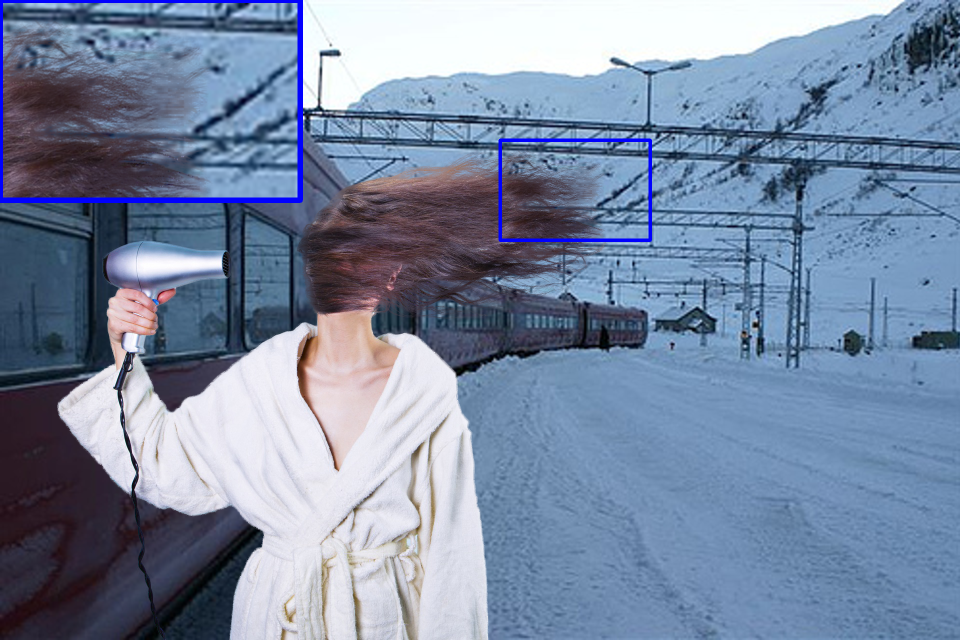}
     \end{minipage}
     }
    \hspace{-2.5mm}
    \subfloat[Trimap]{
     \begin{minipage}{0.12\linewidth}
     \includegraphics[width=\linewidth]{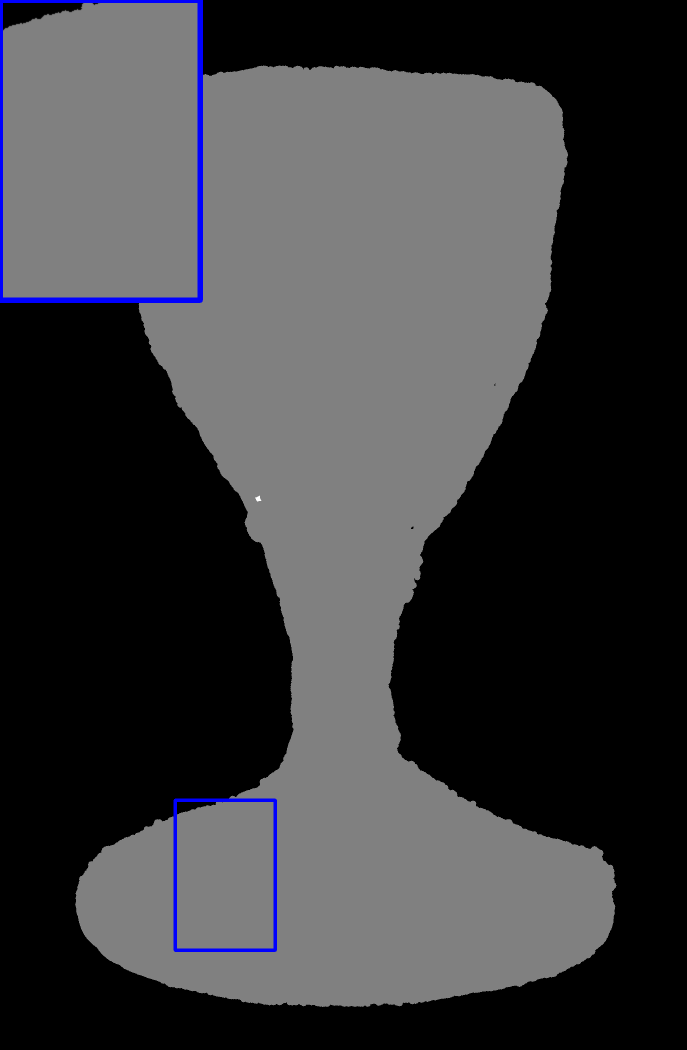}
     \includegraphics[width=\linewidth]{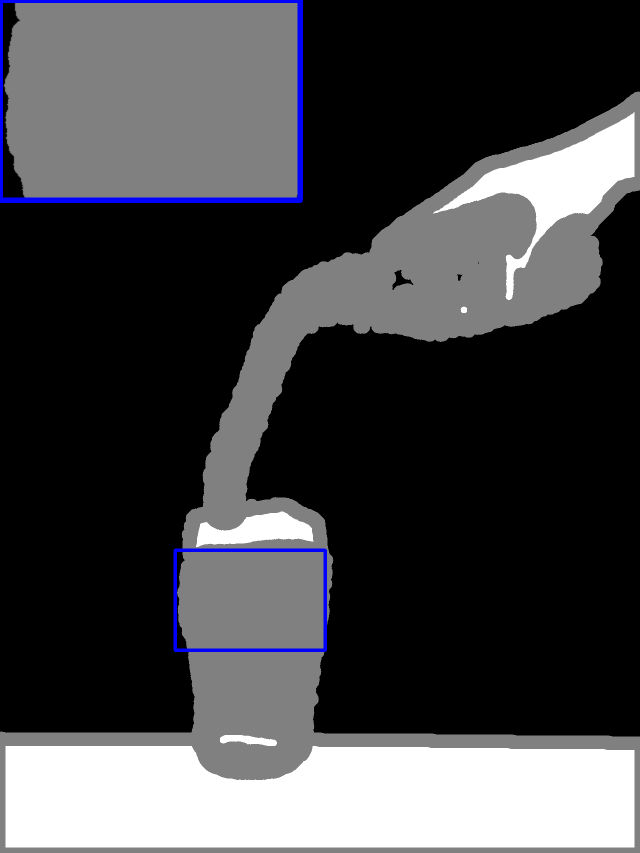}
     \includegraphics[width=\linewidth]{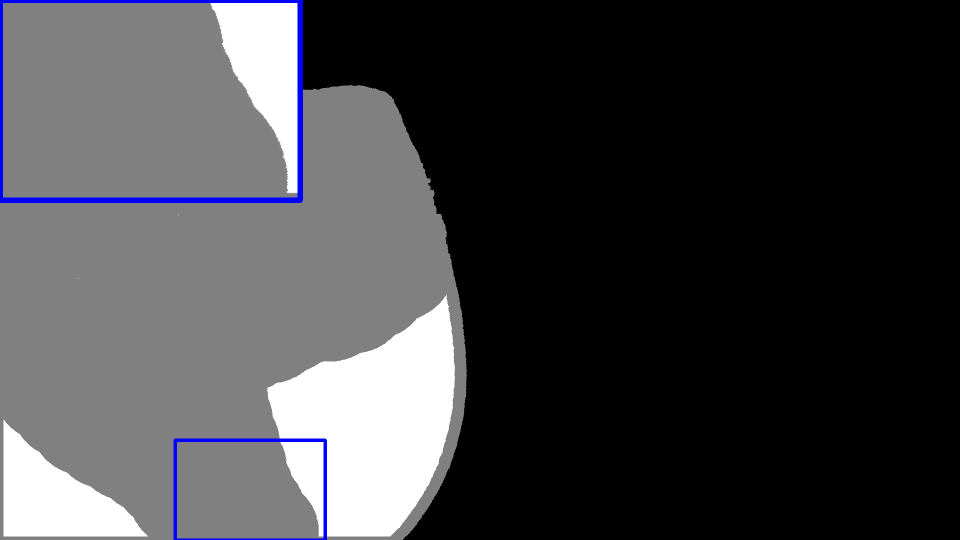}
     \includegraphics[width=\linewidth]{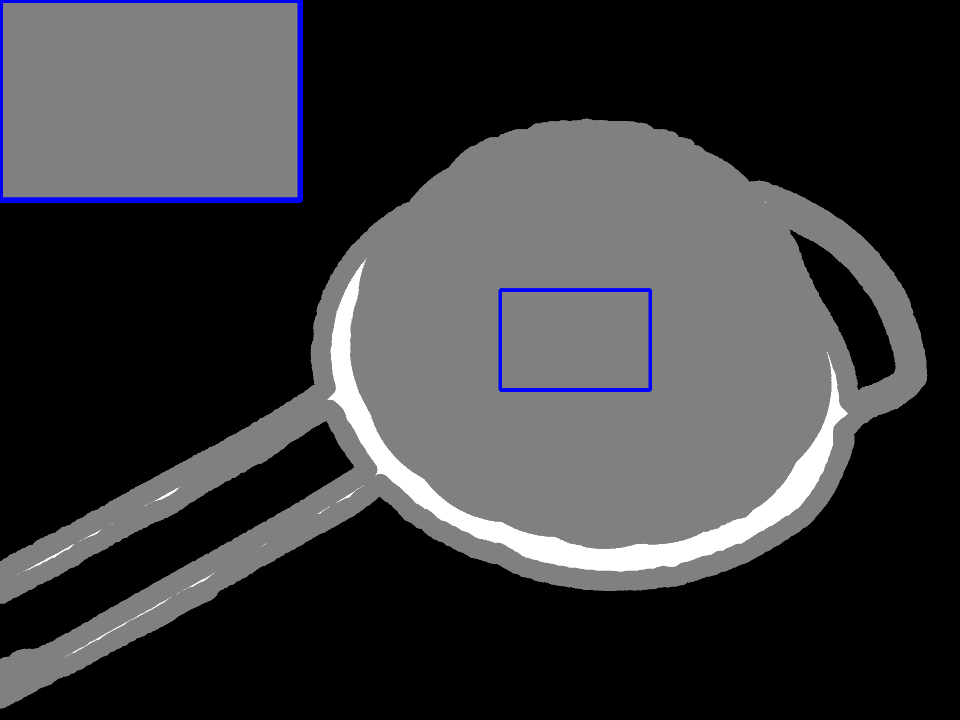}
     \includegraphics[width=\linewidth]{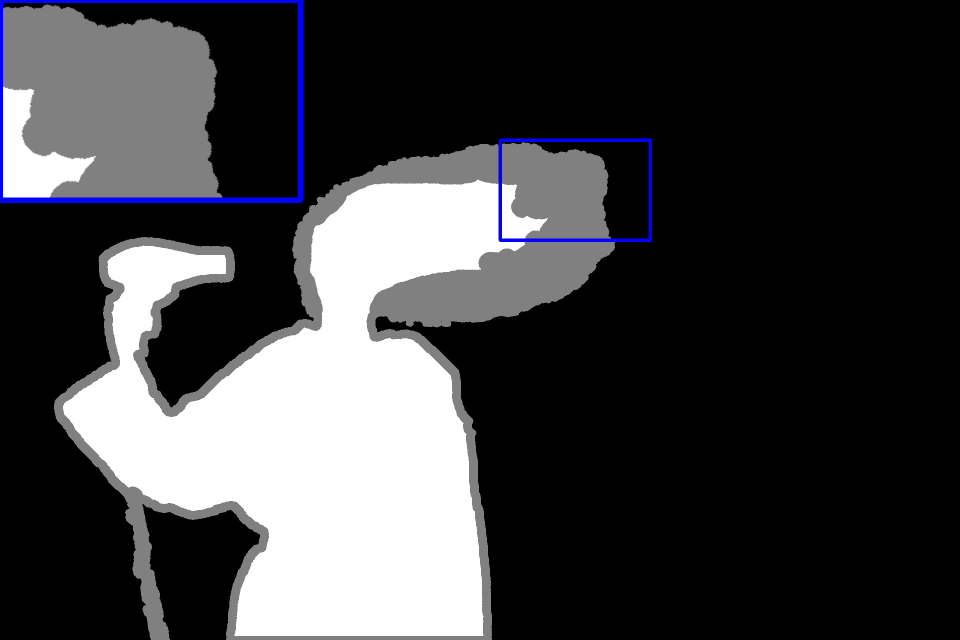}
     \end{minipage}
     }
    \hspace{-2.5mm}
    \subfloat[IndexNet~\cite{lu2019indices}]{
     \begin{minipage}{0.12\linewidth}
     \includegraphics[width=\linewidth]{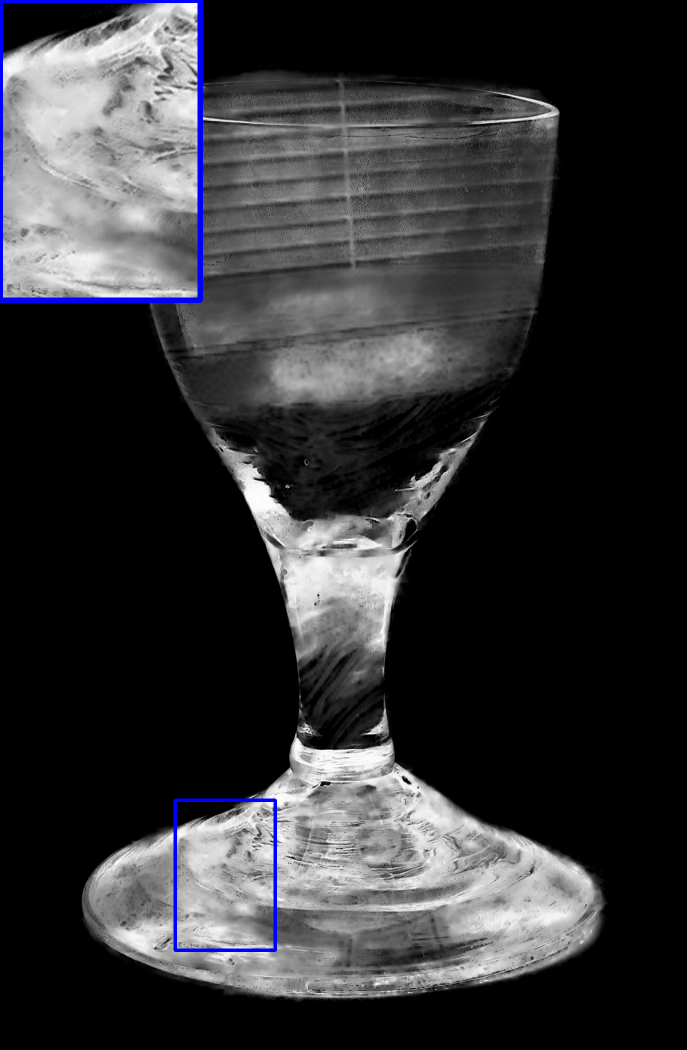}
     \includegraphics[width=\linewidth]{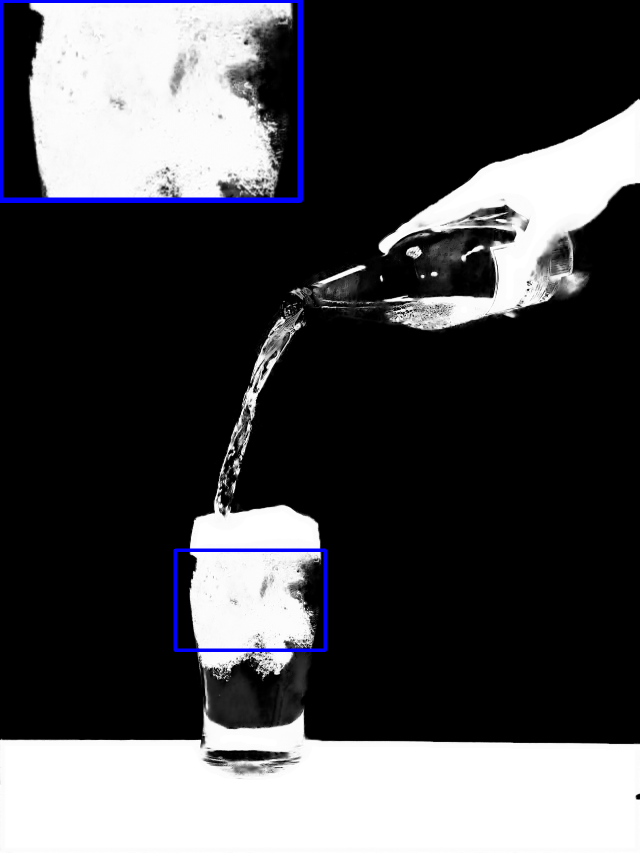}
     \includegraphics[width=\linewidth]{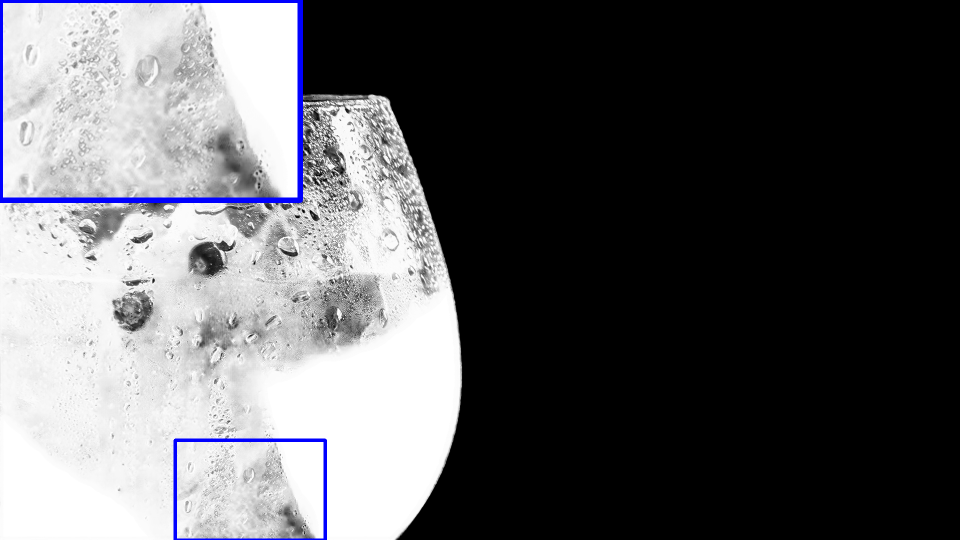}
     \includegraphics[width=\linewidth]{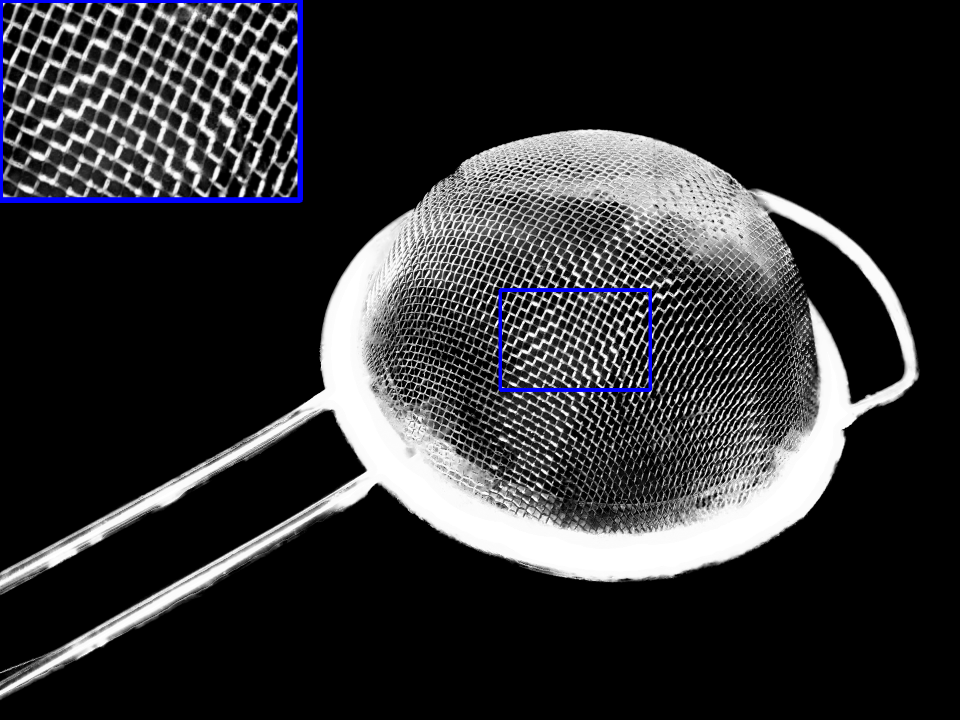}
     \includegraphics[width=\linewidth]{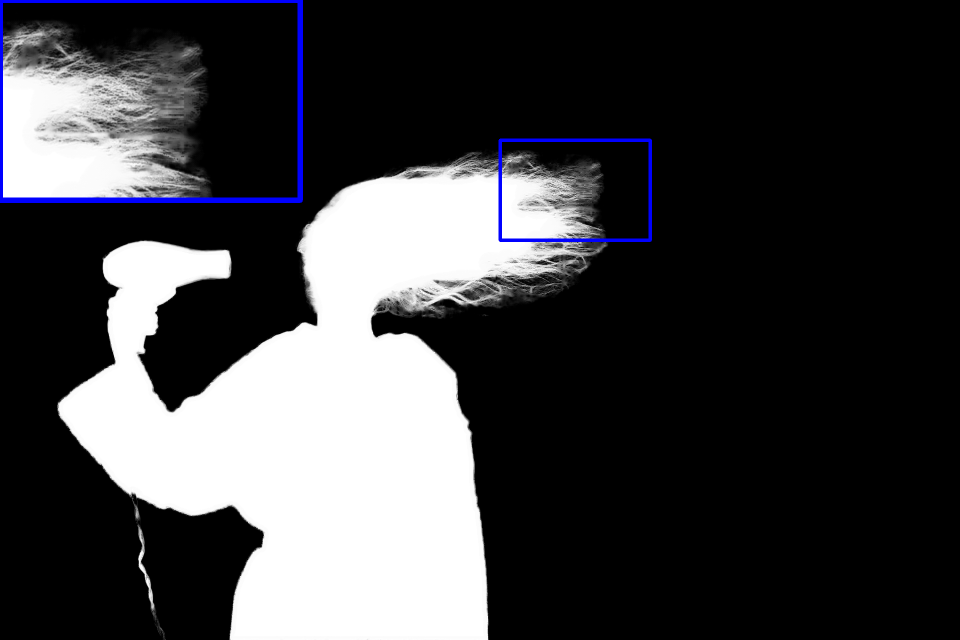}
     \end{minipage}
     }
    \hspace{-2.5mm}
    \subfloat[GCA~\cite{li2020natural} ]{
     \begin{minipage}{0.12\linewidth}
     \includegraphics[width=\linewidth]{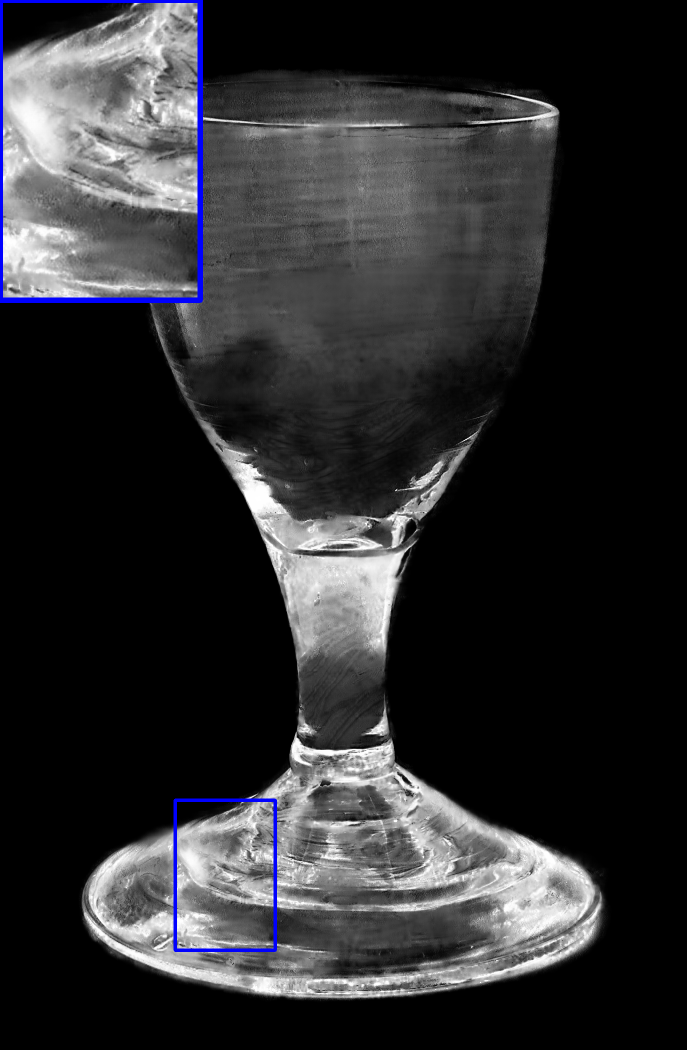}
     \includegraphics[width=\linewidth]{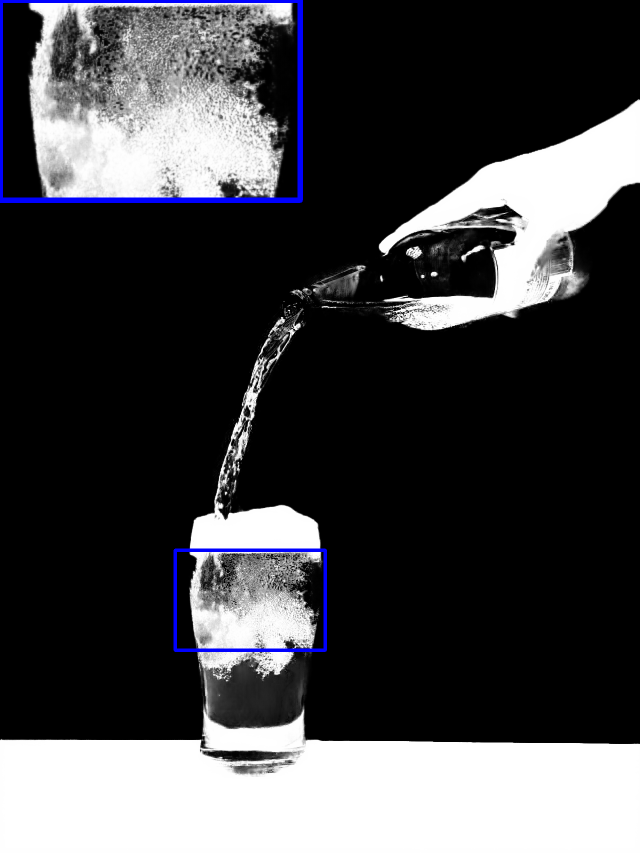}
     \includegraphics[width=\linewidth]{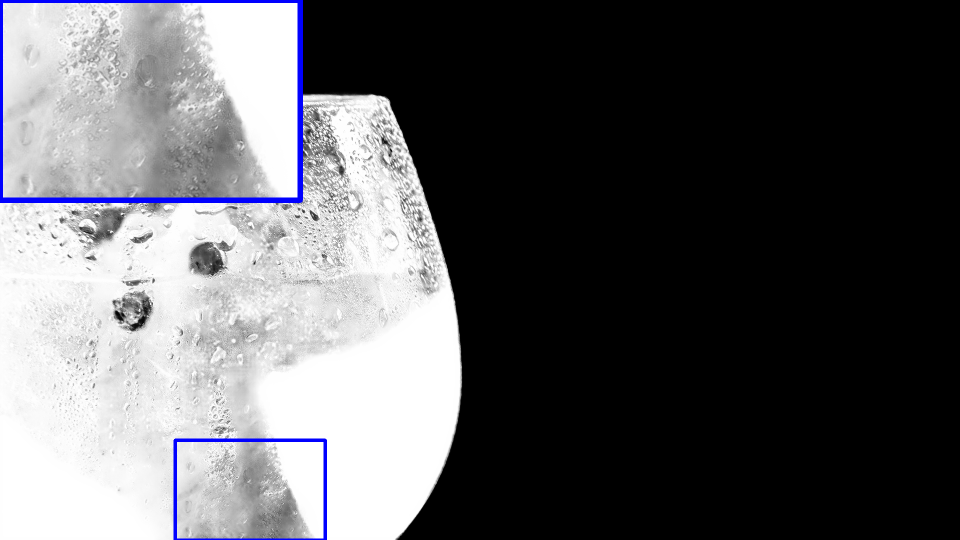}
     \includegraphics[width=\linewidth]{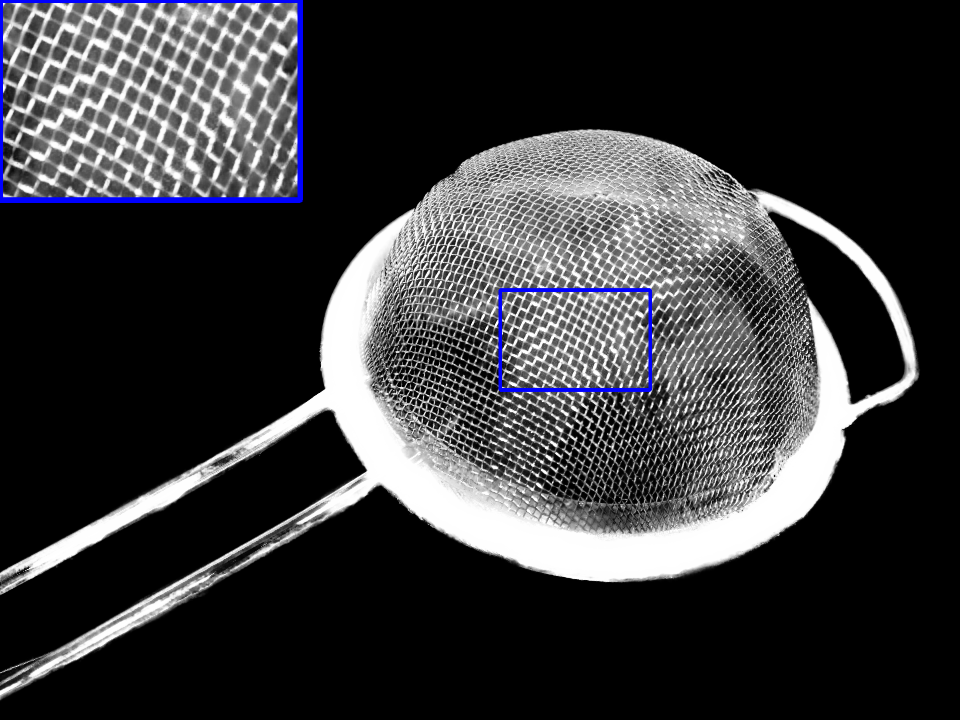}
     \includegraphics[width=\linewidth]{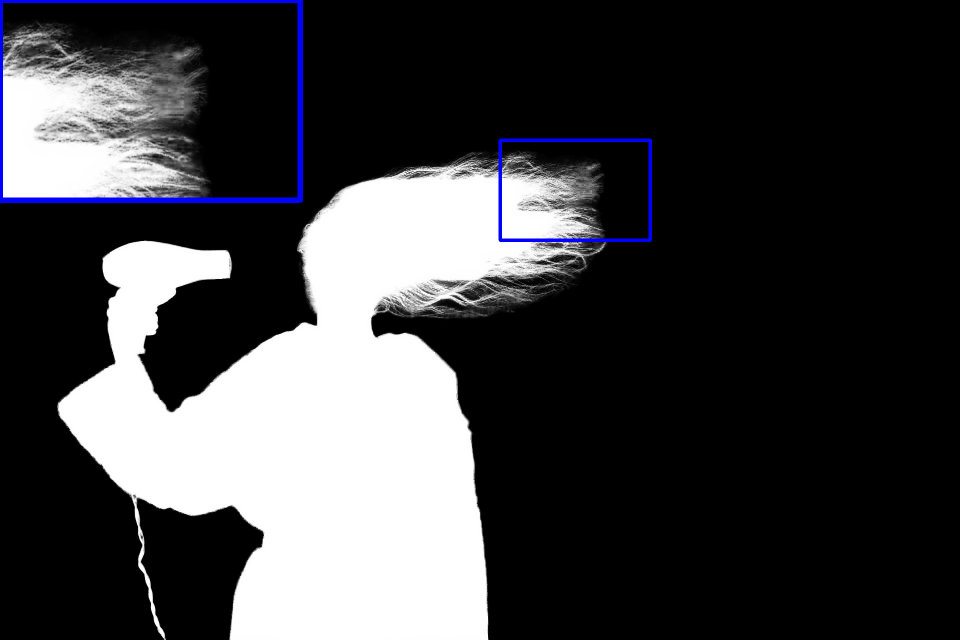}
     \end{minipage}
     }
    \hspace{-2.5mm}
    \subfloat[MG~\cite{yu2021mask}]{
     \begin{minipage}{0.12\linewidth}
     \includegraphics[width=\linewidth]{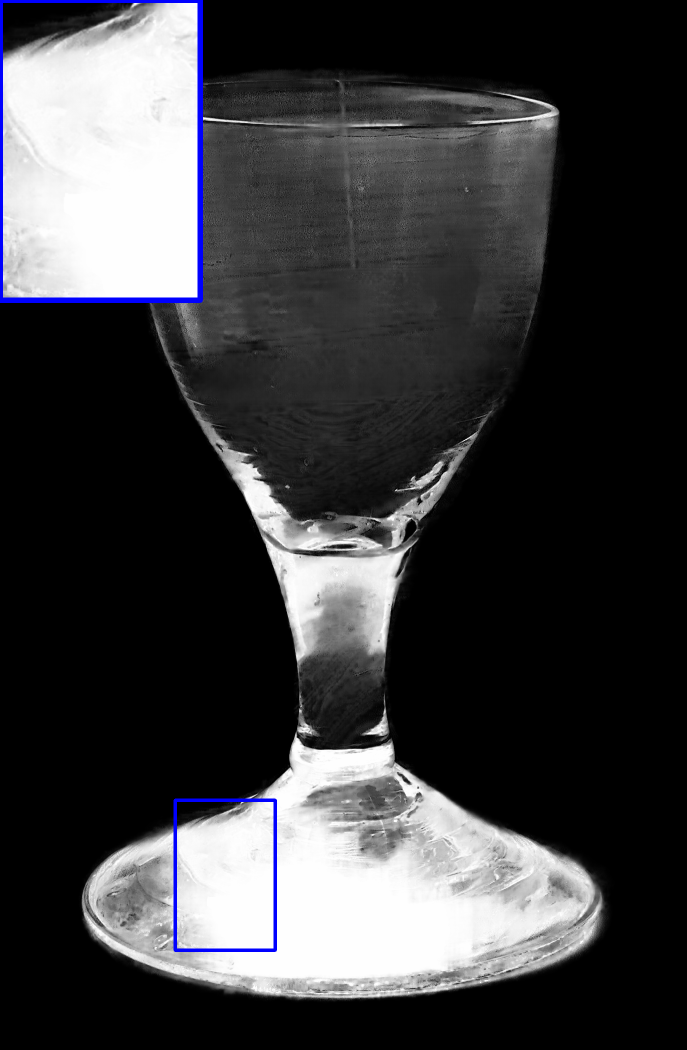}
     \includegraphics[width=\linewidth]{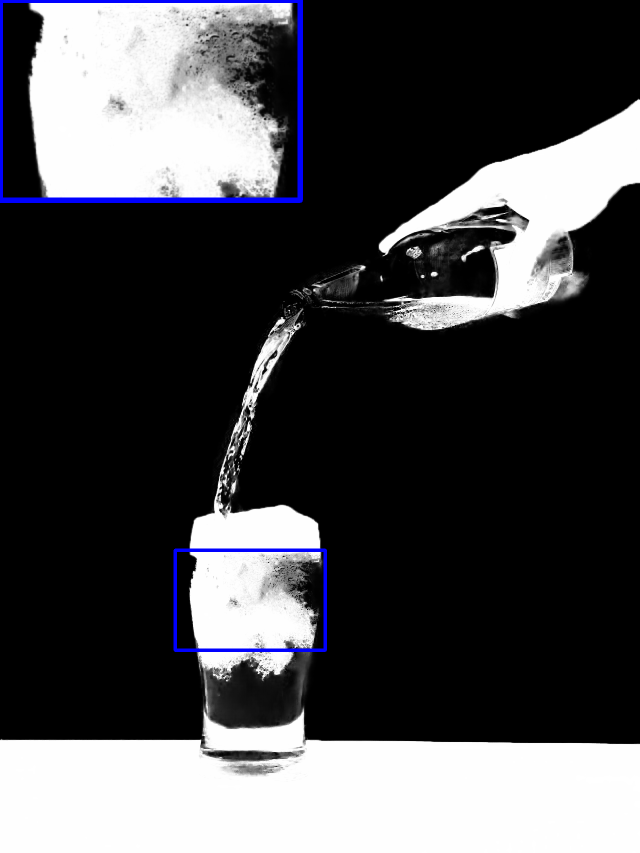}
     \includegraphics[width=\linewidth]{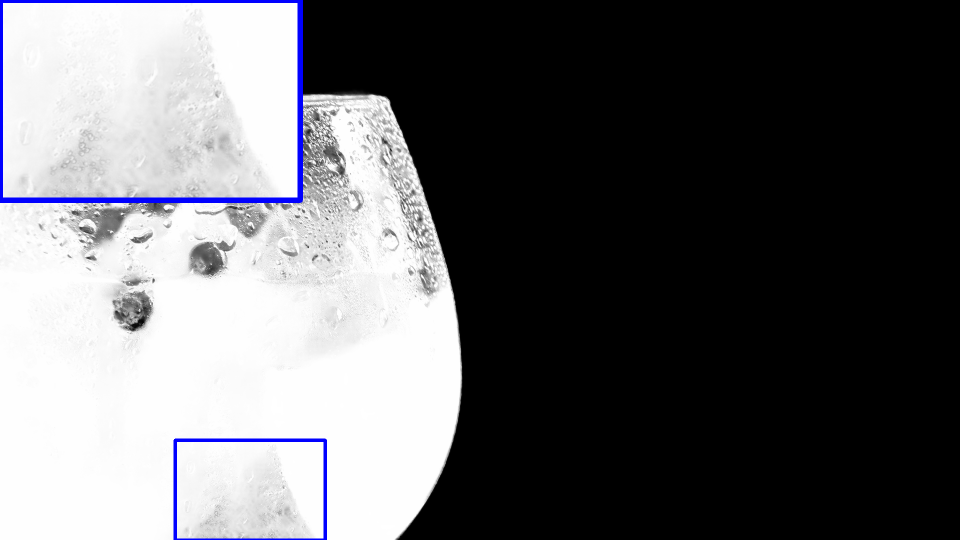}
     \includegraphics[width=\linewidth]{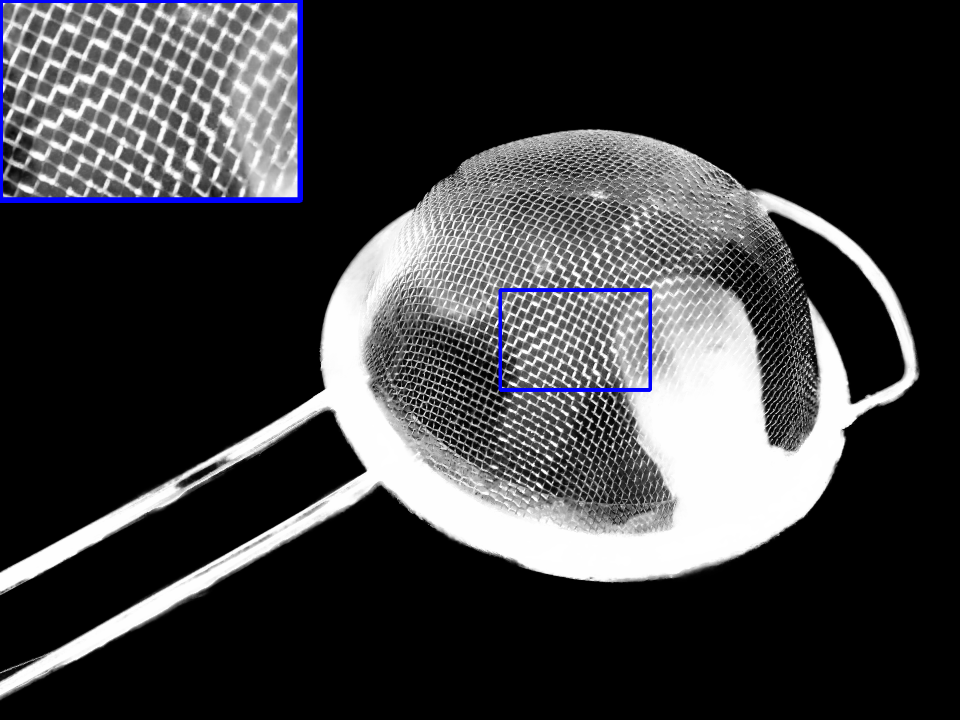}
     \includegraphics[width=\linewidth]{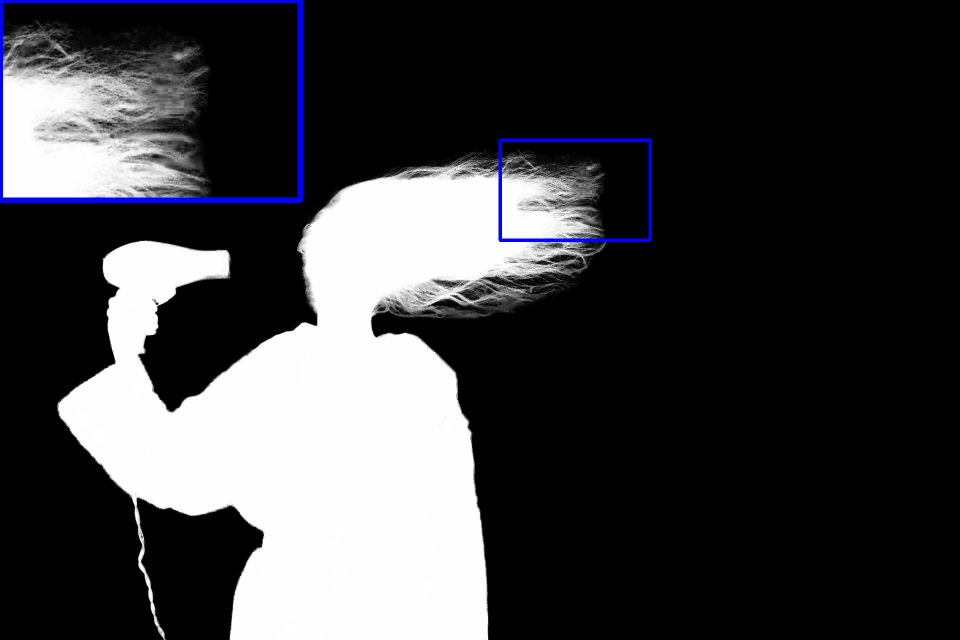}
     \end{minipage}
     }
    \hspace{-2.5mm}
    \subfloat[MatteFormer~\cite{park2022matteformer}]{
     \begin{minipage}{0.12\linewidth}
     \includegraphics[width=\linewidth]{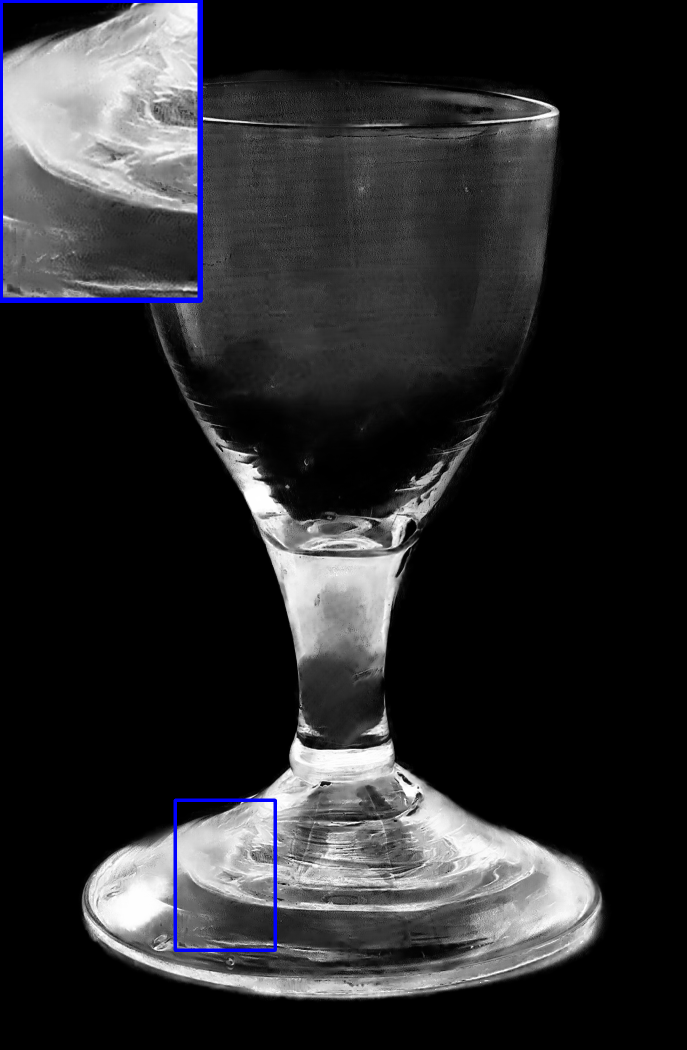}
     \includegraphics[width=\linewidth]{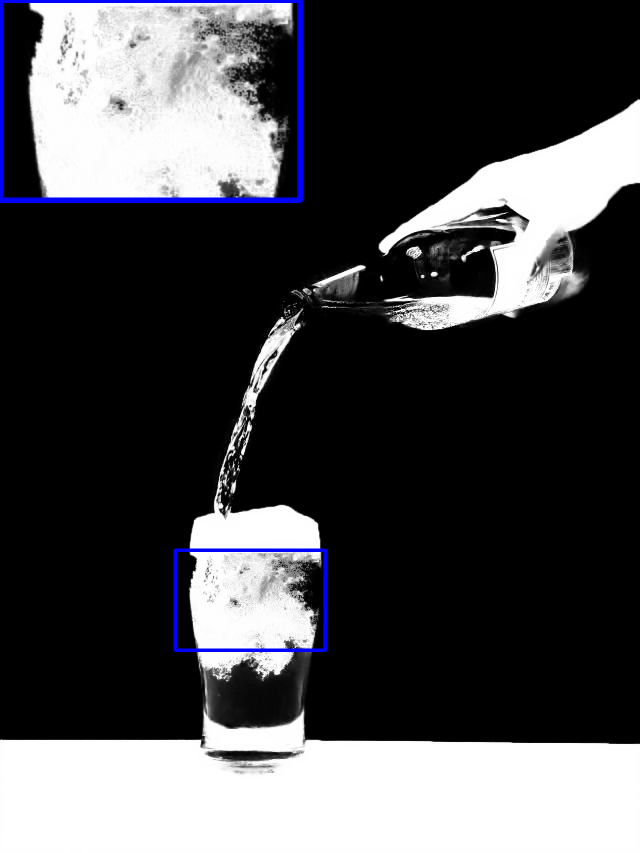}
     \includegraphics[width=\linewidth]{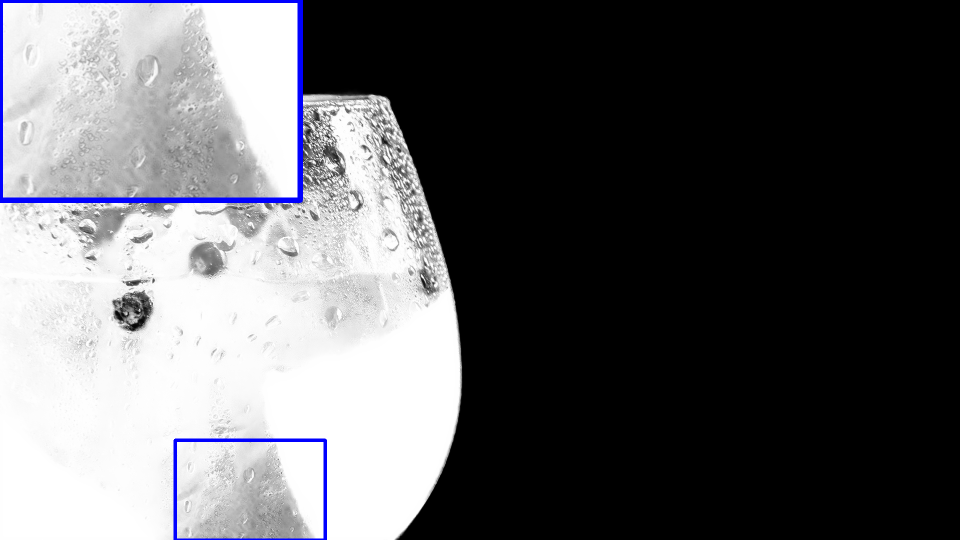}
     \includegraphics[width=\linewidth]{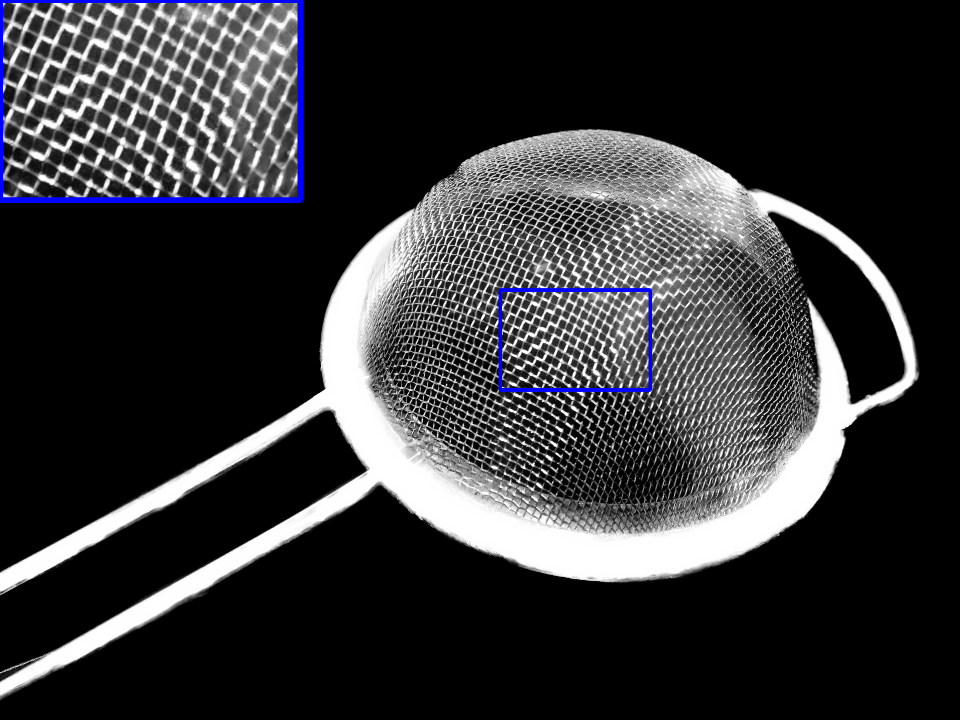}
     \includegraphics[width=\linewidth]{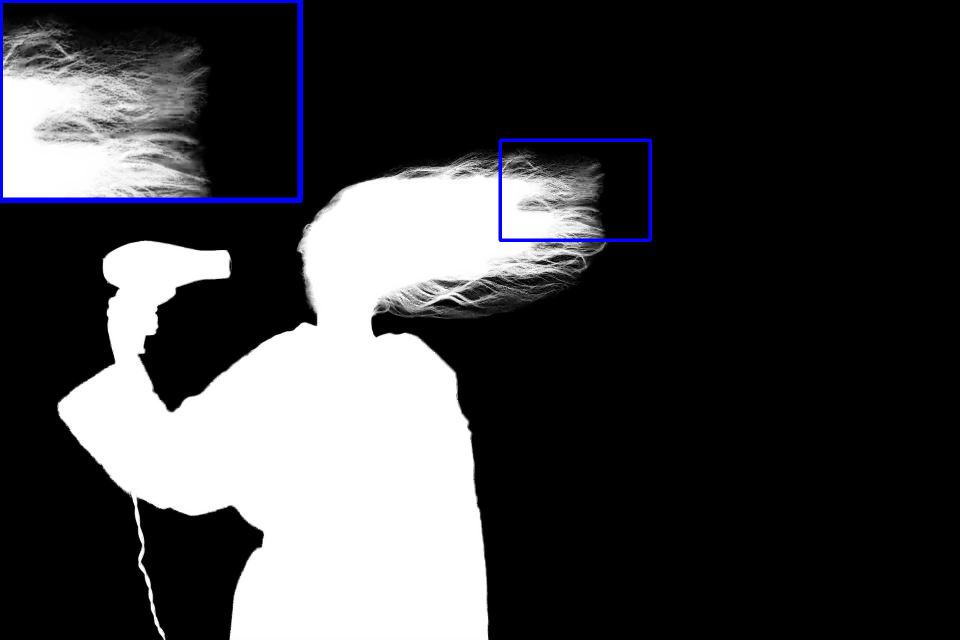}
     \end{minipage}
     }
    \hspace{-2.5mm}
    \subfloat[DiffusionMat]{
     \begin{minipage}{0.12\linewidth}
     \includegraphics[width=\linewidth]{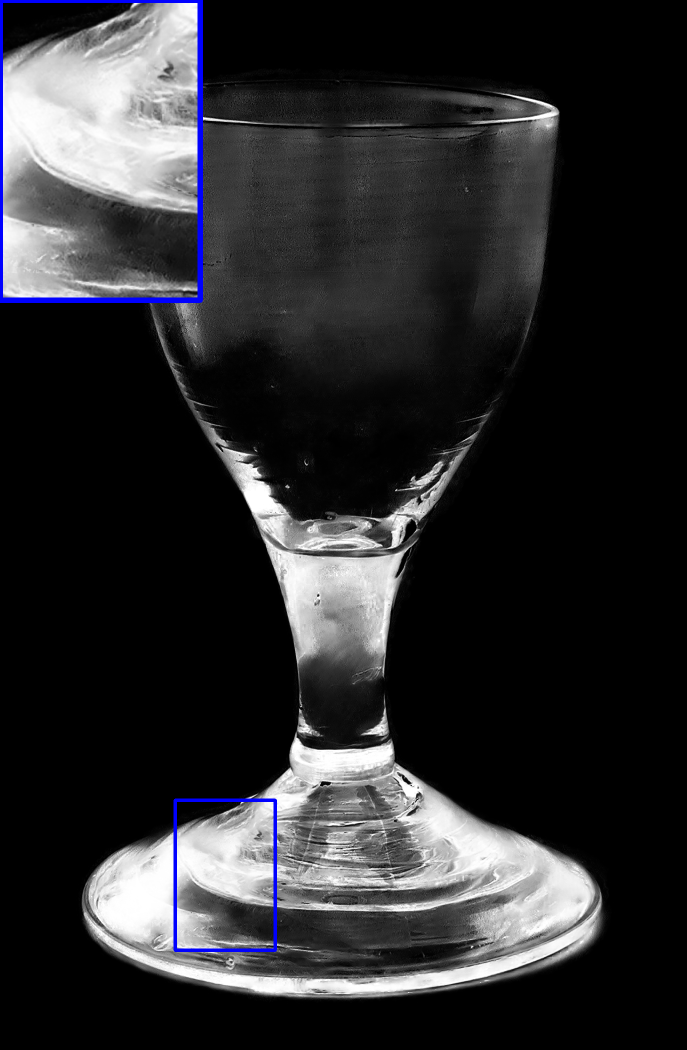}
     \includegraphics[width=\linewidth]{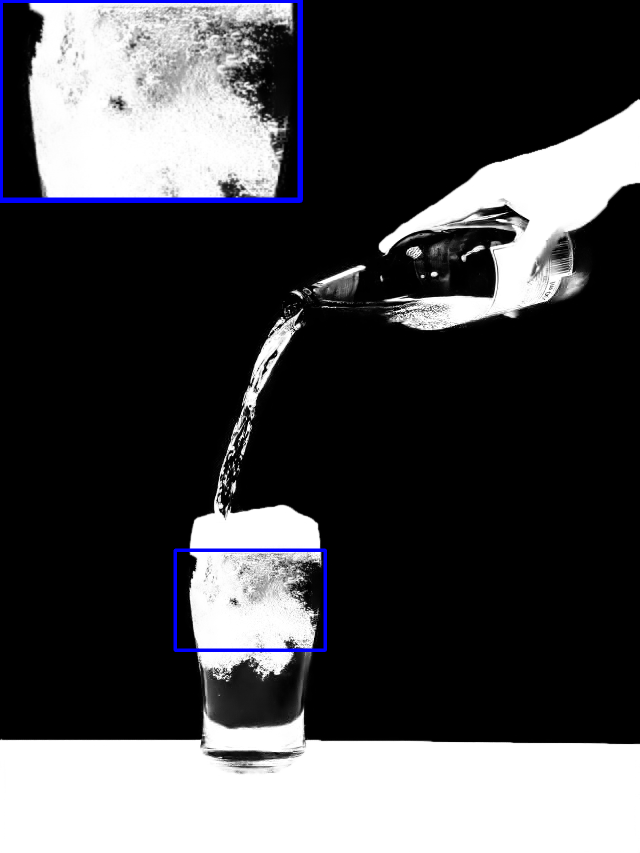}
     \includegraphics[width=\linewidth]{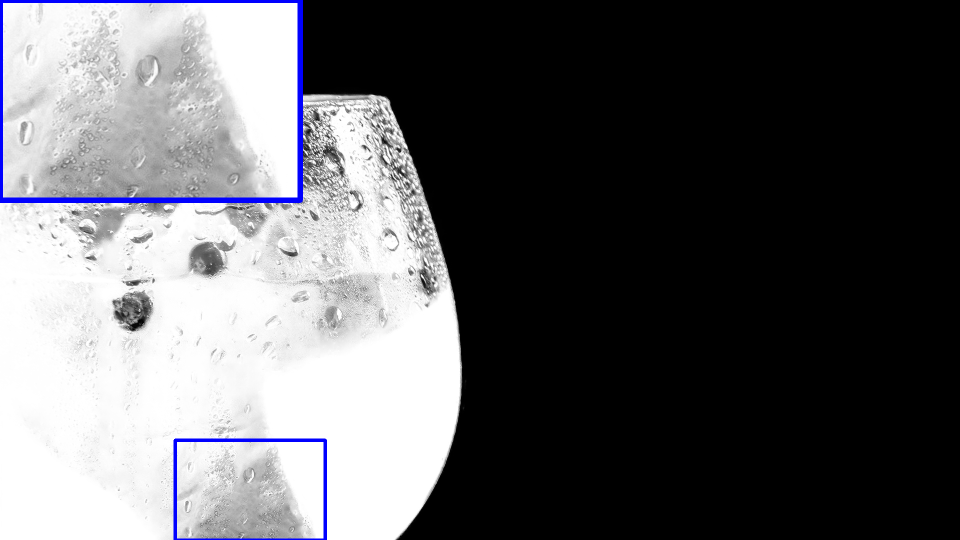}
     \includegraphics[width=\linewidth]{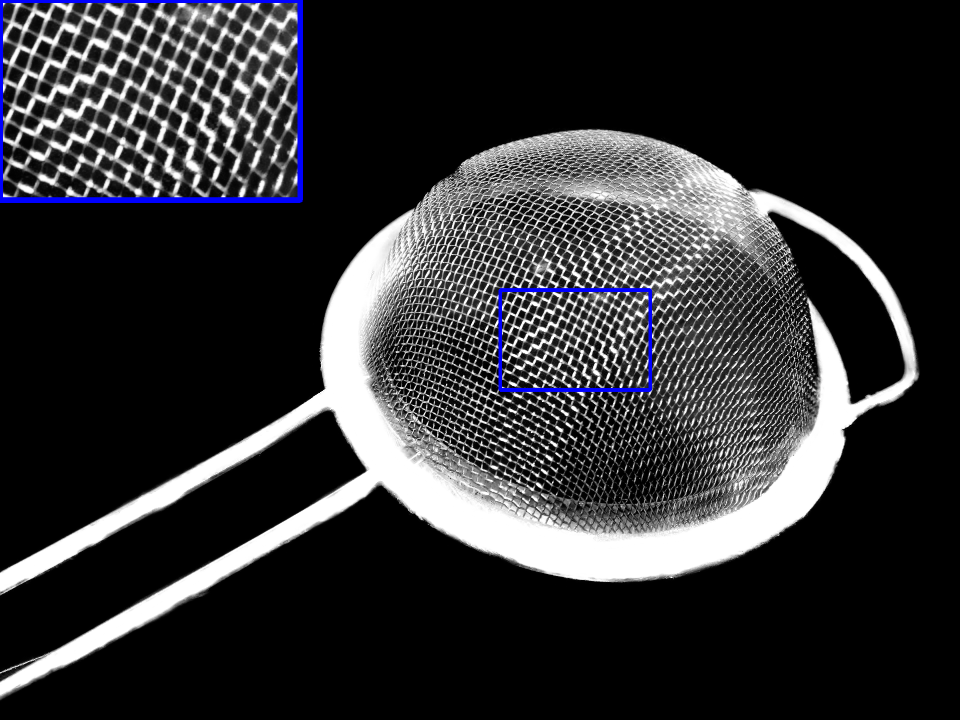}
     \includegraphics[width=\linewidth]{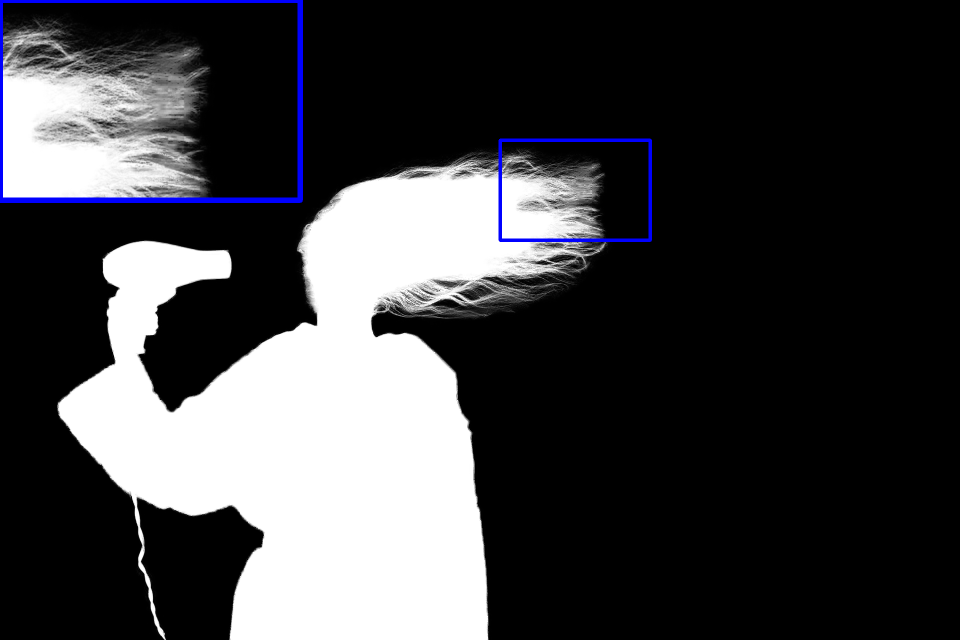}
     \end{minipage}
     }
    \hspace{-2.5mm}
    \subfloat[GT]{
     \begin{minipage}{0.12\linewidth}
     \includegraphics[width=\linewidth]{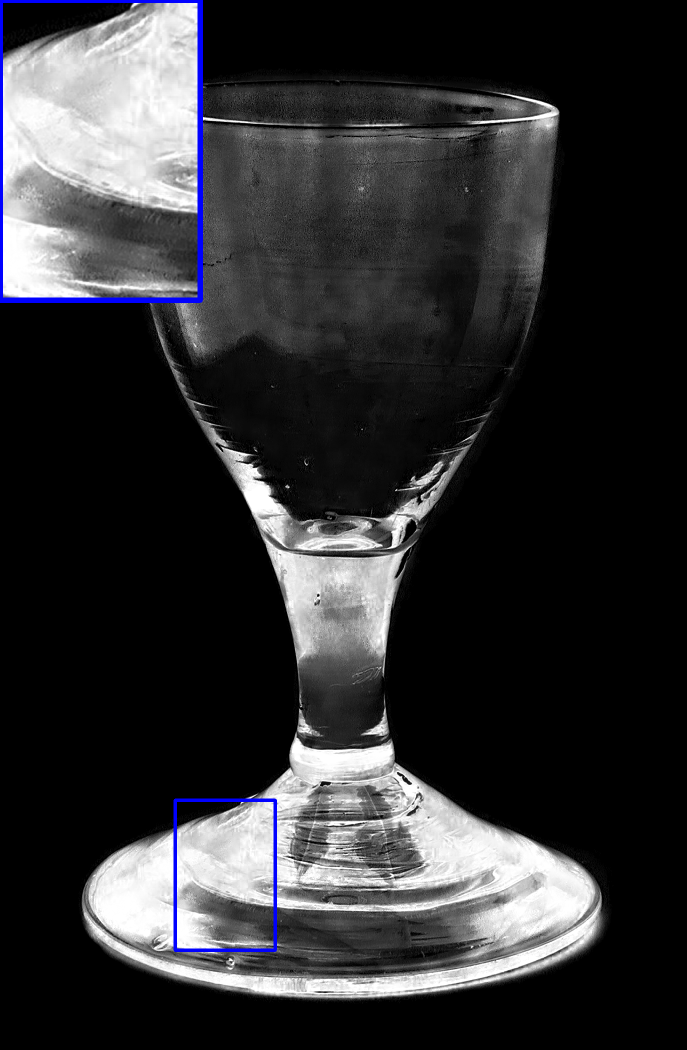}
     \includegraphics[width=\linewidth]{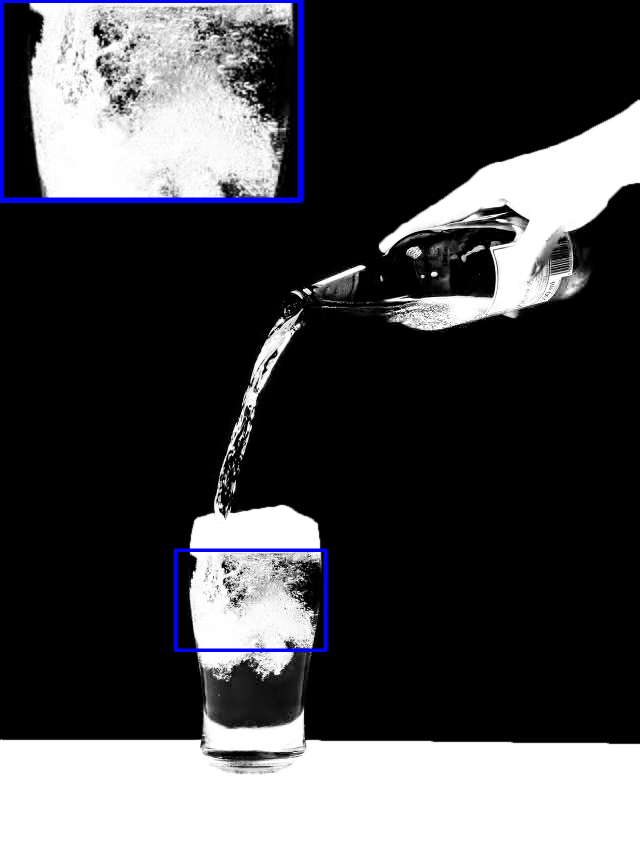}
     \includegraphics[width=\linewidth]{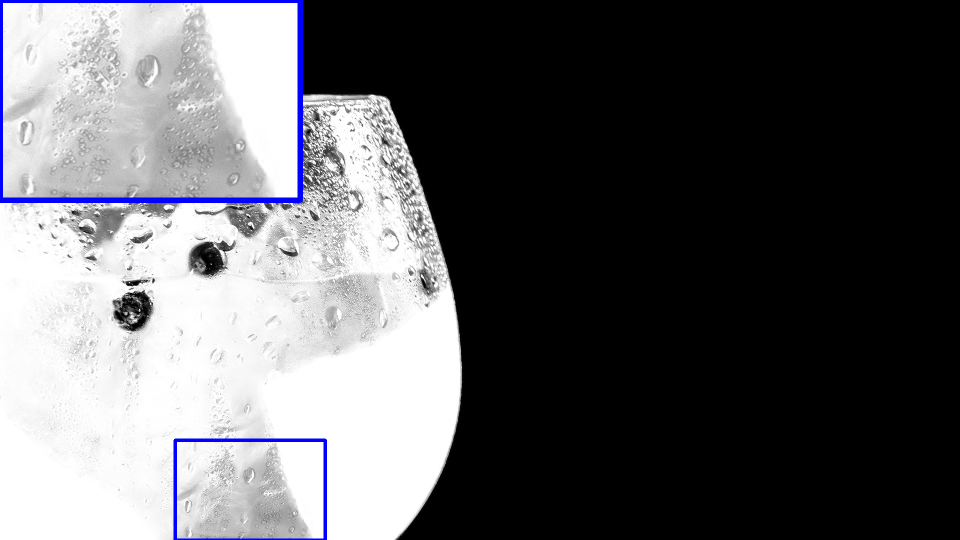}
     \includegraphics[width=\linewidth]{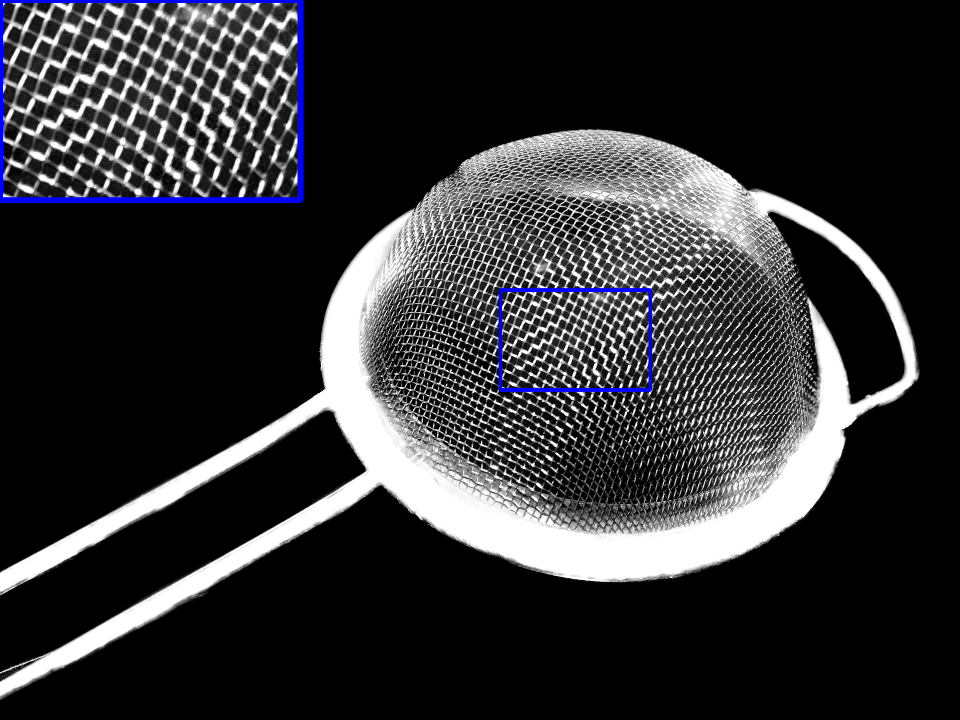}
     \includegraphics[width=\linewidth]{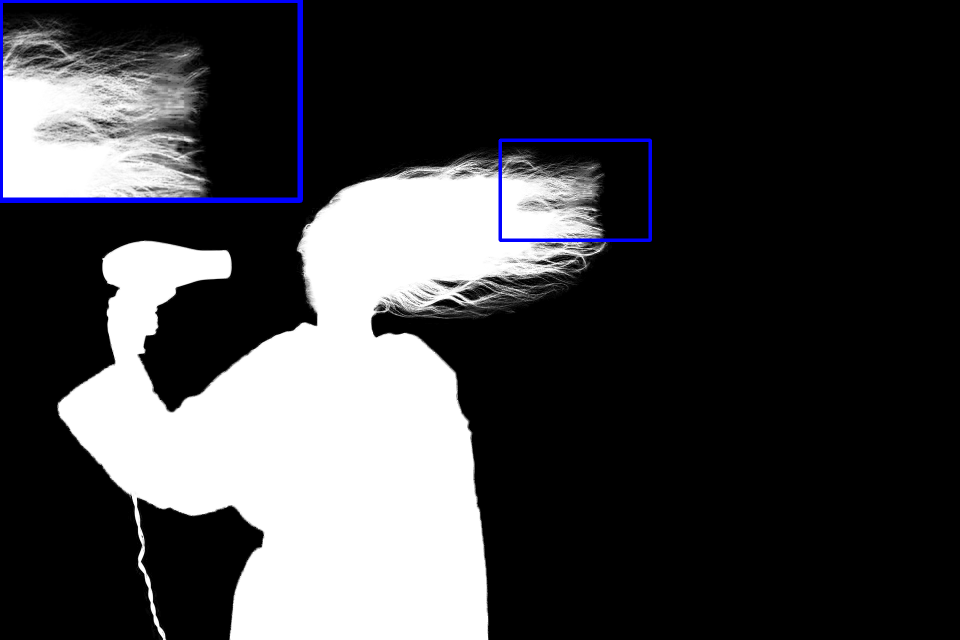}
     \end{minipage}
     }
\vspace{-3mm}
\caption{The qualitative comparison results on Composition-1k dataset. Best viewed by zooming in.}
\label{fig:adobe}
\vspace{-5mm}
\end{figure*}

\vspace{3mm}
In addition, more qualitative comparison with related works on Composition-1k dataset~\cite{xu2017deep} can be seen in Fig.~\ref{fig:adobe}. Our diffusionmat is robust to semitransparent areas, such as \emph{glass}. Moreover, it captures the local fine details effectively.

\vspace{3mm}
We also apply our DiffusionMat on real video matting dataset provided by~\cite{wang2021video}, it consists of 19 human videos results in 711 frames. We use our DiffusionMat trained on P3M dataset~\cite{li2021privacy} and evaluate on this video matting dataset. The quantitative comparison can be seen in Tab.~\ref{table:video}, though our DiffusionMat is trained on image-matting pairs, it also achieves state-of-the-art performance, which demonstrates that effectiveness of our DiffusionMat.

\begin{table}[h]
     \caption{Quantitative evaluation on image matting with state-of-the-art methods on video matting dataset. The lower the better for all metrics. The best results are marked in \textbf{Bold}.}
     \vspace{-3mm}
     \begin{center}
     \setlength{\tabcolsep}{0.25cm}{
     \begin{tabular}{c|c|c|c|c}
         \toprule
         Methods                                 &MSE$\downarrow$         &SAD$\downarrow$  &Grad$\downarrow$ &Conn$\downarrow$  \\
         \hline
         DIM~\cite{xu2017deep}                   & 13.32 & 98.92 & 129.1 & 88.56 \\
         \rowcolor{gray!20}IndexNet~\cite{lu2019indices}                                      & 10.91 & 95.07 & 120.0 & 73.05 \\
         LF~\cite{zhang2019late}                                      & 29.61 & 141.4 & 168.5 &131.7  \\
               \rowcolor{gray!20}CAM~\cite{hou2019context}                                    & 11.62 & 101.0 & 123.9 &78.21  \\
               CRGNN~\cite{wang2021video}              & 9.224 & 73.5 & 112.1 &58.49 \\
     \hline
     \rowcolor{gray!20}DiffusionMat              &\textbf{7.789}    &\textbf{70.3} &\textbf{101.0} &\textbf{52.13}  \\
     \bottomrule
     \end{tabular}
     }
     \end{center}
     \label{table:video}
\end{table}

\clearpage

\begin{algorithm}[t]
\small
\caption{\small  Training Procedure of~{DiffusionMat.}
}
\label{alg:train}
\definecolor{codeblue}{rgb}{0.25,0.5,0.5}
\definecolor{codegreen}{rgb}{0,0.6,0}
\definecolor{codekw}{RGB}{207,33,46}
\lstset{
  backgroundcolor=\color{white},
  basicstyle=\fontsize{7.5pt}{7.5pt}\ttfamily\selectfont,
  columns=fullflexible,
  breaklines=true,
  captionpos=b,
  commentstyle=\fontsize{7.5pt}{7.5pt}\color{codegreen},
  keywordstyle=\fontsize{7.5pt}{7.5pt}\color{codekw},
  escapechar={|}, 
}
\begin{lstlisting}[language=python]
def train(images, trimaps, alphas, steps, T):
  """
  images: [B, H, W, 3]
  trimap: [B, H, W, 1]
  alphas: [B, H, W, 1]
  steps: manual defined sample steps
  T: manual defined corrupted timestep
  B: batch
  """
  
  # Encode image features
  feats = image_encoder(images)

  # Perturb guidance mask
  m_t = perturb_operation(m, T)

  # Get the known mask
  m_known = trimap==0 + trimap==1
  
  #Set time pairs
  time_pairs = set_timepairs(T, steps) 

  #Invert alphas into matte diffusion model
  Alphas_Invs = {}
  for t_now, t_next in zip(time_pairs):
      alphas_inv = ddim_step(alphas)
      Alphas_Invs[t_now]=alphas_inv

  #Invert trimaps into matte diffusion model
  Trimaps_Invs = {}
  for t_now, t_next in zip(time_pairs):
      trimaps_inv = ddim_step(trimaps)
      Trimaps_Invs[t_now]=trimaps_inv

  losses = 0
  for t_now, t_next in zip(reversed(time_pairs)):

      # Correct the perturbed mask 
      m_t^correct = correction_network(m_t, feats, t_now)

      # Get the current timestep's inverted trimap for replacement
      trimap_inv = Trimaps_Invs[t_now]

      # Alpha Reliability Propagation
      m_t^ARP = m_t^correct * (1-m_known) + trimap_inv * m_known

      # DDIM sampling & One step denoising
      m_t, alpha_0 = ddim_step(m_t^ARP)

      # Get the current timestep's inverted alpha for supervision
      alpha_inv = Alphas_Invs[t_now]

      # Set matting loss 
      loss = set_matting_loss(m_t, alpha_0, alpha_inv, alpha)
      losses += loss
  return losses
\end{lstlisting}

\end{algorithm}
\begin{algorithm}[t]
\small
\caption{\small  Inference Procedure of~{DiffusionMat.}
}
\label{alg:sample}
\definecolor{codeblue}{rgb}{0.25,0.5,0.5}
\definecolor{codegreen}{rgb}{0,0.6,0}
\definecolor{codekw}{RGB}{207,33,46}
\lstset{
  backgroundcolor=\color{white},
  basicstyle=\fontsize{7.5pt}{7.5pt}\ttfamily\selectfont,
  columns=fullflexible,
  breaklines=true,
  captionpos=b,
  commentstyle=\fontsize{7.5pt}{7.5pt}\color{codegreen},
  keywordstyle=\fontsize{7.5pt}{7.5pt}\color{codekw},
  escapechar={|}, 
}
\begin{lstlisting}[language=python]
def inference(images, trimaps, steps, T):
  """
  images: [B, H, W, 3]
  trimap: [B, H, W, 1]
  steps: manual defined sample steps
  T: manual defined corrupted timestep
  B: batch
  """
  
  # Encode image features
  feats = image_encoder(images)

  # Perturb guidance mask
  m_t = perturb_operation(m, T)

  # Get the known mask
  m_known = trimap==0 + trimap==1
  
  #Set time pairs
  time_pairs = set_timepairs(T, steps) 

  #Invert trimaps into matte diffusion model
  Trimaps_Invs = {}
  for t_now, t_next in zip(time_pairs):
      trimaps_inv = ddim_step(trimaps)
      Trimaps_Invs[t_now]=trimaps_inv

  for t_now, t_next in zip(reversed(time_pairs)):

      # Correct the perturbed mask 
      m_t^correct = correction_network(m_t, feats, t_now)

      # Get the current timestep's inverted trimap for replacement
      trimap_inv = Trimaps_Invs[t_now]

      # Alpha Reliability Propagation
      m_t^ARP = m_t^correct * (1-m_known) + trimap_inv * m_known

      # DDIM sampling & One step denoising
      m_t, alpha_0 = ddim_step(m_t^ARP)

     
  return m_t
\end{lstlisting}

\end{algorithm}

\clearpage

{\small
\bibliographystyle{ieee_fullname}
\bibliography{egbib}
}

\end{document}